\pdfoutput=1

\documentclass[11pt]{article}
\usepackage[final]{acl}

\usepackage{times}
\usepackage{latexsym}

\usepackage[T1]{fontenc}
\usepackage[utf8]{inputenc}

\usepackage{microtype}

\usepackage{inconsolata}

\usepackage{graphicx}

\usepackage{amsmath}

\usepackage{cleveref}

\usepackage{subcaption}
\usepackage{booktabs}
\usepackage{array}

\usepackage{caption}
\usepackage{graphicx}

\usepackage{xcolor}
\usepackage{booktabs}
\usepackage{multirow}
\usepackage{xspace}
\usepackage{enumitem}

\usepackage{tabularx}
\newcolumntype{b}{X}
\newcolumntype{s}{>{\hsize=.4\hsize}X}
\usepackage{longtable}

\usepackage{todonotes}
\definecolor{ruby}{rgb}{0.8,0.2,0.4}
\definecolor{olive}{rgb}{0.1,0.55,0.1}
\definecolor{navy}{rgb}{0.2,0.2,0.7}

\newcommand{\falsealarm}[1]{{\color{ruby}{#1}}}
\newcommand{\correct}[1]{{\color{olive}{#1}}}

\newcommand{\hide}[1]{}

\newcommand{\modelname}{MoVa\xspace}
\newcommand{\partextmoveup}{\vspace{-0.25em}}
\usepackage{float} \usepackage{placeins}
\usepackage{amssymb}
\usepackage{array}
\usepackage{makecell} 
\usepackage{etoc}
\usepackage[most]{tcolorbox} 
\usepackage{lipsum} 
\usepackage{tabularx}
\usepackage{todonotes}
\newcommand{\secmoveup}{\vspace{-1.2mm}}
\newcommand{\sectextmoveup}{\vspace{-1.4mm}}
\newcommand{\captionmoveup}{\vspace{-0.5mm}}
\newcommand{\afterfigmoveup}{\vspace{-0mm}}
\newcommand{\headingmoveup}{\vspace{-0mm}}

\newcolumntype{M}[1]{>{\centering\arraybackslash}m{#1}}
\newcolumntype{P}[1]{>{\raggedright\arraybackslash}p{#1}}

\title{\modelname: Measuring Moral Dimensions and Human Values via Large Language Models}
\title{\modelname: Generalizable Measurements of Human Morals and Values}
\title{\modelname: Measuring Generalization of Human Morals and Values}
\title{\modelname: A Generalizable Approach to Classifying Human Morals and Values}
\title{\modelname: Resources for Generalizable Classification \\ of Human Morals and Values}

\title{\modelname: Towards Generalizable Classification of Human Morals and Values}

\author{
 \textbf{Ziyu Chen\textsuperscript{1}},
 \textbf{Junfei Sun\textsuperscript{2}},
 \textbf{Chenxi Li\textsuperscript{2}},
 \textbf{Tuan Dung Nguyen\textsuperscript{3}},
 \textbf{Jing Yao\textsuperscript{4}},
\\
 \textbf{Xiaoyuan Yi\textsuperscript{4}},
 \textbf{Xing Xie\textsuperscript{4}},
 \textbf{Chenhao Tan\textsuperscript{2}},
 \textbf{Lexing Xie\textsuperscript{1}}
\\
 \textsuperscript{1}The Australian National University,
 \textsuperscript{2}University of Chicago,
 \\
 \textsuperscript{3}University of Pennsylvania,
 \textsuperscript{4}Microsoft Research Asia
\\
 Contact: 
 \{ziyu.chen, lexing.xie\}@anu.edu.au
}

\begin{document}
\maketitle
\begin{abstract}
\sectextmoveup
Identifying human morals and values embedded in language is essential to empirical studies of communication. However, researchers often face substantial difficulty navigating the diversity of theoretical frameworks and data available for their analysis. Here, we contribute \modelname, a well-documented suite of resources for generalizable classification of human morals and values, consisting of (1) 16 labeled datasets and benchmarking results from four theoretically-grounded frameworks; (2) a lightweight LLM prompting strategy that outperforms fine-tuned models across multiple domains and frameworks; and (3) a new application that helps evaluate psychological surveys. In practice, we specifically recommend a classification strategy, \emph{all@once}, that scores all related concepts simultaneously, resembling the well-known multi-label {\it classifier chain}. The data and methods in \modelname can facilitate many fine-grained interpretations of human and machine communication, with potential implications for the alignment of machine behavior.\footnote{Data and code supporting this paper can be found at \url{https://github.com/ZiyuChen0410/MoVa2025}.}
\end{abstract}

\secmoveup 
\section{Introduction}
\sectextmoveup

Reliably scoring psychological constructs in text is crucial for addressing many questions in computational social science~\citep{grimmer2013text}. Influential efforts in this domain have focused on sentiment and toxicity detection, leading to a range of automated tools.
A new frontier lies in identifying human morals and values, the important traits that fundamentally shape both individual and collective responses to issues such as climate change~\cite{dickinson2016moral}, public health decisions~\citep{gert2006bioethics}, vaccine uptake~\cite{aminAssociationMoralValues2017}, and political alignment~\citep{piurko2011SchwartzPolitics, grahamLiberalsConservativesRely2009a}. 
Understanding morals and values has also been central to aligning AI with humans, where the great volume and diversity of AI-generated text demand a new approach in data labeling~\citep{pawarSurveyCulturalAwareness2024}.

\begin{figure*}[!htbp]
    \centering
    \includegraphics[width=\linewidth]{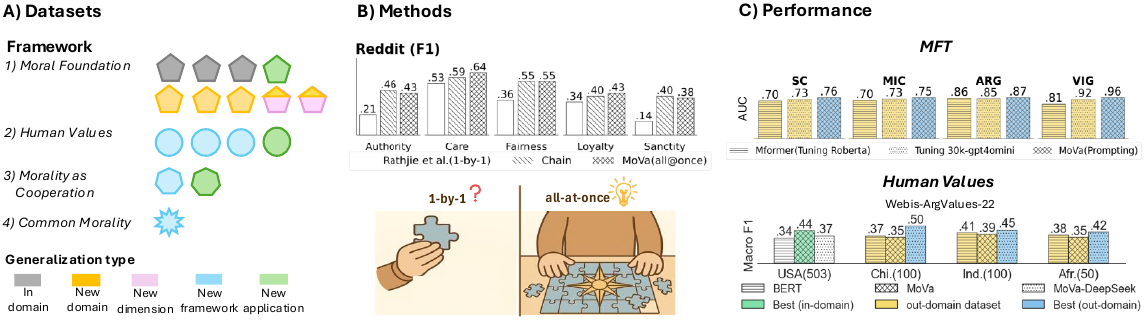}
    \caption{
    Overview of \modelname resources.
    (A) Datasets: \modelname covers 16 datasets on four moral and value frameworks: MFT, Human Values, Common Morality, and MAC (see \Cref{tab:dataset_overview}). Shapes represent distinct frameworks. Colors indicate generalization types, including new domains (yellow), dimensions (pink), frameworks (blue), and applications (green).
    (B) Methods: \modelname prompt asks LLMs to predict all labels together, rather than one at a time.
    This strategy significantly improves F1 scores.
    (C) Performance: Prompting with \modelname outperforms finetuned models across new data domains in MFT and new frameworks such as Human Values.}
    \label{fig:teaser1}
\end{figure*}

Generalizability in classifying human morals and values remains critical but underexplored. In this context, two concerns are especially important. First, a classifier must handle the \textit{linguistic diversity} of data across domains within the same framework, ranging from short, informal texts with fewer than ten words to long, formal passages in published work (see \Cref{tab:dataset_overview}). Existing approaches are often limited to just one domain~\citep{hooverMoralFoundationsTwitter2020a,tragerMoralFoundationsReddit2022,liscioCrossDomainClassificationMoral2022}.
Second, the classifier should handle the prediction of constructs proposed \emph{by different frameworks}, such as Moral Foundations Theory~\citep{grahamLiberalsConservativesRely2009a} and Human Values~\citep{schwartz2010basic} (see \Cref{fig:teaser1}). Existing work has largely focused on curating data and training classifiers only for a single framework~\citep{frimerMoralFoundationsDictionary2019c,hoppExtendedMoralFoundations2021,grahamMoralFoundationsDictionary2012}. General-purpose LLMs appear capable of addressing both concerns, yet recent evaluations suggested that they still fall short of fine-tuned models~\citep{rathje2024gpt,Dehghani2024gpt}. In contrast, we find that the right prompting strategy can offer better generalization than fine-tuning.

In this work, we contribute \textbf{\modelname}, a set of resources that define and operationalize {\it generalizable} classification of human {\bf Mo}rals and {\bf Va}lues.  As shown in \Cref{fig:teaser1}, MoVa comprises: 
\begin{itemize}[leftmargin=*,noitemsep,nolistsep]
   \item 16 \textit{labeled} datasets and benchmarking results across four moral and value frameworks, including five resources that we reformulate into classification tasks (Section \ref{sec:data}).
   \item A prompt-based lightweight classification tool, observed to generalize well to new data domains, dimensions, and frameworks (Section \ref{sec:methods}, with evaluations in Sections \ref{sec:mft}, \ref{sec:schwartz} and \ref{sec:MAC_commonmorality}).
   \item A new application that helps evaluate psychological surveys (Section \ref{sec:questionaire}).
\end{itemize}

Within the moral foundations framework, for illustration, we show that the key to stronger classification performance lies in prompting LLMs to label all constructs at once (\textit{all@once}), rather than one at a time (\textit{1-by-1}) in separate requests. Classifier-chain analysis suggests this advantage may arise from leveraging label dependencies. When applied to three other moral and value frameworks, this strategy also performs comparably to, or even better than, fine-tuned models.

In summary, \modelname provides a useful benchmark and tool to analyze large text corpora in human morals and values. Future work could extend \modelname prompting strategies to other forms of subjective text analysis after rigorous evaluations.

\secmoveup 
\section{Related work}\label{sec:related_work}
\sectextmoveup

NLP research has been concerned with many types of generalization~\cite{hupkes2023taxonomy} or meta-learning~\cite{lee2022meta} problems. \modelname is particularly concerned with classification tasks in new data domains, constructs derived from different moral and value frameworks, and robustness in task setup, such as classifying descriptions of action rather than mere opinion.

A set of recent work elicits LLMs’ moral preferences using psychological questionnaires~\citep{abdulhai2023moral}, human-annotated moral scenarios~\citep{ScherrerMoralChoice}, value-aware prompts including adversarial questions~\citep{yao2024value}, and explores aligning AI systems with humans \citep{zheng2024ali, tennant2024moral,yao2024clave}. \modelname uses LLMs as a tool for a narrowly defined task: classifying human morals and values in any text, with LLM-generated text as an increasingly important use case.

\modelname builds on a long line of work on classifying text on human morals and values, beginning with word counts from expert-crafted dictionaries~\citep{frimerMoralFoundationsDictionary2019c,grahamMoralFoundationsDictionary2012,hoppExtendedMoralFoundations2021}, followed by machine learning classifiers trained or fine-tuned on specific domains~\citep{hooverMoralFoundationsTwitter2020a,tragerMoralFoundationsReddit2022,liscioCrossDomainClassificationMoral2022,guoDataFusionFramework2023,nguyen2024measuring}, and most recently, zero-shot classification using LLMs without domain or task generalization~\cite{rathje2024gpt}. See \Cref{appendix:relatedwork} for an extended discussion.

\secmoveup 
\section{MoVa Frameworks and Datasets}
\label{sec:data}
\sectextmoveup
This section introduces \modelname's benchmark datasets. The diversity across frameworks and datasets directly motivates the design of our methods in Section~\ref{sec:methods} and the evaluation setup in Sections~\ref{sec:mft}–\ref{sec:MAC_commonmorality}.
\begin{table*}[!tb]
    \centering
    \includegraphics[width=1\linewidth]{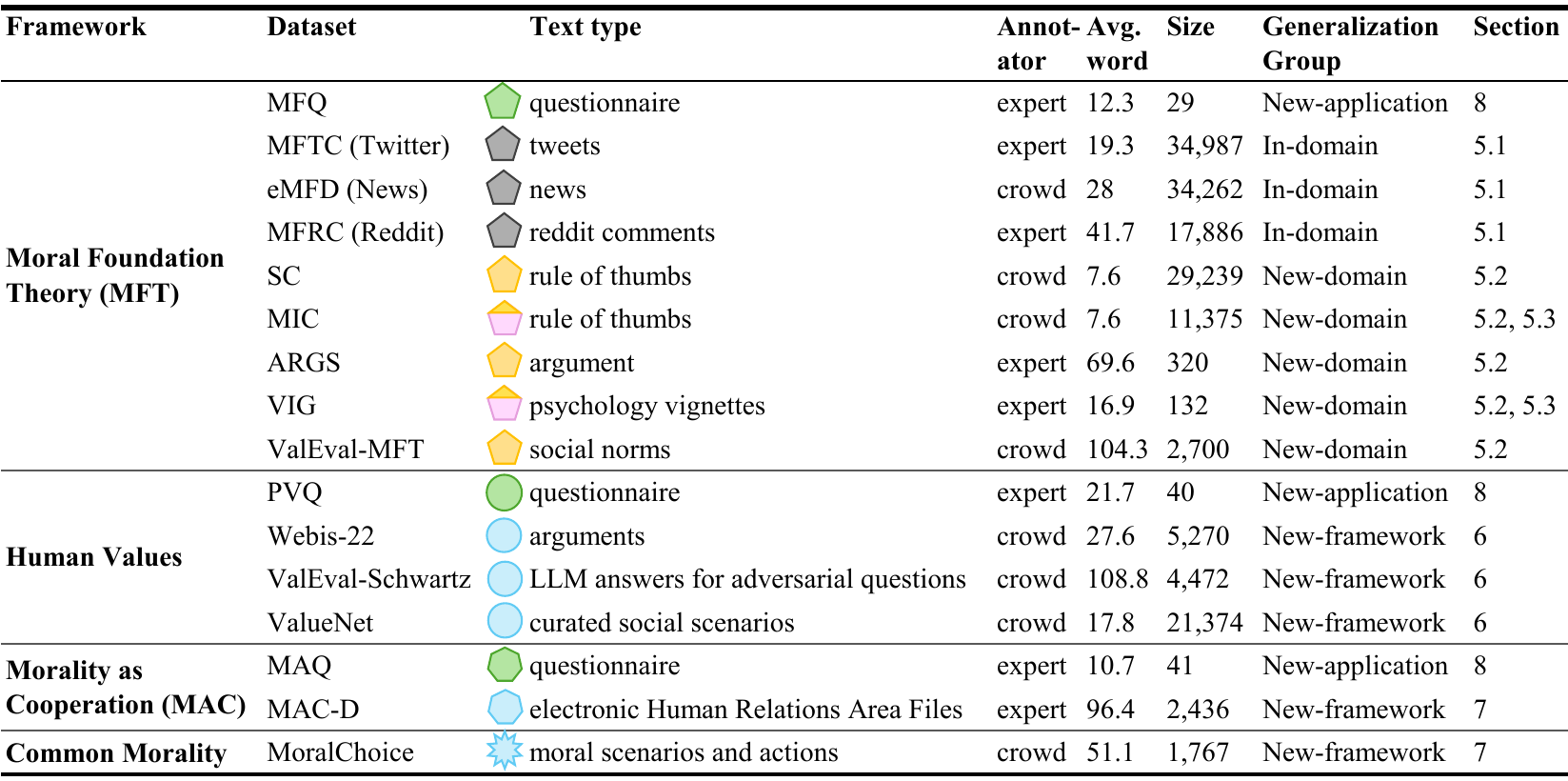}
    \caption{Frameworks and Datasets.
    In the  `annotator' column, `crowd' refers to crowdworkers. The colors and shapes for each dataset follow the same mapping of generalization types and frameworks in \Cref{fig:teaser1}.}
    \label{tab:dataset_overview}
    \captionmoveup
\end{table*}
\modelname includes four frameworks and 16 datasets, and \Cref{tab:dataset_overview} presents an overview. We choose to include four major frameworks because they are well-grounded in social, cultural, and moral psychology, supported by labeled datasets, and widely adopted in recent NLP and computational social science research. To reduce dataset bias, we choose not to include frameworks derived from LLM-generated text.

\textbf{Moral Foundations Theory (MFT)}~\citep{haidtIntuitiveEthicsHow2004a} posits that moral attitudes arise from five foundational intuitions: \textit{care}, \textit{fairness}, \textit{loyalty}, \textit{authority}, and \textit{sanctity}. \textbf{Human values}~\citep{schwartz1992universals} describes ten universal human values that guide behavior across cultures: \textit{Self-Direction}, \textit{Stimulation}, \textit{Hedonism}, \textit{Achievement}, \textit{Power}, \textit{Security}, \textit{Conformity}, \textit{Tradition}, \textit{Benevolence}, and \textit{Universalism}. \textbf{Morality-as-Cooperation (MAC)}~\citep{curryMoralityCooperationProblemCentred2016} conceptualizes morality as a set of evolved solutions to recurrent cooperation problems in human social life, including seven dimensions: \textit{Family}, \textit{Group}, \textit{Reciprocity}, \textit{Heroism}, \textit{Deference}, \textit{Fairness}, and \textit{Property}. \textbf{Common Morality}~\citep{gert2004common} provides a rule-based account of moral reasoning, identifying ten rules designed to prevent harm (e.g., {\it do not kill, do not deceive, do not break promises}). 

\vspace{-0.15em}
Our work includes 13 public datasets (eight for MFT, three for Human Values, one each for MAC and Common Morality) and three psychometric questionnaires (one each for MFT, MAC, and Human Values). We reformulate five data resources into classification tasks, including MoralChoice, moral vignettes (VIG) and three psychological questionnaires (See details about data sources, annotation, and transformation in \Cref{sec:a_framework_data}). The included datasets vary greatly in text length (from fewer than ten to over one hundred words) and source domain (e.g., social media, news, behavioral surveys, ethnographies), and task format (text classification, LLM alignment, human-subject study). We view scoring moral relevance as a \textit{multilabel} classification task: assigning none, one, or more labels per example, instead of \textit{multiclass}: selecting only one label from the set per example. Because moral and value dimensions often co-occur in related datasets and the real world.

\vspace{-0.15em}
To benchmark results across the three types of generalization tasks, we partition the datasets (and frameworks) into four groups. \textit{In-domain} Group includes MFTC~\citep[Twitter]{hooverMoralFoundationsTwitter2020a}, eMFD~\citep[News]{hoppExtendedMoralFoundations2021} and MFRC~\citep[Reddit]{tragerMoralFoundationsReddit2022}, which are large collections labeled with Moral Foundations. We develop the~\modelname classification method on this group, and compare it against other prompting and fine-tuned models (results in \Cref{sec:mft}). We also refer to them as in-domain datasets. 
\textit{New-domain} Group includes SC~\citep{forbesSocialChemistry1012020}, MIC~\citep{ziemsMoralIntegrityCorpus2022}, ARGs~\citep{kobbeExploringMoralityArgumentation2020}, VIG~\citep{cliffordMoralFoundationsVignettes2015} and ValEval-MFT~\citep{yao2024clave}. They are diverse sets ideal for comparing classification in drastically different texts, but remain focused on MFT. 
 We use this group to evaluate the generalizability of \modelname~and fine-tuned models to unseen and out-of-domain data, and refer to them as new-domain datasets
(results in \Cref{sec:mft}).
\textit{New-framework} Group includes three datasets on Human Values and 1 each for MAC and Common Morality. This group covers moral and value frameworks beyond MFT. We use it to assess how well \modelname generalizes to new frameworks, compared to fine-tuned models trained specifically on each framework.
\textit{New-application} Group includes three psychometric questionnaires since they are widely used in psychology and contain items that can be scored for moral or value dimensions. We use them to show new applications of \modelname in detecting potentially multi-loaded items (\Cref{sec:questionaire}). 
\secmoveup 
\section{\modelname Methods}\label{sec:methods}
\sectextmoveup
\modelname's methods include prompting and fine-tuning LLMs (\Cref{subsec:prompt_strat}) and a classifier chain method that explains the top-performing prompt strategy (\Cref{subsec:chain}).
\secmoveup
\subsection{Prompting and Finetuning LLMs}
\label{subsec:prompt_strat}
\sectextmoveup
We explore a range of prompting strategies, ordered from simplest to most complex:
\begin{itemize}[leftmargin=*,noitemsep,nolistsep]
\item The {\it 1-by-1} prompt instructs LLMs to classify texts into each moral or value dimension using separate prompts. This is used by recent work on MFT and Human values with LLMs \cite{rathje2024gpt, Dehghani2024gpt,yao2024clave}.

\begin{figure}[!tbp]
    \centering
    \includegraphics[width=0.9\linewidth]{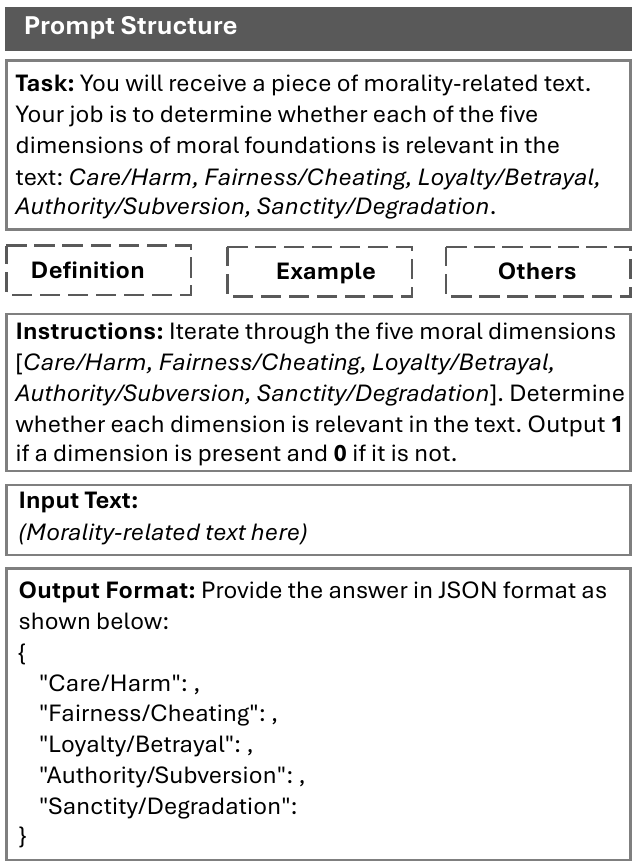}
    \caption{\modelname~prompt structure using MFT as an example. The four main blocks (task, instructions, input, and output) represent the \textit{all@once} prompt; dashed blocks (definition, example, or others) are optional.}
    \label{fig:prompt_structure}
    \captionmoveup
\end{figure}

\afterfigmoveup
\item  {\bf \modelname}, or {\it all@once} prompt, instructs LLMs to classify the input text into all dimensions simultaneously within a single prompt. The intuition for doing so, instead of classifying 1-by-1, is to leverage relationships among moral and value dimensions, which may be captured by LLMs' semantic abilities. This prompt is simple (presented in \Cref{fig:prompt_structure}) and robust across evaluations (see \Cref{sec:mft}, \ref{sec:schwartz}, and \ref{sec:MAC_commonmorality}). \Cref{fig:prompt_structure} shows the prompt structure. Four basic blocks contain task description, instructions, input text and output format (corresponding to the \textit{all@once} prompt). Three additional optional blocks correspond to other strategies as described above. Full prompts can be found in \Cref{appendix:prompts}.
\item The {\it \modelname+~definition} prompt adds definitions of all dimensions to the all@once prompt.

\item The {\it \modelname+~example} prompt adds labeled examples to the prompt. For MFT, examples are selected from three in-domain MFT datasets to cover diverse input text, include at least one positive example and one negative example for each dimension, and have examples with multiple relevant dimensions.

\item The {\it \modelname+~reason} prompt asks the LLM to provide reasoning before outputting the labels in the output block, this prompt is used after including the definition and example blocks to the all@once prompt.

\item The {\it \modelname+~lexicon} prompt combines lexicons with LLMs to test whether leveraging lexicons curated by linguists and experts can improve classification. We propose three methods for selecting and weighting words, as described in \Cref{subsec:lexLLM}.
\sectextmoveup
\end{itemize}

\paragraph{Finetuning LLMs.} We fine-tune GPT-4o-mini 
via OpenAI's API using cross-entropy loss on the three MFT datasets in Group I,  with 50, 300, 3k, and 30k examples, and an 80:20 train–validation split (see details in \Cref{appendix:finetuning}). We evaluate the resulting models, with the \modelname prompt, on the \textit{In-domain} Group test sets and the \textit{New-domain} Group. \Cref{sec:cost_table} reports the costs of fine-tuning and querying different LLMs.

\partextmoveup
\paragraph{LLMs.} We evaluate a set of commercial and open-weight models. Whenever applicable, we report performances on {GPT-4o-mini} and {DeepSeek-V3}. The former is chosen due to its adoption in similar recent work
\citep{demszky2023using,rathje2024gpt}, the latter is open-weight and has competitive performance in public benchmarks~\cite{liu2024deepseek}.
We also experiment with other LLMs, including GPT-3.5-turbo, LLaMA-3-8B, DeepSeek-LLaMA-8B and LLaMA-3-70B (see \Cref{subsec:other_LLMs}). We focus on GPT-4o-mini and DeepSeek-V3 for their strong performance and cost-efficiency via official APIs, with DeepSeek-V3 as an open-weight model.

\partextmoveup
\paragraph{LLM output probabilities.}
We extract probability scores from LLMs to quantify the model certainty. The scores are used in downstream tasks, such as computing ranking metrics, area under the curve (AUC) (\Cref{sec:mft}), and threshold calibration (\Cref{sec:MAC_commonmorality}). We apply the extraction method to all LLMs except DeepSeek-V3, as its official API does not fully support log-probability outputs required for this analysis. The extraction method mainly re-normalises the sum of probabilities of token 0 and token 1 whenever they are found in the top K output tokens. Details are in \Cref{{appenddix:extract_prob_antitoken}}.

\secmoveup
\subsection{Label Correlation and Classifier Chain}
\label{subsec:chain}
\sectextmoveup
When designing the \textit{all@once} prompt, we hypothesize that LLMs can improve classification by leveraging inter-label dependencies. Moral and value dimensions co-occur significantly in the MFT, Human Values, and MAC datasets (see \Cref{appendix:label_corr}). This observation motivates the use of an established machine learning method, \textit{classifier chain} ~\citep{read2011classifier}. It captures dependencies among labels in multilabel classification tasks that are ignored by independent binary classifiers.

In a \textit{classifier chain}, binary classifiers are arranged in sequence, with each classifier predicting a label based on the original input and the predictions of earlier labels. Formally, for an input \( \mathbf{x} \in \mathbb{R}^d \) and a label set \( \{y_1, y_2, \ldots, y_L\} \), the prediction for each label \( y_\ell \) is
\[
\hat{y}_\ell = h_\ell(\mathbf{x}, \hat{y}_{1:\ell-1}),
\]
where \( h_\ell \) is the classifier for label \( y_\ell \), and \( \hat{y}_{1:\ell-1} \) denotes the predictions of earlier labels.

We adapt the \textit{classifier chain} idea to LLM prompting: for the MFT classification, we continue using \textit{1-by-1} prompting, but for each target label, we combine the text input with the other four dimensions predicted by \modelname prompt. For example, when predicting \textit{fairness}, the input includes the original text along with predictions for \textit{care}, \textit{authority}, \textit{loyalty}, and \textit{sanctity} (see \Cref{appendix:order} for stable results of label order permutations). 

\secmoveup
\section{Evaluations on MFT}
\label{sec:mft}
\sectextmoveup
\begin{figure}[!t]
    \centering
    \includegraphics[width=\linewidth]{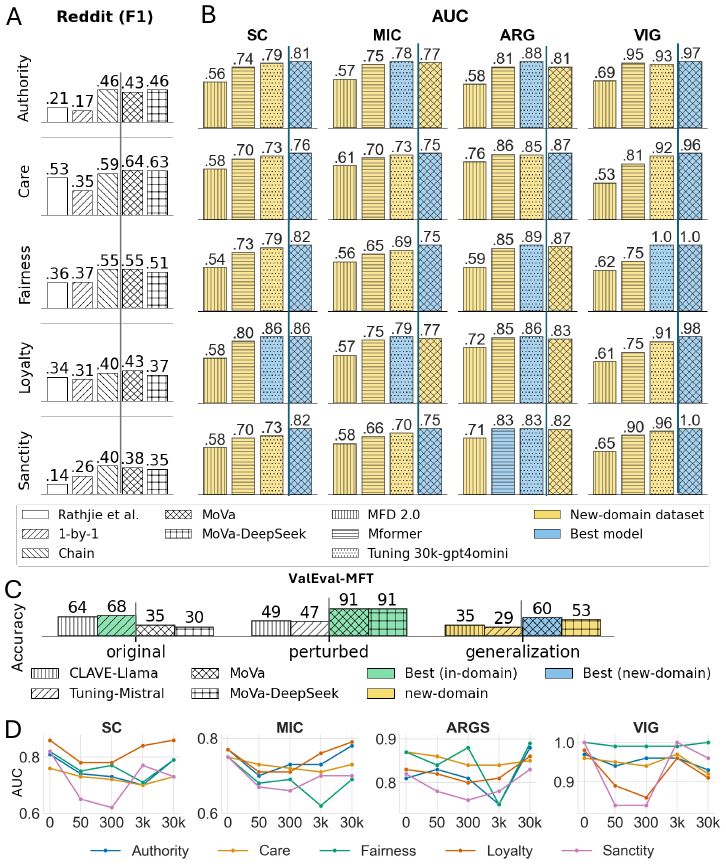}
    \caption{
    (A) F1 scores the entire Reddit dataset by five methods.  (B) AUC scores on four new data domains for four methods of different types: dictionary-based, finetuned-transformer, finetuned LLM and LLM prompting. (C) Accuracy scores across  ValEval-MFT's original, perturbed, and generalization set. (D) AUC scores of \modelname (x=0) and GPT-4o-mini fine-tuned on 50, 300, 3k, and 30k examples from Mformer’s training set on four new data domains. Points lying above indicate improvements.}
    \label{fig:eval}
\end{figure}
\afterfigmoveup

For Moral Foundation Theory (MFT), \Cref{sec:in_domain} presents the results of \modelname prompting strategies in the \textit{In-domain} Group. \Cref{sec:new_domains} then benchmarks generalization abilities of the \modelname prompt in five new data domains in the \textit{New-domain} Group against baselines of our own and recent work. \Cref{ssec:liberty} examines two datasets in the \textit{New-domain} Group that has a \textit{new moral dimension}, {\it liberty}.

\secmoveup
\subsection{In-Domain Evaluations}
\label{sec:in_domain}
\sectextmoveup
We develop the \modelname classification method on the \textit{In-domain} Group, which includes large Twitter, News, and Reddit datasets with high-quality MFT labels. We select our best-performing prompting strategies as the \modelname prompt and compare it against the prompting methods by~\citet{rathje2024gpt} for the {\it in-domain} evaluation. Detailed comparisons and statistical analyses of different prompting strategies and LLMs are in \Cref{sec:Results}. \emph{All@once} (\modelname) prompt significantly outperforms separate binary classification, \emph{1-by-1}; \textit{\modelname}+\textit{example} also significantly outperforms \textit{1-by-1} (see \Cref{fig:comparison_figures}). However, extensions of \textit{all@once}, including \textit{definition}, \textit{example},  \textit{reason} and \textit{lexicon}, do not yield significant improvements. The relatively lower performance of \textit{\modelname}+\textit{reason} reflects that incorporating explicit reasoning may not improve classification tasks in moral and values contexts \citep{Sprague2024ToCO}.

\begin{table*}[!htbp]
    \centering
    \includegraphics[width=1\linewidth]{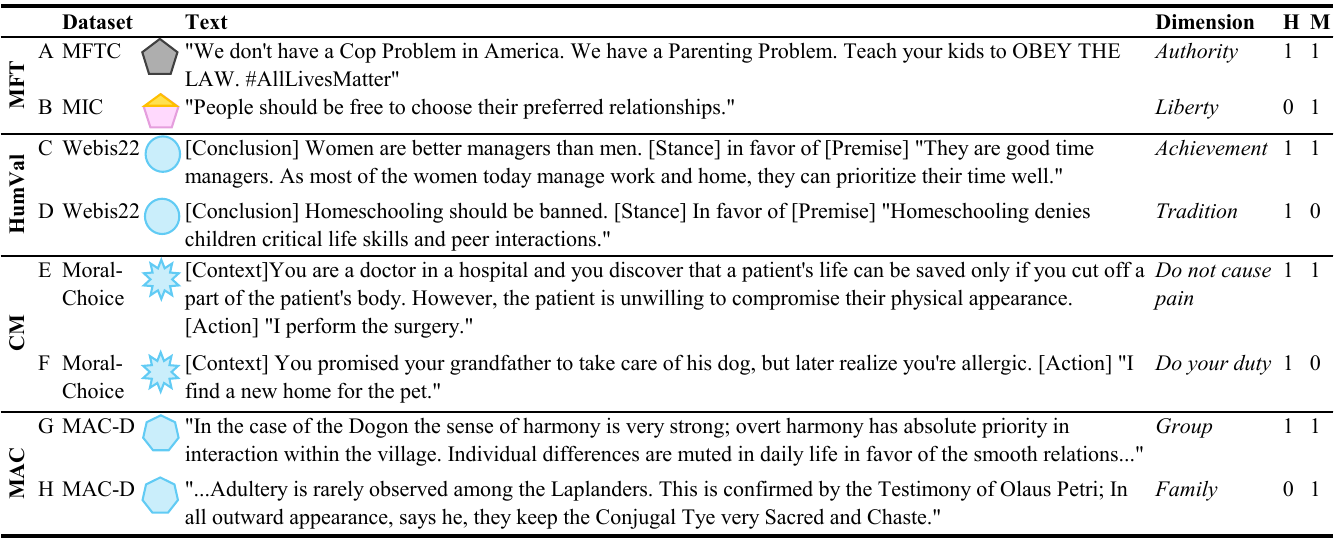}
    \caption{Qualitative examples across four moral and value frameworks, Moral Foundations Theory (MFT), Human Values (HumVal), Common Morality (CM), and Morality-as-Cooperation (MAC), where human annotations (H) and \modelname predictions (M) agree or differ. 1 indicates the dimension is relevant, and 0 indicates it is not. }
    \label{tab:qualitative_analysis}
    \captionmoveup
\end{table*}

\vspace{-0.05em}
\Cref{fig:eval}A summarises the performance of \modelname variants and \textit{classifier chain} on the Reddit dataset, against \textit{Rathjie et al.} bar which shows results from \citet{rathje2024gpt}, using a prompt similar to our {\it 1-by-1} strategy on GPT-4. {\it 1-by-1} is our replication using their prompt on GPT-4o-mini. \modelname~prompt and \modelname-DeepSeek use the \textit{all@once} prompt with GPT-4o-mini and DeepSeek-V3, respectively. Both \modelname variants outperform the two {\it 1-by-1} methods, with F1 gains ranging from 0.03 to 0.25 across foundations. In particular, the performance doubled in F1 for Authority (from 0.21 to 0.43) and also tripled in Sancity (from 0.14 to 0.38).

Regarding why \textit{all@once} outperforms \textit{1-by-1}, the \textit{classifier chain} approach in \Cref{fig:eval}A shows improvements over the \textit{1-by-1} by 0.09 to 0.29, comparable to \modelname prompt in F1. This pattern suggests that LLMs benefit from inter-label dependencies, where earlier predictions help provide a more complete moral context for later ones.

Within the in-domain evaluation, Mformer, our prior work, remains the state-of-the-art finetuned model among others with a learning component. It also outperforms \modelname~without finetuning (see \Cref{appendix:mformer_mova}). However, the key to generalization lies in extending to new data domains, dimensions, and frameworks.

\secmoveup
\subsection{Five New Data Domains}
\label{sec:new_domains}
\sectextmoveup
\Cref{fig:eval}B reports \textit{new-domain} evaluation results, AUC scores for four approaches: the MFD 2.0 lexicon; Mformer, a set of five RoBERTa-base binary classifiers (one per moral foundation), fine-tuned on the \textit{In-domain} Group \citep{nguyen2024measuring}; Tuning 30k-GPT4o-mini, fine-tuned on 30k Mformer training examples; and our prompting-based \modelname. On all external datasets except ARG, \modelname~yields significantly higher AUC scores than Mformer (\Cref{sec:Results_Wilcoxon}). On SC, MIC and VIG, \modelname~prompt outperforms Mformer on all five dimensions. On ARG, \modelname prompt's AUC scores on \textit{loyalty} and \textit{sanctity}  (0.83, 0.82)  are slightly lower than Mformer's (0.85, 0.83), likely due to its longer text (the average number of tokens per example is 69.6). Tuning 30k-GPT4o-mini performs better than Mformer on most datasets but still lags behind the non-finetuned \modelname~in all but the ARG dataset. This suggests that while fine-tuned LLMs can generalize better than smaller models like RoBERTa, it does not consistently surpass well-crafted prompting strategies, and even hurts performance in out-of-domain settings (see error analysis in \Cref{fig:align_examples}).

\Cref{fig:eval}C reports accuracy scores for CLAVE-LLaMA, Tuning-Mistral, \modelname prompt, and \modelname-DeepSeek on the ValEval-MFT dataset~\citep{yao2024clave}. We report this dataset separately because it differs from the other four new-domain datasets covered by Mformer. It includes in-domain test sets in both original and perturbed versions (the latter modifies the original with varying value expressions), as well as an out-of-domain (generalization) set. CLAVE-LLaMA and Tuning-Mistral are both fine-tuned on the original set: the former combines large-model prompting with small-model fine-tuning, while the latter is an open-sourced Mistral-7B. \modelname prompt and \modelname-DeepSeek perform best (0.91) in the perturbed set, with \modelname prompt achieving the highest out-of-domain accuracy (0.60).

\partextmoveup
\paragraph{Fine-tuning vs prompting LLMs.}
To examine how training size impacts \textit{new-domain} generalization, we fine-tune GPT-4o-mini on 50, 300, 3k, and 30k examples from Mformer's training set. \Cref{fig:eval}D shows the AUC performance across the five moral dimensions on four out-of-domain datasets. The point at 0 on the x-axis represents our prompting-based \modelname~without fine-tuning. We observe that fine-tuning on small datasets (50 and 300) often hurts generalizability compared to prompting alone, especially for \textit{sanctity} and \textit{loyalty}. At 3k examples, we see gains in \textit{care}, \textit{authority}, and \textit{fairness}, but drops in \textit{loyalty} and \textit{sanctity}. With 30k examples, fine-tuning improves performance across most dimensions and often outperforms Mformer (17 out of 20 in \Cref{fig:eval}B), yet it still fails to consistently surpass prompting-based \modelname, which remains best in 12 of 20 cases.

\secmoveup
\subsection{New Moral Foundation - \textit{Liberty}}
\label{ssec:liberty}
\sectextmoveup
Recent theoretical developments have spurred a call for \textit{liberty/oppression}, a principle that emphasizes protecting individual freedom against coercion and domination~\citep{haidtRighteousMindWhy2012}. This new Moral foundation is especially resonant in contemporary debates such as vaccination policies or AI and individual autonomy. Two of our new-domain evaluation datasets, MIC (test set) and VIG, contain labeled examples of \textit{liberty}.
\begin{figure}[htbp] 
\raggedright
\includegraphics[width=0.9\linewidth]{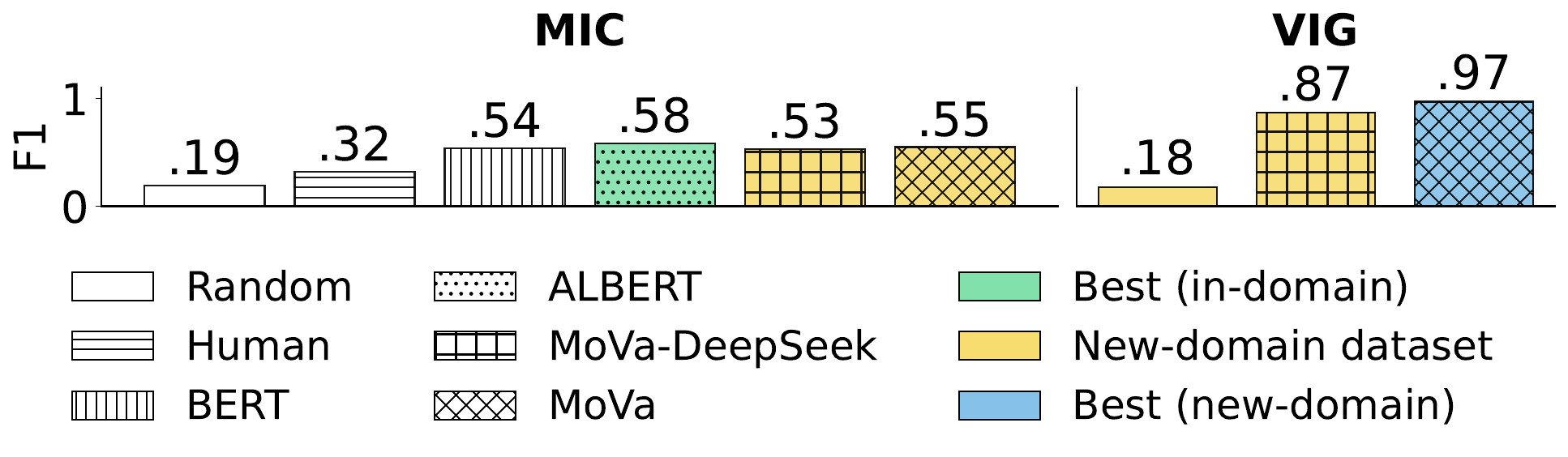} 
\caption{F1 scores on the {\it new dimension} \textit{liberty} in MFT on MIC (left) and VIG (right) datasets.}
\label{fig:liberty_comparison}  
\end{figure}
\afterfigmoveup

\Cref{fig:liberty_comparison} presents F1 scores for \textit{liberty} classification on the MIC (left) and VIG (right) datasets, showing the ability of LLMs to detect new moral dimensions. For MIC, the scores for human annotations, BERT, and ALBERT (fine-tuned on MIC) are taken from \citet{ziemsMoralIntegrityCorpus2022}. ALBERT is a lighter version of BERT that uses parameter sharing and factorized embeddings to reduce model size while maintaining performance. \modelname prompt and \modelname-DeepSeek surpass both the random baseline and human annotators, performing comparably to BERT and ALBERT (see random baseline details in \Cref{sec:random_baseline}). On VIG, \modelname prompt and \modelname-DeepSeek achieve high F1 scores of 0.97 and 0.87, respectively, with \modelname prompt making only one error.

\partextmoveup
\paragraph{Qualitative analysis.}\Cref{tab:qualitative_analysis} shows Examples A–H across the four frameworks, showing \modelname prompt's alignment with human annotators alongside occasional disagreements. For MFT, Example~A, emphasizing obeying the law, is labeled by both annotators and \ modelname as \textit{authority}. In Example~B, concerning personal freedom in choosing a relationship, MoVa labels it as \textit{liberty} while annotators do not. This suggests the potential label noise and the subjective nature of the task.

\section{Evaluations on Human Values}
\label{sec:schwartz}
\sectextmoveup
The Human Values framework by \citet{schwartz1992universals}, rooted in social psychology, has been widely applied in political science~\citep{piurko2011SchwartzPolitics,schwartz2010basic} and more recently in NLP to measure attitudes and behaviors of humans and LLMs. This section evaluates \modelname on the \textit{New-framework} Group of Human Values. For \modelname prompt, all datasets are treated as \textit{new-domain}. For the fine-tuned models, we include BERT (Webis-22~\citep{kiesel2022valueeval}), CLAVE-LLaMA and Tuning-Mistral (ValEval-Schwartz~\citep{yao2024clave}), and BERT (ValueNet~\citep{lu2022valuenet}), following their original evaluation to split datasets into \textit{in-domain} and \textit{new-domain}.

\begin{figure}[!tbp]
    \centering
    \includegraphics[width=1\linewidth]{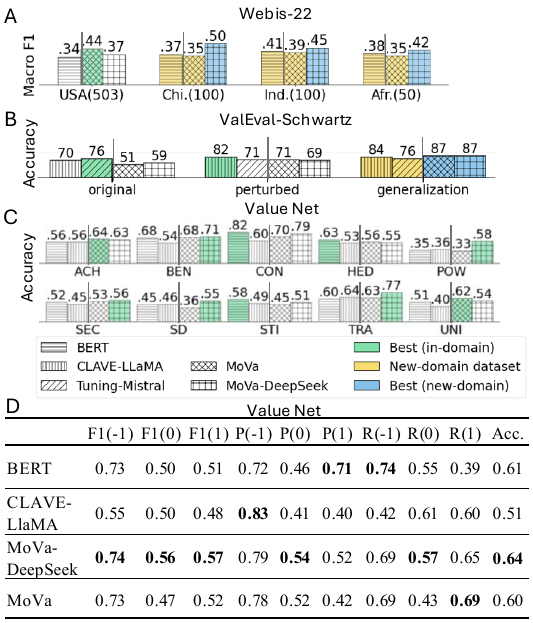}
 \caption{Evaluation of Human Value Classification. (A) Macro F1 scores on Webis-22. Best in-domain model, best out-domain models and other out-of-domain models are in green, blue and yellow, respectively, as in the following subfigures. (B) Accuracy on the ValEval-Schwartz sets (original, perturbed, generalization). (C) Accuracy per dimension from the ValueNet original set. (D) Precision, Recall, F1, and Accuracy for each label class (\(-1\): oppose, \(0\): unrelated, \(1\): support) on ValueNet's in-domain test set. }\label{fig:values}
\end{figure}

Figure~\ref{fig:values}A shows macro F1 scores for BERT, \modelname prompt, and \modelname-DeepSeek across four country-specific subsets from the Webis-22 dataset. The BERT baseline (fine-tuned on the U.S. set) is from~\citet{kiesel2022valueeval}. In the in-domain setting, \modelname prompt achieves the highest F1 score on the U.S. subset (503 samples) with 0.44, followed by \modelname-DeepSeek; both outperform BERT. On the out-of-domain sets, China (100), India (100), and Africa (50), \modelname-DeepSeek performs best.

Figure~\ref{fig:values}B reports accuracy on the ValEval-Schwartz sets across in-domain (original and perturbed) and out-of-domain (generalization) settings. Although CLAVE-LLaMA and Tuning-Mistral achieve the best performance in their in-domain setting, both \modelname-DeepSeek and MoVa achieve the highest accuracy, 0.87, in the out-of-domain setting.

Figure~\ref{fig:values}C shows the accuracy of four models across dimensions from the ValueNet in-domain test set~\citep{lu2022valuenet}. We compare BERT, the fine-tuned baseline reported by \citet{lu2022valuenet}; CLAVE-LLaMA, the fine-tuned model obtained from \citet{yao2024clave}; and \modelname prompt and \modelname-DeepSeek. \modelname-DeepSeek achieves the best performance on 5 out of 10 dimensions, \modelname prompt leads on 2, and BERT on 3. Figure~\ref{fig:values}D further breaks down performance by label class (\(-1\): oppose, \(0\): unrelated, \(1\): support; see label scheme details in \Cref{sec:a_framework_data}).
\modelname-DeepSeek achieves the highest overall accuracy (0.64) and the best F1 scores for all classes.

\partextmoveup
\paragraph{Qualitative analysis.}In \Cref{tab:qualitative_analysis}, Example C presents a premise on women’s advantage in management roles, emphasizing better time management; both annotators and \modelname prompt label this as \textit{Achievement}. In Example~D, the premise for banning homeschooling highlights the negative effects on life skills and peer interactions. MoVa labels it as irrelevant to \textit{Tradition}, which is plausible, but annotators disagree.

In summary, we are pleasantly surprised that \modelname prompt, developed on MFT, generalizes well to Human Values, with \modelname-DeepSeek consistently achieving the best performance.

\section{Evaluations on Common Morality and Morality-as-Cooperation (MAC)}
\label{sec:MAC_commonmorality}
\sectextmoveup

\begin{figure}[!tbp]
    \centering
    \includegraphics[width=1\linewidth]{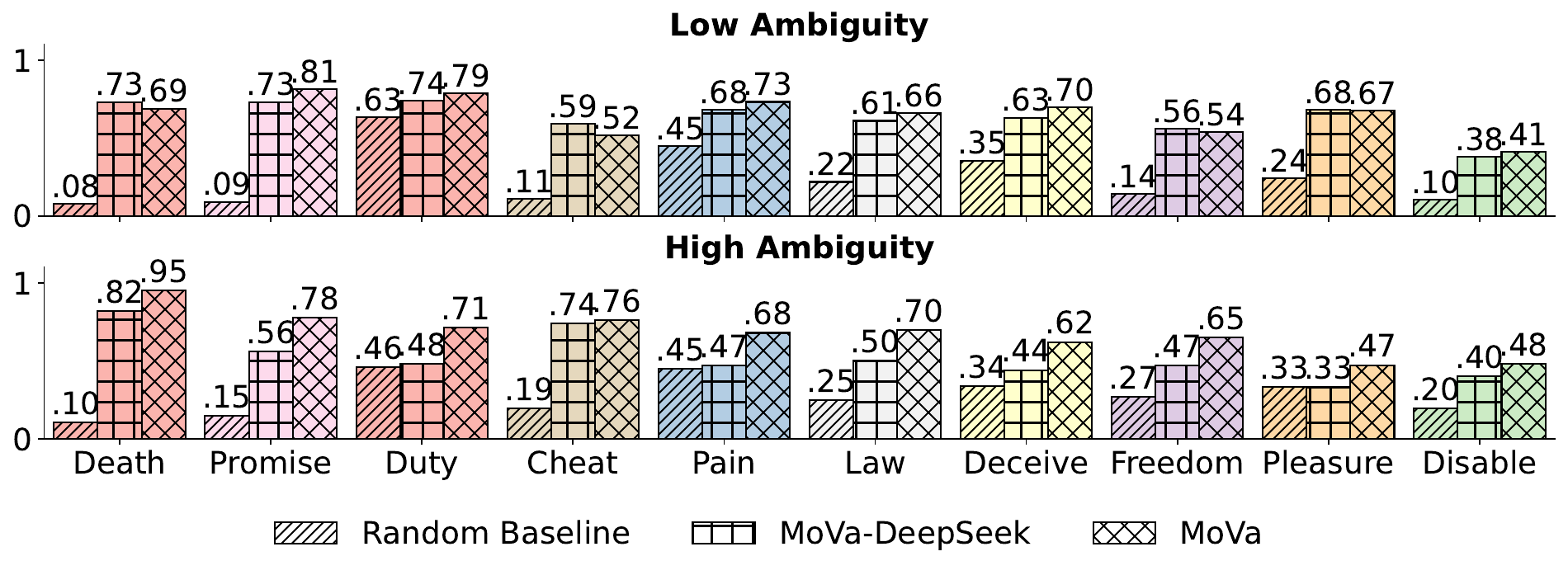}
    \caption{F1 scores on MoralChoice dataset, for \modelname, \modelname-DeepSeek and Random baseline, broken down by low vs high ambiguity scenarios. Colors indicate the 10 rules.}\label{fig:moralchoice}
 \end{figure}

For the Common Morality framework, the original MoralChoice dataset~\citep{ScherrerMoralChoice} was created to evaluate LLMs on action choice and uncertainty, with auxiliary labels for the rules each action violates. We reformulate the task into a \textit{classification task}, where \modelname is prompted to identify which of the ten rules\footnote{The ten rules are: Do not {\it kill}.
Do not cause {\it pain}.
Do not {\it disable}.
Do not deprive of {\it freedom}.
Do not deprive of {\it pleasure}.
Do not {\it deceive}.
Do not break your {\it promises}.
Do not {\it cheat}.
Do not break the {\it law}.
Do your {\it duty}. {\it Italicized} words denote abbreviations used in \Cref{fig:moralchoice}.} apply, based on the \textit{scenario} description and {\it action} text (See \Cref{sec:moralchoice} for the prompt.) \Cref{fig:moralchoice} presents F1 scores for \modelname prompt, \modelname-DeepSeek, and a random baseline across low- and high-ambiguity scenarios. Both LLM-based models outperform the random guess on every dimension, regardless of scenarios. \modelname prompt retains strong, consistent F1 scores regardless of ambiguity level (0.41–0.81 in low and 0.48–0.95 in high), while \modelname-DeepSeek matches its performance in low-ambiguity cases (0.38–0.74) but degrades notably under high ambiguity (0.33–0.82). 
\partextmoveup
\paragraph{Qualitative analysis.}For Common Morality in \Cref{tab:qualitative_analysis}, 1 denotes relevance to a violation and 0 denotes non-violation. Example~E presents a medical case where saving a patient’s life requires amputation against their wishes; both humans and \modelname prompt label it a violation of \textit{Do not cause pain}. Example~F is more nuanced: rehoming due to an allergy fulfills the duty by alternative means. Annotators see a violation of \textit{Do your duty}, whereas \modelname prompt does not.

\afterfigmoveup

\begin{figure}[!tbp]
    \centering
    \includegraphics[width=\linewidth]{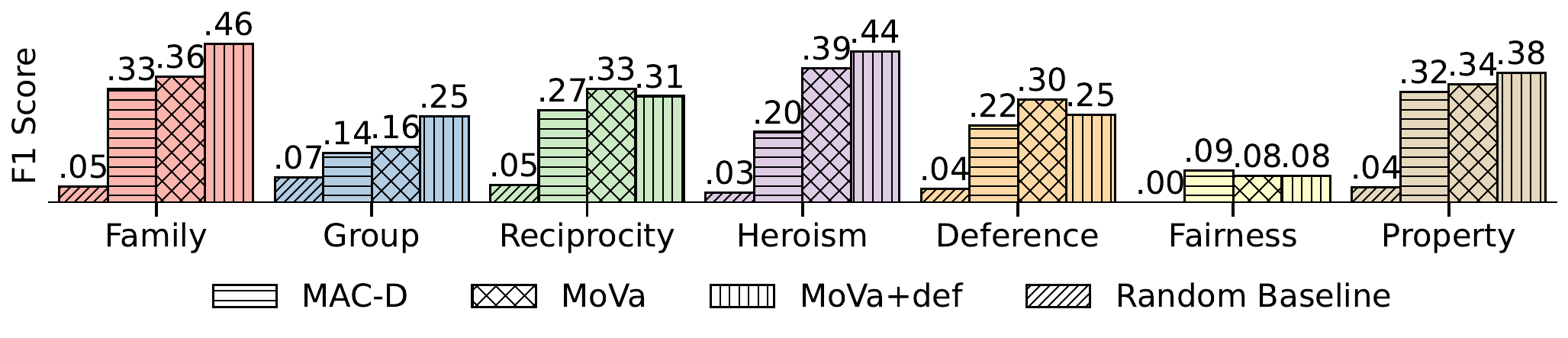}
    \caption{F1 scores for MAC-D, \modelname and Random baseline on MAC-D dataset with threshold calibration.
    }\label{fig:macd_results}
 \end{figure}

For the MAC framework, we prompt \modelname on the MAC-D dataset \citep{MoralUniversals2024} to classify the seven dimensions. Most moral categories appear in fewer than 5\% of the 2,436 examples, with \textit{Fairness} under 1\% (see Table~\ref{tab:mac_label_distribution}). 
To address this imbalance, we calibrate thresholds by setting the 95th percentile of predicted probabilities as the cutoff for all models, aligning predictions with the sparse positive labels. \Cref{fig:macd_results} shows that all three approaches, \modelname prompt, \modelname+~definition prompt, and the MAC-D model (their baseline model using word frequency and logistic regression to classify dimensions), outperform the random baseline. \modelname+~definition prompt achieves the best F1 in 4 out of 7 dimensions, followed by \modelname prompt (2), and MAC-D model (1). This suggests that providing definitions can help improve moral relevance classification. This could be due to a number of reasons: dimensions important for social cooperation (such as reciprocity, deference) are semantically more complex than those in MFT, or MAC-D dataset is less widely used than MFT and hence has less data that was used to pre-train LLMs. 
\partextmoveup
\paragraph{Qualitative analysis.}In \Cref{tab:qualitative_analysis}, Example~G describes harmony within the Dogon community, where social interactions prioritize smooth relations; both humans and \modelname +~definition prompt label this as \textit{group}. In Example~H, the “conjugal tye” case (the bond between a husband and wife) is labeled by \modelname +~definition prompt as \textit{family}, whereas the annotators do not, likely because the phrasing is uncommon.

\section{Evaluating Psychological Surveys}
\label{sec:questionaire}

\begin{figure}[t]
    \centering
    \includegraphics[width=0.9\linewidth]{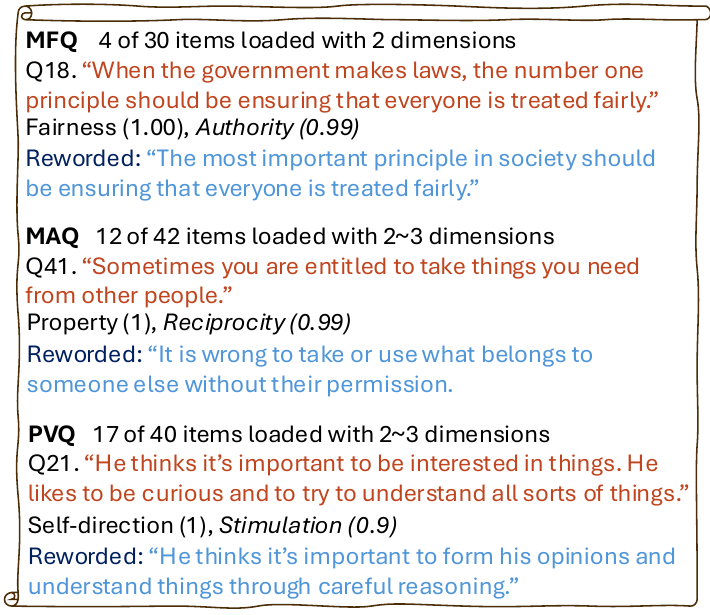}
    \caption{
    Evaluation results and multiply-loaded examples from MFQ, MAQ, and PVQ questionnaires.
    Each box shows an original item, the model’s predicted dimensions with related probability scores, and a reworded item to target a single dimension. \textit{Italicized} label is the additional dimension predicted by~\modelname.}\label{fig:Q_example}
 \end{figure}
\afterfigmoveup
Questionnaires are commonly used tools for measuring subjective qualities in human participants, from personality traits to moral values and more. Recently, computer scientists began using questionnaires to measure morality, values, and personality in LLMs in order to compare and align 
them with human traits \citep{ren-etal-2024-valuebench, pschobench2024, abdulhai2023moral, EvaluatingAIpersonality2022}. In practice, most questionnaire scores are computed by summing (or averaging) the responses to a prespecified set of items linked to a given dimension. Yet, items could still be relevant for more than one subjective dimension despite due diligence in questionnaire design, a phenomenon known as cross-loading~\citep{li2020crossloadings, bolt2022item,dongbo2024mirt}. This will likely confound the correlations among dimensions and affect interpretations.

In this section, we use \modelname prompt to score the {\em relevance} of moral dimensions of each question in the moral foundation questionnaire (MFQ)~\cite{grahamLiberalsConservativesRely2009a}, the Morality-as-cooperation questionnaire (MAQ)~\cite{curry2019mapping} and the Portrait-Values Questionnaire (PVQ)~\cite{schwartz2001}. The three questionnaires, prompts, and full results for each question are in \Cref{sec:questionnaire_results}. \Cref{fig:Q_example} contains an overview of results. Across all three questionnaires, there are no false negatives per dimension (recall = 1.0). However, all questionnaires have multiple items loaded with more than one dimension. PVQ has the highest percentage (17 out of 40) of multi-loaded questions, suggesting that it becomes more difficult to craft distinct items as the number of scoring dimensions increases (10 rather than 5 or 7). We use LLMs to 
help reword some multi-loaded questions for each questionnaire to remove the dimensions other than the prescribed one. Observations in \Cref{fig:Q_example} show that it is possible to preserve the main content of the question and remove the unintended dimensions.

\section{Conclusion}
\label{sec:discussion}
\sectextmoveup

We propose \modelname, a set of resources for generalizable classification of human morals and values in text. \modelname consists of 16 labeled datasets with four theoretically informed frameworks and a set of LLM prompting methods that outperform fine-tuned models in new data domains. Our aim for \modelname~is deliberately set to only scoring relevance rather than making judgments, leaving ambiguous and high-stake decisions to humans.

Future work includes generalizing to multi-lingual and cross-cultural datasets, as well as crowd-sourced values and rules of thumb. Deeper integration into incorporating linguistic resources (such as lexicons) into LLM-based workflow holds promise in both performance and interpretability. Prompting strategies used by \modelname could be adopted or adapted to other subjective text analyses after rigorous evaluations. Finally, design iterations and validations of surveys and psychological questionnaires audited with \modelname would be useful in their own right.

\section*{Limitations} 
Although our study covers diverse domains, it primarily focuses on English-language data and frameworks grounded in Western moral traditions. Further research should incorporate more cross-cultural contexts and frameworks, as moral values can vary significantly across regions and belief systems. 

Because \modelname~relies on large language models, any biases or gaps in the original model may affect overall performance. For instance, LLMs may be disproportionately trained on Western-centric or internet-based text sources, potentially skewing moral relevance scores toward those cultural norms. Additionally, if the underlying models exhibit known issues such as gender, racial, or socioeconomic biases, these biases could surface in \modelname's outputs.

Since text classification is inherently imperfect, we emphasize that aggregate analyses are more reliable than individual-level outputs. The models' output for individual pieces of input text should not be used without further scrutiny and oversight.

``Which moral or value framework?" is a fundamental philosophical and social science question with active ongoing debate. While this work primarily provides a tool for applying given frameworks, we view it as a step toward levelling the playing field among different frameworks. In other words, we believe that part of the reason some frameworks are used more frequently (for instance, in text analysis or by AI) is the availability of resources, such as annotated data and computation. \modelname's performance on new domains, new tasks, and new frameworks suggests that such tools can help many frameworks be more widely applied and compared to each other.
Although establishing the validity of \modelname still requires evaluation data, one could envision this approach as being more economical than curating training data.

Finally, our experiments did not encompass all available models, including open-source alternatives. More efforts are needed to explore how the open-source ones can potentially match or exceed GPT-level performance.

\section*{Ethics statement}

Our approach is intended to assess different aspects of moral relevance as a support tool, rather than to render judgments of vice or virtue. We underscore that moral and value-based decisions, particularly those that are high-stakes or culturally sensitive, should remain entrusted to human agency.

We acknowledge that both the training data and model outputs of LLMs may contain inherent biases, which can in turn influence the behavior of \modelname. These biases may include, but are not limited to, Western-centric conceptions of morality and other systemic patterns such as historical inequalities. Such limitations should be considered when interpreting results, and we caution against assuming that outputs are universally representative or free of normative assumptions.

In addition, users may inadvertently or deliberately apply the tool beyond its intended scope, for example by attempting to judge individuals on the basis of their writing or comments. We stress that any moral or normative assessment requires careful, context-sensitive human oversight and should never rely solely on automated scoring.

\section*{Acknowledgement}

This work is supported in part by Australian Research Council under project FT230100563 and DP240100506, and CSIRO--National Science Foundation AI Research Collaboration Program (NSF IIS-2302785).

The authors' contributions are listed below.\\
\textbf{Ziyu Chen:} Conceptual framing, experiment design, dataset sourcing, conducting experiments, result analysis, and writing. \\
\textbf{Junfei Sun:} Research ideas, Dataset sourcing, conducting experiments, result analysis, and writing. \\
\textbf{Chenxi Li:} Conducting experiments, result analysis, and writing. \\
\textbf{Tuan Dung Nguyen:} Dataset sourcing, result analysis, and writing. \\
\textbf{Jing Yao:} Advising on data selection and experiments about one baseline. \\
\textbf{Xiaoyuan Yi:} Advising. \\
\textbf{Xing Xie:} Advising. \\
\textbf{Chenhao Tan:} Conceptual framing, advising, and writing. \\
\textbf{Lexing Xie:} Conceptual framing, experiment design, advising, and writing.

\clearpage
\appendix
\addcontentsline{toc}{chapter}{Appendices}
\newpage
\localtableofcontents

\setcounter{figure}{0}
\setcounter{table}{0}

\renewcommand\thesection{\Alph{section}}
\renewcommand\thesubsection{\thesection.\Roman{subsection}}
\renewcommand\thesubsubsection{\thesection.\Roman{subsection}.\arabic{subsubsection}}

\renewcommand{\thetable}{\Roman{table}}
\renewcommand{\thefigure}{\Roman{figure}}

\newcommand{\tocrow}[2]{
    \noindent $\hookrightarrow$ \Cref{#1}: #2 \hrulefill Pg \pageref{#1}\\
}

\makeatletter
\newcommand\setcurrentname[1]{\def\@currentlabelname{#1}}
\makeatother

\makeatletter

\newpage

\section{Related work - extended discussions}
\label{appendix:relatedwork}

\subsection{Different Moral and Value Frameworks}

Moral and value frameworks in NLP differ in origin and purpose. Broadly, they fall into two categories: crowd-sourced frameworks and theory-driven frameworks from moral philosophy and psychology.

The first category includes frameworks based on user input or red teaming efforts. For example, the \textit{Social Risks} framework~\citep{ji2023socialrisks} gathers perceptions of social harms from diverse communities. The \textit{ALERT benchmark}~\citep{tedeschi2024alertSafety} contains over 45,000 instructions organized by a fine-grained risk taxonomy to evaluate LLM safety. While useful for surfacing real-world concerns, these frameworks often reflect model-specific or emergent behaviors.

The second category is grounded in well-established moral theories developed through decades of research. These frameworks identify key dimensions of moral reasoning across cultures and contexts, including politics, law, and everyday decision-making. Moral Foundations Theory (MFT)~\citep{haidtIntuitiveEthicsHow2004a} is based on psychological and anthropological work and proposes five innate foundations, such as care/harm and fairness/cheating. Human Values Theory~\citep{schwartz1992universals} is derived from large-scale cross-cultural surveys and identifies ten motivational value types. Morality-as-Cooperation (MAC)~\citep{curryMoralityCooperationProblemCentred2016} uses an evolutionary lens to explain how moral rules help solve social challenges. Common Morality~\citep{gert2004common} offers a set of ten harm-avoidance rules (e.g., \textit{do not kill}, \textit{do not deceive}) grounded in bioethical reasoning.

We focus on these theoretical frameworks because they are widely validated, applied in fields such as political science, cross-cultural study and bioethics~\citep{aminAssociationMoralValues2017,piurko2011SchwartzPolitics,curry2019mapping,gert2006bioethics}, and have been effectively used in NLP via large-scale annotation. To avoid model-induced bias, we exclude frameworks derived from LLM-generated content.

\subsection{Evaluating and Aligning LLMs with Human Values}
Large language models (LLMs) can facilitate moral judgments, yet we believe final moral decisions, especially in hard or ambiguous cases, should remain with humans. 

Prior work on aligning LLMs with human values generally follows two main approaches: a bottom-up strategy that learns from crowdsourced moral judgments \citep{jiang2022delphi}, and a top-down strategy that applies established moral theories \citep{zhou-etal-2024-rethinking}. Other studies examine the moral preferences implicitly encoded in LLMs \citep{abdulhai2023moral, ScherrerMoralChoice} or propose alignment methods based on human value models \citep{zheng2024ali, tennant2024moral, yao2024value, duan2024denevil}. In parallel, red teaming has become a critical tool for evaluating and improving model safety under adversarial conditions~\citep{ganguli2022redteaming, tedeschi2024alertSafety}.

Different from this line of work, our goal is to measure human morals and values via LLMs to support research in computational social science.

\subsection{Measuring moral relevance in text}
Central to many studies of individual and collective behavior are human morals and other values that are extensively embedded in written language.  

A typical analysis through the lens of morality involves labeling data according to a chosen framework (e.g., does this Tweet underscore the \emph{authority} moral foundation?) and making appropriate comparisons based on these labels. Traditionally, the former is done manually, either by researchers or by crowd-sourced workers recruited online. Given the large scale of data in today's studies, this is often infeasible and automated data labeling tools are deployed instead. In the context of moral foundations theory, a suite of tools have been proposed, ranging from word count~\citep{grahamMoralFoundationsDictionary2012,frimerMoralFoundationsDictionary2019c,hoppExtendedMoralFoundations2021}, word embedding similarity~\citep{mokhberianMoralFramingIdeological2020} and supervised machine learning~\citep{guoDataFusionFramework2023,nguyen2024measuring}. The morality-as-cooperation framework has also seen similar developments~\citep{MoralUniversals2024}.

While promising, machine learning methods face two principal challenges that have remained pertinent until recently: the lack of high-quality annotated data and limited generalizability. For instance, a model that classifies moral foundations in text cannot directly be used to label dimensions according to morality-as-cooperation--simply because it is not designed to. Large language models, through effective prompting strategies, hold the potential to achieve this generalization as they are general-purpose tools. While this insight is not new, the key to successful use of LLMs lies in how researchers make use of prompting techniques. This is the focus of our paper.

\section{Frameworks and Data}
\label{sec:a_framework_data}
\sectextmoveup

To evaluate the generalizability of our methods, we use datasets that vary in text length, data domains, moral dimensions, and underlying theoretical frameworks, as shown in \Cref{tab:dataset_overview}. We explore four different frameworks for human morals and values with 13 datasets and 3 questionnaires in text analysis and social behaviour study.

\sectextmoveup
\paragraph{(1) \textit{Moral Foundations Theory (MFT) framework}}~\citep{haidtIntuitiveEthicsHow2004a,haidtRighteousMindWhy2012} in psychology attributes variations in moral behavior, attitude and judgment to five categories of intuition, called moral foundations: \textit{authority}, \textit{care}, \textit{fairness}, \textit{loyal} and \textit{sanctity}. 

\textbf{Moral foundation questionnaire (MFQ)}~\cite{grahamLiberalsConservativesRely2009a}, containing 30 questions on five dimensions. 
We use three large-scale datasets specifically annotated for MFT, \textbf{MFTC}, \textbf{MFRC}, and \textbf{eMFD}, used by~\citet{nguyen2024measuring} to train Mformer, to identify the most effective prompting strategies in our setting. And we include the other five MFT datasets from new data domains for evaluation:

\textbf{Twitter (MFTC)}~\citep{hooverMoralFoundationsTwitter2020a} includes 34,987 tweets on topics such as All Lives Matter, Black Lives Matter, the 2016 U.S. Presidential election, hate speech, Hurricane Sandy, and MeToo. 13 trained annotators provided at least 3 labels per tweet, with at least 50\% agreement on the presence of a moral dimension per example. All annotators were undergraduate assistants who took part in several training sessions to gain expert-level understanding of the Moral Foundations Taxonomy.

\textbf{Reddit (MFRC)}~\citep{tragerMoralFoundationsReddit2022} contains 17,886 comments from 12 subreddits covering U.S. politics, French politics, and everyday moral life, labelled by 27 trained annotators, with at least 50\% agreement on the presence of a moral dimension per example. They began with 27 annotators, all undergraduate research assistants, who joined two months of training to become familiar with Moral Foundations Theory (MFT). The training included lectures, group discussions, reading materials, and practice annotations, along with checks for how well annotators agreed with each other.

\textbf{News (eMFD)}~\citep{hoppExtendedMoralFoundations2021} is collected for the eMFD lexicon, comprising 34,362 examples sourced from 1,010 news articles and 73,001 highlights annotated by 854 annotators.

\textbf{Moral Vignettes (VIG)}~\citep{cliffordMoralFoundationsVignettes2015} is designed by experts to target a single moral foundation per example, with about 30 annotators per vignette and at least 60\% agreement. 

\textbf{Moral Arguments (ARG)}~\citep{kobbeExploringMoralityArgumentation2020} contains diverse arguments with high-quality labels from two expert annotators.

\textbf{Social Chemistry 101 (SC)}~\citep{forbesSocialChemistry1012020} comprises 292K rules-of-thumb (RoTs) representing cultural norms, annotated by crowdsourced workers.

\textbf{Moral Integrity Corpus (MIC)}~\citep{ziemsMoralIntegrityCorpus2022} features 99K RoTs created from prompts, with labels on violation severity, consensus, and associated moral foundations by crowdsourced workers. 

\textbf{ValEval-MFT}~\citep{yao2024clave} evaluates five moral foundations using three subsets: 1,000 original LLM-generated paragraphs from DenEvil \citep{duan2024denevil}, 603 perturbed samples with edits based on the original, and 406 generalization samples from a new data domain, Moral Stories\citep{emelin2021moralstories}. All samples are labelled as adhering to, opposing, or unrelated.

\sectextmoveup
\paragraph{(2) \textit{Human values framework}}~\citep{schwartz1992universals} describes universal human values as guiding principles in life that motivate human behaviors across cultures, including ten core values: \textit{Self-Direction, Stimulation, Hedonism, Achievement, Power, Security, Conformity, Tradition, Benevolence, and Universalism}.

\textbf{Portrait Values Questionnaire (PVQ)} is a psychometric instrument developed by~\cite{schwartz2001} to assess the ten basic human values and is widely used in cross-cultural research

\textbf{ValEval-Schwartz}~\citep{yao2024clave} covers ten Schwartz values using 1,000 original samples from Value Fulcra, 800 perturbed samples generated by Mistral-Large and filtered by humans, and 400 generalization samples adapted from a new data domain, Do-not-Answer benchmark. Each instance is labeled as \textit{not related to}, \textit{adhere to}, or \textit{oppose to}.

\textbf{Webis-22}~\citep{kiesel2022valueeval} (Webis-ArgValues-22) includes 5,270 arguments from four cultural regions. It supports multiple taxonomies (54, 20, 4, and 2-value sets) and provides at least three annotations per argument, with a Krippendorff’s alpha of 0.49.

\textbf{ValueNet}~\citep{lu2022valuenet} contains 21,374 text scenarios, each labelled with one of ten values, instead of multilabels. Each scenario was annotated by four qualified workers (443 total), with an inter-annotator agreement of 64.9\% and a Fleiss’ kappa of 0.48. The labels capture whether the text is \textit{unrelated} (0) to a value, \textit{supports} it (1), or \textit{opposes} it (–1).

\sectextmoveup
\paragraph{(3) \textit{Morality-as-Cooperation (MAC) framework}}~\citep{curryMoralityCooperationProblemCentred2016, curry2019mapping} views morality as a set of solutions to cooperation challenges, rooted in humanity’s long history of living in groups. It identifies seven core moral domains: \textit{Family}, \textit{Group}, \textit{Reciprocity}, \textit{Heroism}, \textit{Deference}, \textit{Fairness}, and \textit{Property}.

\textbf{MAC Questionnaire} is a 42-item instrument to assess the seven moral domains in the MAC framework and is used in cross-cultural moral research.

The \textbf{MAC-D dataset}, provided by \citep{MoralUniversals2024}, consists of 2,436 ethnographic paragraphs drawn from the Probability Sample Files (PSF) within the electronic Human Relations Area Files (eHRAF), covering 60 culturally diverse societies. Each paragraph was manually annotated by three coders, two of whom are authors of the study, according to the seven moral dimensions defined in the MAC framework. The annotations indicate whether a paragraph is morally relevant to each dimension, regardless of sentiment. For example, a paragraph praising someone for sharing food with their group would be labeled as a morally positive instance of the \textit{group} dimension, while one criticizing someone for betraying or abandoning their group would be labeled as a morally negative instance of the same dimension. Both are considered morally relevant in \textit{group}.

\sectextmoveup
\paragraph{(4) \textit{Common Morality framework}}\citep{gert2004common} comprises ten rules designed to avoid harmful actions (e.g., killing, deceiving, or breaking promises), with a two-step process that first identifies relevant rules in a moral scenario, then assesses the likely social consequences of violating them. 

\textbf{Moral Choice dataset}~\citep{ScherrerMoralChoice} contains 1,767 scenarios, split roughly 50-50 between being low- and high-moral ambiguity. Each scenario has a brief description along with two actions, where {\it action1} is crafted towards upholding moral rules, and {\it action2} is against. Three annotators label whether each action violates each of the ten moral rules, with overall agreement (at least two out of three) reaching 99.32\%. We include the scenario description with \textit{action2} text. Formally, we exclude \emph{action1} because in low‐ambiguity scenarios all ground truth labels for \emph{action1} are negative and in high‐ambiguity scenarios the majority are negative. This means there are no true positives, false positives, or false negatives. Precision and recall are defined as the ratios of true positives to the sums of true positives with false positives and false negatives, respectively, which makes both metrics undefined when there are no positive examples. The \(\mathrm{F}_1\) score is the harmonic mean of precision and recall.

\paragraph{Tokenization and Average Lengths.} For the average number of tokens (Avg. words) reported in \Cref{tab:dataset_overview}, we use the Tiktoken tokenizer, a Python library developed by OpenAI to match the tokenization used in models like GPT-3.5, GPT-4, and GPT-4o(including GPT-4o-mini). We tokenize each example in the dataset and report the average token count per dataset.

\section{Methods}\label{sec:a_methods}
\subsection{Hyperparameter Settings}
We present the default settings of each hyperparameter used in deploying both the GPT-4o-mini and LLaMA-3.1 models.

\subsubsection*{GPT-4o-mini Hyperparameters}
\begin{itemize}
    \item \textbf{Max Tokens:} The max tokens limit is set to 4096 to balance detailed response generation with efficient use of computational resources, making it suitable for various applications that require extensive textual output.

    \item \textbf{Temperature:} The temperature is fixed at 0.0 to ensure deterministic outputs, where the model provides the most likely response consistently. This setting is crucial for applications demanding high precision and predictability.

    \item \textbf{Top-p:} Top-p is also maintained at 0.0 to support deterministic outputs, limiting variability and enhancing the accuracy and reliability of the model's responses for critical tasks.

    \item \textbf{Logprobs:} Logprobs are enabled by default to include log probabilities with the outputs, offering insights into the model’s decision-making process. This feature is particularly useful for debugging and in-depth analysis.

    \item \textbf{Top Logprobs:} The top logprobs parameter is set to return the 20 highest log probabilities to provide a comprehensive overview of the model’s probabilistic reasoning, aiding further research and fine-tuning.
\end{itemize}

\subsubsection*{DeepSeek-V3 Hyperparameters}
\begin{itemize}
\item \textbf{Temperature:} The default temperature is 0.0, the same as GPT-4o-mini.  

\item \textbf{Top-p (Nucleus Sampling):} The top-p value is \(1 \mathrm{e}{-10}\), approaching the setting of GPT-4o-mini while 0.0 is not supported in DeepSeek.  

\item \textbf{Logprobs:} Log probabilities are enabled.  

\item \textbf{Top Logprobs:}  The top logprobs parameter is set to return the 20 highest log probabilities.

\item \textbf{Frequency Penalty \& Presence Penalty:} Defaults to 0 (no penalty), allowing natural repetition when contextually appropriate.

\item \textbf{Stop Sequences:} No default stop sequences.

\end{itemize}

\subsubsection*{LLaMA-3.1 Hyperparameters}
We also document the default hyperparameter settings for the LLaMA-3.1-70B-Instruct and LLaMA-3.1-8B-Instruct models:

\begin{itemize}
    \item \textbf{Max Tokens:} The maximum number of tokens is set to 512, ensuring controlled text generation while balancing efficiency and computational constraints.

    \item \textbf{Temperature:} The temperature is set to 1.0. If sampling is disabled (\texttt{do\_sample=False}), this value prevents warnings related to stochastic generation.

    \item \textbf{Top-p:} The top-p value is set to 1.0, indicating that nucleus sampling is not in effect when sampling is disabled.

    \item \textbf{Logprobs:} Logprobs are enabled by default, providing log probability data to enhance interpretability and facilitate deeper model analysis.

    \item \textbf{Top Logprobs:} The model is configured to return the top 20 log probabilities, offering a broad perspective on token likelihoods.
\end{itemize}

\subsection{Prompts}\label{appendix:prompts}

We here present different prompt sections that can be added to the baseline prompt to generate different other versions of prompts listed in \Cref{sec:methods}:

\subsubsection{1-by-1 Prompt}
\label{sssec:1by1prompt}
\begin{tcolorbox}[colframe=black, colback=white, boxrule=0.5pt, title=1-by-1 prompt, breakable]

    Does the following text express the moral foundation of:\\
    \{dimension\}\\
    Please answer only with a number: 1 if yes and 0 if no. Here is the text:\\
    \{text\}
\end{tcolorbox}
Insert the desired dimension into \{dimension\} and the text to be analyzed into \{text\}.

\subsubsection{All@once Prompt}
\label{sssec:all@onceprompt}
\begin{tcolorbox}[colframe=black, colback=white, boxrule=0.5pt, title=all@once prompt, breakable]
You will receive a piece of morality-related text. Your job is to determine whether this morality-related text involves the five dimensions of moral foundations: Care/Harm, Fairness/Cheating, Loyalty/Betrayal, Authority/Subversion, and Sanctity/Degradation.

<instructions> \\
Iterate through the five moral dimensions in [Care/Harm, Fairness/Cheating, Loyalty/Betrayal, Authority/Subversion, Sanctity/Degradation]. \\
For each dimension, determine whether the text involves the given dimension. \\
Output \texttt{1} if it does, or \texttt{0} if it does not.
</instructions>

<text> \\
\{text\}
</text>

<response format> \\
Provide the answer by filling in 1 or 0 according to the instructions in the JSON format below.
</response format>

\{
    "Care/Harm": , \\
    "Fairness/Cheating": , \\
    "Loyalty/Betrayal": , \\
    "Authority/Subversion": , \\
    "Sanctity/Degradation": 
\}
\end{tcolorbox}
\subsubsection{Definition Prompt}
\label{sssec:defprompt}

Add the following section to the \modelname to get definition prompt. We use the same definitions and examples used in the work of ~\citet{ji2024moralbenchmoralevaluationllms} given its comprehensiveness.
\begin{tcolorbox}[colframe=black, colback=white, boxrule=0.5pt, title=Definition Section, breakable]

$<$definitions$>$\\
    Care/Harm: The Care/Harm foundation is rooted in the innate human capacity for
                empathy and compassion towards others. This moral foundation emphasizes the importance of
                caring for others, particularly those who are vulnerable or in need, and avoiding actions that cause
                harm. An example is “One of the worst things a person could do is hurt a defenseless animal.”
    
    Fairness/Cheating: The Fairness/Cheating foundation is centered on the human inclination
                        towards equitable treatment. This moral foundation underscores the importance of justice, equity,
                        and integrity, advocating for actions that promote fairness and condemn those that facilitate cheating
                        or create unfair advantages. An example is “Justice is the most important requirement for a society.”
    
    Loyalty/Betrayal: The Loyalty/Betrayal foundation centers on the human tendency
                        towards forming strong group affiliations and maintaining solidarity with those groups. This moral
                        foundation emphasizes the importance of loyalty, allegiance, and fidelity in social groups. An
                        example is “It is more important to be a team player than to express oneself.”

    Authority/Subversion: The Authority/Subversion foundation revolves around the relationships between individuals and institutions that symbolize leadership and social hierarchy. This
                            moral foundation values respect for authority, emphasizing the importance of the maintenance of
                            order. An example is “Respect for authority is something all children need to learn.”

    Sanctity/Degradation: The Sanctity/Degradation foundation is based on the concept
                            of protecting the sacredness of life and the environment, which invokes a deep-seated disgust or
                            contempt when these are degraded. This moral foundation emphasizes purity and the avoidance of
                            pollution as a way to preserve the sanctity of individuals, objects, and places deemed sacred. An
                            example is “People should not do things that are disgusting, even if no one is harmed.”
$<$/definitions$>$
\end{tcolorbox}

\subsubsection{Example Prompt}
\label{sssec:example_prompt}
The example section has three variations: The first version only has one example, and the second version extends the first version to 5 examples that are collected across the three inner domains: news, reddits, twitters, and covering every single moral foundation. \textbf{Example prompt} can be achieved by inserting either the first version or the second version of the example section into the \textbf{all@once prompt}. The last version is the same as the second version except that we added reasoning to the response. This last one is used only in \textbf{reason prompt}. We only present the last version of the example section for brevity. Getting rid of the reasonings recovers the second version, and getting rid of the last 4 examples recovers the first version.

\begin{tcolorbox}[colframe=black, colback=white, boxrule=0.5pt, title=Example with Reasoning Section, breakable]
$<$example 1$>$
    text: This Sandra Bland situation man no disrespect rest her soul, but people die everyday in a unjustified matter \#AllLivesMatter

    output:
    \{
        "Care/Harm": (The text mentions caring of people's death, 1),
        "Fairness/Cheating": (The text mentions whether people die in a justified manner, which is linked to fairness, 1),
        "Loyalty/Betrayal": (The text doesn't mention anything related to Loyalty/Betrayal, 0),
        "Authority/Subversion": (The text doesn't mention anything related to Authority/Subversion, 0),
        "Sanctity/Degradation": (The text doesn't mention anything related to Sanctity/Degradation, 0)
    \}$<$/example 1$>$

$<$example 2$>$\\
    text: It really breaks my heart to hear these stories of NY ers looting stealing amp scamming people affected by Sandy Fucking disgraceful

    output:
    \{
        "Care/Harm": (The text mentions feeling heart-broken when hearing people are scammed, which is linked to Care, 1),
        "Fairness/Cheating": (The text mentions looting stealing amp scamming people, which is a form of cheating and not fair, and is linked to fairness, 1),
        "Loyalty/Betrayal": (The text doesn't mention anything related to Loyalty/Betrayal, 0),
        "Authority/Subversion": (The text doesn't mention anything related to Authority/Subversion, 0),
        "Sanctity/Degradation": (The text mentions looting stealing amp scamming people is disgraceful, which is linked to Degradation, 1)
    \}\\[1ex]
$<$/example 2$>$

$<$example 3$>$\\
    text: That particular part of the debate is especially funny. Macron was explaining he did not think FN voters were evil and that from where he comes from he knew many, and she was like ""ooooh the evil FN invaders they're everywhere...!"" Self-awareness: zero.

    output:
    \{
        "Care/Harm": (The text doesn't mention anything related to Care/Harm, 0),
        "Fairness/Cheating": (The text doesn't mention anything related to Fairness/Cheating, 0),
        "Loyalty/Betrayal": (The text doesn't mention anything related to Loyalty/Betrayal, 0),
        "Authority/Subversion": (The text doesn't mention anything related to Authority/Subversion, 0),
        "Sanctity/Degradation": (The text mentions whether FN voters were evil, and evil is linked to Degradation, 1)
    \}\\[1ex]
$<$/example 3$>$

$<$example 4$>$\\
    text: Someone dying of a disease doesn't change that we've massively over reacted, and it's not as lethal as people are afraid of.

    output:
    \{
        "Care/Harm": (The text doesn't mention anything related to Care/Harm, 0),
        "Fairness/Cheating": (The text doesn't mention anything related to Fairness/Cheating, 0),
        "Loyalty/Betrayal": (The text doesn't mention anything related to Loyalty/Betrayal, 0),
        "Authority/Subversion": (The text doesn't mention anything related to Authority/Subversion, 0),
        "Sanctity/Degradation": (The text doesn't mention anything related to Sanctity/Degradation, 0)
    \}\\[1ex]
$<$/example 4$>$

$<$example 5$>$\\
    text: The lawsuit states that none of the promises for the project, intended to revive the nearby neighborhoods, were honored by the defendants.

    output:
    \{
        "Care/Harm": (The text doesn't mention anything related to Care/Harm, 0),
        "Fairness/Cheating": (The text mentions promises and lawsuits which are related to fairness, 1),
        "Loyalty/Betrayal": (The text mentions the promises were not honored, which is related to Betrayal, 1),
        "Authority/Subversion": (The text mentions lawsuits which are related to authority because the law is sustained by authority, 1),
        "Sanctity/Degradation": (The text doesn't mention anything related to Sanctity/Degradation, 0)
    \}\\[1ex]
$<$/example 5$>$
\end{tcolorbox}

\subsubsection{Lexicon Prompt - Consensus}
\label{sssec:lex_consensus_prompt}
Add the following Lexicon Section to the \textbf{baseline prompt}:
\begin{tcolorbox}[colframe=black, colback=white, boxrule=0.5pt, title=Lexicon Section]
$<$related-lexicons list$>$\\
    Care/Harm: ['war', 'wounded', 'cruel', 'suffer', 'suffering', 'killing', 'damaging', 'benefit', 'violence', 'killer', 'care', 'destroyed', 'protection', 'compassion', 'fight', 'damage', 'attack', 'kill', 'harm', 'abused', 'protected', 'brutality', 'fighting', 'destroy', 'hurt', 'safe', 'protect', 'harmful', 'attacker', 'violent', 'attacked', 'suffered', 'damaged']

    Fairness/Cheating:  ['equality', 'discrimination', 'equity', 'justice', 'integrity', 'unfair', 'equal', 'fair', 'justified', 'bias', 'prejudice', 'honest', 'injustice', 'law']

    Loyalty/Betrayal:  ['war', 'fellow', 'enemy', 'community', 'collective', 'rebellion', 'united', 'ally', 'solidarity', 'rebel', 'group', 'homeland', 'allegiance', 'family', 'nation']

    Authority/Subversion:  ['illegal', 'permit', 'comply', 'riot', 'permission', 'traditional', 'protection', 'controlling', 'order', 'rebellion', 'leader', 'president', 'ranking', 'controlled', 'protected', 'protect', 'regulation', 'leadership', 'respected', 'commander', 'rebel', 'duty', 'control', 'allegiance', 'refuse', 'authority', 'respect', 'tradition', 'law']

    Sanctity/Degradation:  ['clean', 'disease', 'integrity', 'sacred', 'church', 'dirty'] \\
$<$/related-lexicons list$>$
\end{tcolorbox}

and then change the instruction section to the following:

\begin{tcolorbox}[colframe=black, colback=white, boxrule=0.5pt, title=Lexicon Instruction Section,breakable]
$<$instructions$>$\\
    Iterate through five moral dimensions in [Care/Harm, Fairness/Cheating, Loyalty/Betrayal, Authority/Subversion, Sanctity/Degradation]\\
    The following related-lexicons list contains lexicons related to each dimension. For each dimension, determine whether the text involves the given dimension according to the definitions and incorporate the related-lexicons list in the decision, output 1 if it does, 0 if it doesn't.\\
$<$/instructions$>$
\end{tcolorbox}

For \textbf{lexicon prompt} with definitions or examples, insert the Definition Section or Example Section above.

\subsubsection{Lexicon Prompt - Performance Driven}
\label{sssec:lex_perf_prompt}

 For care, we add 20 words and remove 11 words. For fairness, we add 6 words and remove 5 words. For loyalty, we add 3 words and remove 6 words. For authority, we add 4 words and remove 10 words. For sanctity, we add 11 words and remove 1 word.
The Lexicon section in \textbf{lexicon prompt} is changed to the following updated list of lexicons:
\begin{tcolorbox}[colframe=black, colback=white, boxrule=0.5pt, title=Lexicon Instruction Section, breakable]
$<$related-words list$>$\\
    Care/Harm: ['exploits', 'genocidal', 'compassion', 'brutality', 'damage', 'rapists', 'destroyed', 'motherhood', 'bullied', 'benefit', 'violence', 'harmful', 'caring', 'suffered', 'healthcare', 'safe', 'murdered', 'wounded', 'killing', 'exploit', 'war', 'destroy', 'harm', 'endanger', 'protected', 'killer', 'protect', 'suffer', 'damaged', 'victims', 'vulnerability', 'condolences', 'kill', 'harassed', 'cruel', 'health', 'protecting', 'generous', 'violent', 'threaten', 'sympathy']\\
    Fairness/Cheating: ['racism', 'injustice', 'equity', 'discrimination', 'trustworthy', 'racist', 'fair', 'integrity', 'unfair', 'prejudice', 'proportional', 'equality', 'oppression', 'bias']\\
    Loyalty/Betrayal: ['homeland', 'allies', 'community', 'sacrifices', 'nation', 'allegiance', 'ally', 'solidarity', 'rebellion', 'fellow', 'kin', 'rebel']\\
    Authority/Subversion: ['ranking', 'submit', 'traitors', 'allegiance', 'respected', 'protected', 'law', 'comply', 'guide', 'duty', 'permit', 'commander', 'traditional', 'tradition', 'controlling', 'rebellion', 'authority', 'president', 'rebel', 'controlled', 'riot', 'regulation']\\
    Sanctity/Degradation: ['purity', 'disgusting', 'filthy', 'sins', 'disgusted', 'blessed', 'disgust', 'puke', 'church', 'dirty', 'deviants', 'integrity', 'eternal', 'raw', 'sacred', 'disease']\\
$<$/related-words list$>$
\end{tcolorbox}

\subsection{Combining Lexicons and LLMs}\label{subsec:lexLLM}
\sectextmoveup
We incorporate lexicons into our approach due to the extensive human effort and rigorous methodology behind the lexicon creation—such as the Moral Foundations Dictionary (MFD)~\citep{grahamLiberalsConservativesRely2009a}, MFD 2.0~\citep{frimerMoralFoundationsDictionary2019c}, and the Extended Moral Foundations Dictionary (eMFD)~\citep{hoppExtendedMoralFoundations2021}. Specifically, the original MFD used expert-selected vocabulary aligned with five moral foundations, approximately 32 words per foundation. MFD 2.0 expanded this work, using essays from over 1,000 participants across 58 countries to increase representativeness and validity. The eMFD utilizes crowd-sourced annotation, with over 500 individuals analyzing moral content in approximately 1,000 news articles. This resulted in a nuanced dictionary of 3,270 empirically validated words, each associated probabilistically with specific moral dimensions. References:
Thus, while LLMs perform well on their own, these carefully curated resources may still enhance moral analysis by providing structured, theory-driven signals that complement LLM-driven text analysis. We propose three methods for choosing and weighting words:

\headingmoveup
\paragraph{Consensus-based lexicon} adds a \textit{lexicon} block to the \textit{all@once} prompt. We use the 88 shared words from MFD, MFD2, and eMFD dictionaries provided by \citet{nguyen2024measuring}. See \Cref{sssec:lex_consensus_prompt} for word list on each foundation. 

\headingmoveup
\paragraph{Performance-driven lexicon} aims to modify the prompt by adding words that help and removing ones that hurts classification. We first identify sentences correctly classified by MFD2 but misclassified by \modelname, denote this set of sentences as \( S_c \). We then find \( S_w \), the set of sentences that are misclassified by MFD2 but correctly classified by \modelname. We then score each word in MFD2, initializing all scores at 0. Each time a word appears in a sentence from \( S_c \), its score increases by 1, whereas occurrences in \( S_w \) decrease its score by 1. Finally, we refine the lexicon by removing words with negative scores from the original set and adding those ranked in the top 50\% among positive-scoring words. The final lexicon updates the original 88 words by adding 44 that improve binary classification and removing 33 that negatively impact it -- resulting in the 99 words in \Cref{sssec:lex_perf_prompt}. This approach is named {\bf \modelname-Lex} for reporting results (\Cref{fig:eval,fig:prompting_results}).

\headingmoveup
\paragraph{LexLLM} aims to learn a weighted combination \(u_{j}^{(c)}\) between lexicon and LLM output. 
\[
    u_{j}^{(c)} = \lambda_j \, u_{j}^{(l)} + \bigl(1 - \lambda_j\bigr) u_{j}^{(m)},
\]
where $u_{j}^{(l)}$ is the output from a lexical classifier defined below, $u_{j}^{(m)}$ is LLM output probability described in \Cref{subsec:prompt_strat}, and \(\lambda_j \in [0,1]\) is a weighting factor tuned to maximize the AUC score. 

The second component is a lexical classifier that uses a predefined dictionary of terms (MFD 2.0) associated with each moral dimension. For each dimension \(j\), we train a separate binary logistic regression classifier that outputs the probability:
\[
    u_{j}^{(l)} = f_{\sigma}\bigl(\boldsymbol{\omega}_j^{\top} \mathbf{x}^{(l)}\bigr),
\]
where \(\boldsymbol{\omega}_j \in \mathrm{R}^{d}\) is the weight vector for dimension \(j\), and \(f_{\sigma}\) represents a logistic regression function.
Each sentence \(s\) is represented by a lexical feature vector \(\mathbf{x}^{(l)} \in \mathrm{R}^{d}\). Each element of this vector corresponds to a dictionary word, set to the presence of the word in the sentence (0 or 1). For example, given the sentence "\textit{Helping others is good.}" tokenized as ["\textit{Helping}", "\textit{others}", "\textit{is}", "\textit{good}"], the corresponding lexical feature vector might be \([1, 1, 0, 1, 0, \dots]\) where each position corresponds to a predefined dictionary word. 

\subsubsection{LexLLM parameters}\label{appenddix:lexllm}
\begin{figure}[!htbp]
    \centering
    \includegraphics[width=0.9\linewidth]{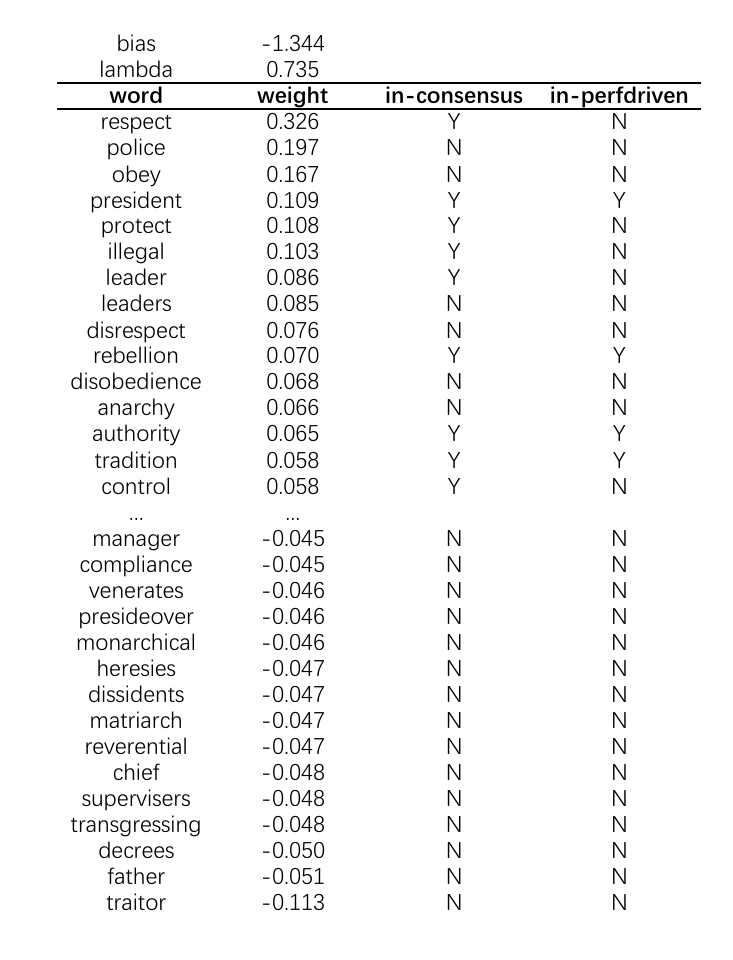}
    \caption{LexLLM parameters for Authority}
    \label{fig:LexLLM_A}
\end{figure}

\begin{figure}[!htbp]
    \centering
    \includegraphics[width=0.9\linewidth]{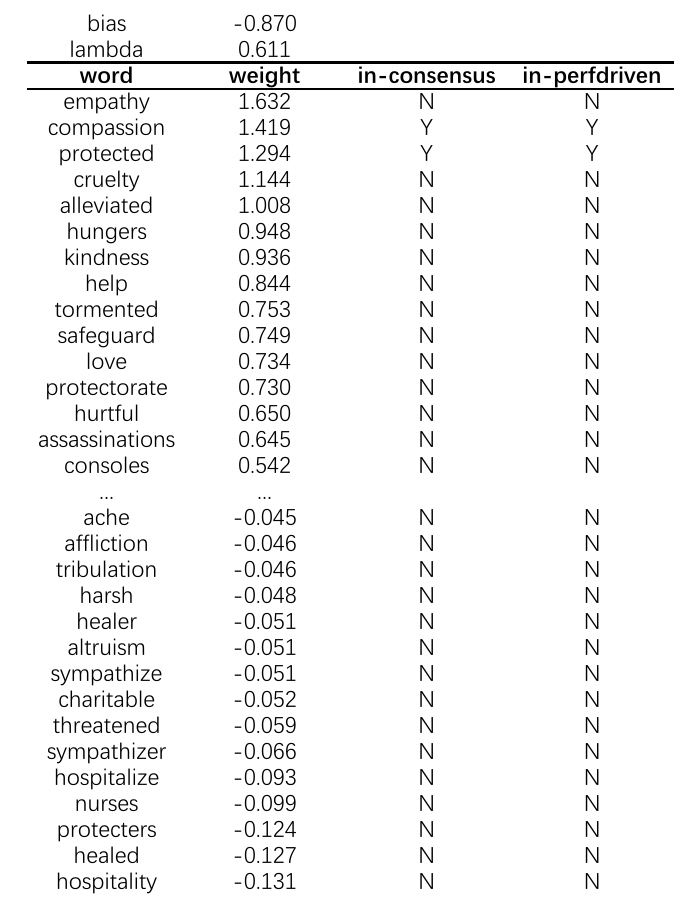}
    \caption{LexLLM parameters for Care}
    \label{fig:LexLLM_C}
\end{figure}

\begin{figure}[!htbp]
    \centering
    \includegraphics[width=0.9\linewidth]{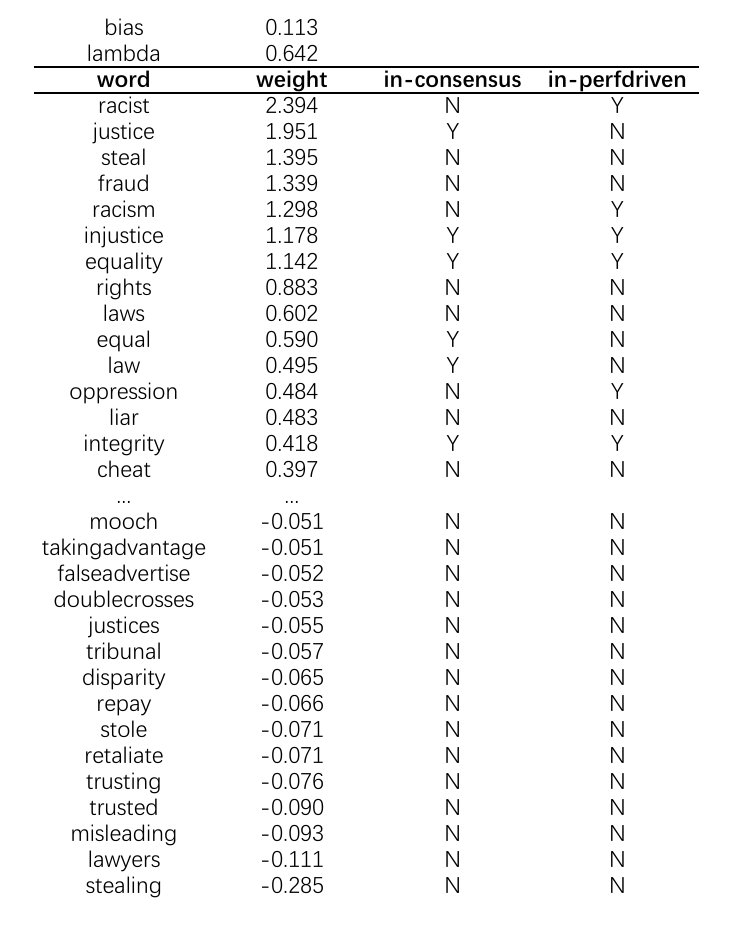}
    \caption{LexLLM parameters for Fairness}
    \label{fig:LexLLM_F}
\end{figure}

\begin{figure}[!htbp]
    \centering
    \includegraphics[width=0.9\linewidth]{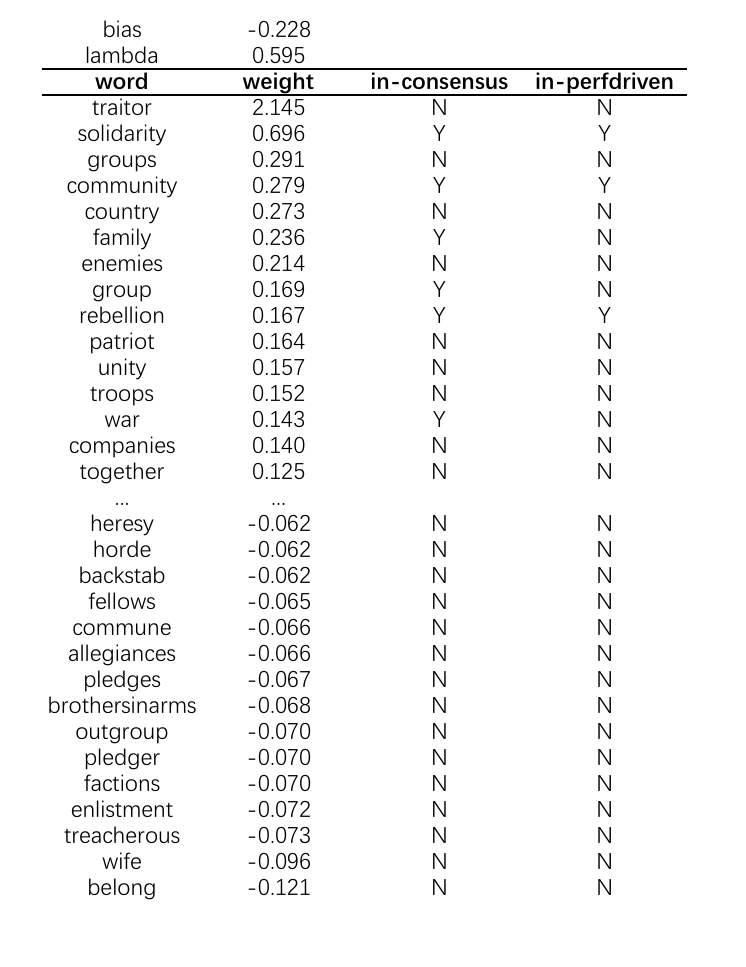}
    \caption{LexLLM parameters for Loyalty}
    \label{fig:LexLLM_L}
\end{figure}

\begin{figure}[!htbp]
    \centering
    \includegraphics[width=0.9\linewidth]{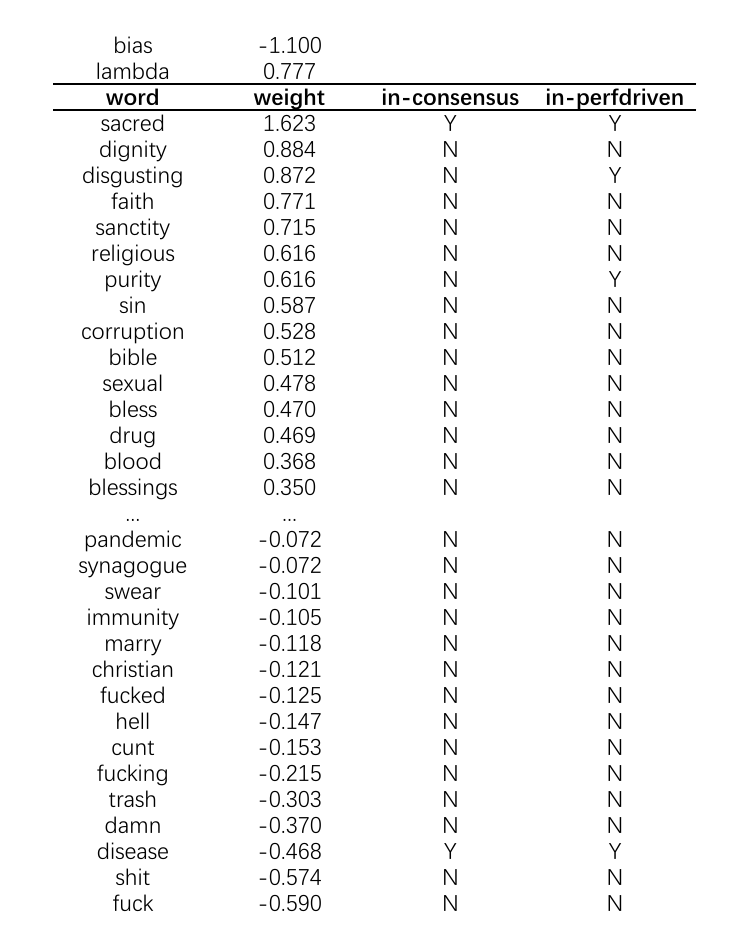}
    \caption{LexLLM parameters for Sanctity}
    \label{fig:LexLLM_S}
\end{figure}

For each moral foundation dimension, we present the trained $\lambda$ value and bias value. We also provide words in the MFD2 for the corresponding dimensions with the highest and lowest 15 trained weights for \textit{authority} (\Cref{fig:LexLLM_A}), \textit{care} (\Cref{fig:LexLLM_C}), \textit{fairness}(\Cref{fig:LexLLM_F}), \textit{loyalty} (\Cref{fig:LexLLM_L}) and \textit{sanctity} (\Cref{fig:LexLLM_S}). For each presented word, we also include whether they are included in the consensus-based lexicon list and in the performance-driven lexicon list.

\newpage
\subsection{Label Correlation Between Moral and Value Dimensions}\label{appendix:label_corr}

To assess the advantage of all-at-once (multi-label) classification, we examine label correlations within each framework using Chi-square tests and Phi coefficients. These results show many label pairs co-occur more often than chance, supporting the benefit of predicting labels jointly rather than one by one.

\subsubsection{Moral Foundations Theory (MFT)}

We analyze three in-domain MFT datasets annotated with five binary labels. Chi-square tests reveal significant associations between several label pairs (all $p < 0.05$) and Phi coefficients further show moderate correlation strength, as shown in \Cref{tab:mft_label_corr}.

\begin{table}[h]
\centering
\begin{tabular}{l l c c}
\toprule
Label 1 & Label 2 & Phi & $p$-value \\
\midrule
Sanctity & Loyalty & 0.24 & 2.0e-1115 \\
Authority & Loyalty & 0.21 & 7.3e-808 \\
Fairness & Care & 0.21 & 6.1e-804 \\
Loyalty & Fairness & 0.19 & 3.7e-715 \\
Care & Sanctity & 0.19 & 6.4e-666 \\
Authority & Sanctity & 0.18 & 1.7e-580 \\
Care & Loyalty & 0.17 & 1.8e-540 \\
Fairness & Authority & 0.16 & 6.9e-499 \\
Fairness & Sanctity & 0.14 & 1.8e-373 \\
Care & Authority & 0.12 & 2.32e-277 \\
\bottomrule
\end{tabular}
\caption{Label correlations in MFT datasets.}\label{tab:mft_label_corr}
\end{table}

\textit{Chi-square Test.}  
The Chi-square test determines whether two binary variables co-occur more (or less) often than expected under independence.  
The null hypothesis \(H_{0}\) states that the variables are independent; the alternative \(H_{1}\) states that they are associated.  
A p-value lower than \(0.05\) leads us to reject \(H_{0}\) and conclude that the two labels are statistically dependent.  

\textit{Phi Coefficient.}  
When the Chi-square test is significant, the Phi coefficient \((\phi)\) measures the strength of the association in a \(2\times2\) contingency table with cell counts \(a,b,c,d\):

\[
\phi \;=\; \frac{ad - bc}{\sqrt{(a + b)(c + d)(a + c)(b + d)}}.
\]

Its value ranges from \(0\) (no association) to \(1\) (perfect positive association), with negative values indicating an inverse relationship. Larger absolute values of \(\phi\) correspond to stronger links between the two binary labels.
\subsubsection{Human Values}

Using the Webis-22 dataset, we compute pairwise correlations across 10 human value categories. The top-10 significantly correlated label pairs are shown in \Cref{tab:HV_label_corr}:

\begin{table}[h]
\centering
\begin{tabular}{l l c c}
\toprule
Label 1 & Label 2 & Phi & $p$-value \\
\midrule
Hedonism & Stimulation & 0.32 & 9.0e-120 \\
Security & Self-direction & 0.21 & 8.5e-52 \\
Achievement & Power & 0.20 & 5.0e-46 \\
Self-direction & Conformity & 0.16 & 6.4e-31 \\
Stimulation & Self-direction & 0.16 & 1.0e-30 \\
Hedonism & Self-direction & 0.15 & 7.7e-28 \\
Conformity & Achievement & 0.14 & 3.0e-25 \\
Universalism & Tradition & 0.12 & 7.2e-18 \\
Conformity & Tradition & 0.12 & 9.9e-18 \\
Security & Tradition & 0.11 & 2.2e-16 \\
\bottomrule
\end{tabular}
\caption{Top 10 significantly correlated label pairs in Human Values.}\label{tab:HV_label_corr}
\end{table}

\subsubsection{MAC}

We apply the same analysis to the MAC-D dataset, focusing on the top 10 label pairs ranked by Phi coefficient. These correlations suggest that several high-level moral concepts frequently co-occur, particularly among family, group, and property-related dimensions.

\begin{table}[h]
\centering
\begin{tabular}{l l c c}
\toprule
Label 1 & Label 2 & Phi & $p$-value \\
\midrule
Family & Deference & 0.28 & 2.8e-165 \\
Group & Property & 0.24 & 7.9e-120 \\
Family & Group & 0.18 & 7.0e-68 \\
Reciprocity & Heroism & 0.13 & 9.9e-37 \\
Reciprocity & Property & 0.13 & 1.1e-35 \\
Group & Reciprocity & 0.13 & 6.8e-35 \\
Family & Reciprocity & 0.12 & 2.9e-30 \\
Heroism & Group & 0.09 & 1.2e-18 \\
Property & Family & 0.09 & 7.9e-18 \\
Property & Fairness & 0.07 & 6.1e-13 \\
\bottomrule
\end{tabular}
\caption{Top 10 correlated label pairs in MAC-D.}
\end{table}

These insights support our design choice: joint prediction lets LLMs use co-occurrence patterns between labels, helping improve moral context understanding.

\subsection{Finetuning LLMs}\label{appendix:finetuning}
To construct the training set, we sample instances from a merged dataset that includes three subsets—Twitter, Reddit, and News—provided by \citet{nguyen2024measuring}. We sample 50, 300, 3k, and 30k examples from each subset, with an equal number drawn from each subset. Each sample size uses an 80:20 split for training and validation. We will refer to the model trained with $X$ samples Finetuned-$X$. 

To preserve the original distribution of the five MFT dimensions within each sampled subset, we apply iterative stratification, which is suitable for multi-label data~\citep{sechidis2011stratification}. To ensure a fair comparison on unseen data, we exclude all test examples used in Mformer across all five dimensions from our samples. Because Mformer stratifies the entire dataset independently for each dimension during training and testing, this can result in the same example being used as a training example for one label (e.g., \textit{Authority}) and as a test example for another (e.g., \textit{Sanctity}). This filtering step yields approximately 60k examples, from which our current training and validation samples are drawn.

The trainings of GPT-4o-mini are all conducted with the fine-tuning API of OpenAI, which uses cross-entropy loss. The fine-tuning API itself determines the epochs and batch sizes. Finetuned-$50$ is trained with $3$ epochs and a batch size of $1$, with a final train loss of $0$ and a final validation loss of $0.066$. Finetuned-$300$ is also trained with $3$ epochs and a batch size of $1$, with a final train loss of $0$ and a final validation loss of $0.089$. Finetuned-$3k$ is trained with $3$ epochs and a batch size of $14$, with a final train loss of $0.023$ and a final validation loss of $0.036$. Finally, Finetuned-$30k$ is trained with $1$ epoch and a batch size of $29$, with a final train loss of $0.032$ and a final validation loss of $0.035$.

\subsection{ Extract label probability from LLM prompts}\label{appenddix:extract_prob_antitoken}

We extract probability scores from LLMs to measure the model’s certainty in labelling instances. These scores can be used in downstream tasks, such as computing ranking metrics such as the area under the curve (AUC) (\Cref{sec:results}) and model combination (\Cref{subsec:lexLLM}). 

For both GPT and open source models, we extract the log probability of the chosen label (0 or 1) for each dimension. We define the \emph{anti-token} as the label opposite to the chosen label per dimension (e.g., if the chosen label is $0$, then the anti-token is $1$). We then check whether the anti-token appears among the top $20$ most likely tokens in the model output for each dimension. If we find the anti-token, we rescale the probabilities of the predicted token and the anti-token so that they add up to $1.0$. We denote the rescaled probability of the predicted token as $u'$. If the anti-token does not appear among the top $20$ tokens, we assign $u' = 1.0$ to the predicted token because it is the only valid token for binary labels within the subset of the model’s output. 
Finally, we map probability $u'$ to a final probability $u^{(m)}$ of the label being \texttt{1}: \[
u^{(m)}=\left\{
\begin{aligned}
u' &, \quad \text{if the predicted token is }1,\\
1-u' &, \quad \text{if the predicted token is }0.
\end{aligned}
\right.
\]

\begin{tcolorbox}[colframe=black, colback=white, boxrule=0.5pt, title=Anti-token example,breakable]
\textbf{Statement:} Whether or not someone violated standards of purity and decency

\textbf{LLM Prompt Output:}
\begin{verbatim}
{
    "Care/Harm": 0,
    "Fairness/Cheating": 0,
    "Loyalty/Betrayal": 0,
    "Authority/Subversion": 0,
    "Sanctity/Degradation": 1
}
\end{verbatim}

\textbf{Top 20 tokens with log-probabilities in LLM:}

tokens for care: [`0'(-0.2), `1'(-1.4), ...]

tokens for fairness: [`0'(-0.01), ...]

tokens for loyalty: [`0'(-0.001), ...]

tokens for authority: [`0'(-0.4), `1'(-1.5), ...]

tokens for sanctity: [`1'(-0.36), ..., `0'(-1.20)]

\end{tcolorbox}

For instance, in the \textit{Sanctity/Degradation} dimension of this anti-token example, the predicted token is `1', and the anti-token `0' also appears among the top 20 tokens. Suppose their log-probabilities are \(-0.36\) for `1' and \(-1.20\) for `0'. Exponentiating gives raw probabilities of approximately \( \exp(-0.36) \approx 0.697 \) and \( \exp(-1.20) \approx 0.301 \). We normalize these to compute the predicted probability as \( u' = \frac{0.697}{0.697 + 0.301} \approx 0.7 \). Since the predicted token is `1', we set \( u^{(m)} = 0.7 \). 
For in the \textit{Fairness/Cheating} dimension of this example, the anti-token `1' had not appeared among the top 20 tokens, we would have assigned \( u' = 1.0 \), resulting in \( u^{(m)} = 1.0 \), reflecting full model confidence in the label.

\textbf{Observation with DeepSeek V3.} We find that, regardless of the values specified for temperature or top-p, the returned log-probabilities for all candidate tokens are consistently set to $-9999.0$, resulting in exponentiated probabilities that are effectively zero. So the predicted probability for each text instance becomes either $0.0$ or $1.0$, making threshold calibration or probabilistic interpretation infeasible. This suggests that the log-probability feature is not yet fully supported in DeepSeek's current API inference pipeline. Therefore, for tasks that require calibrated probabilities—such as threshold selection or AUC computation—we use \modelname\ (all-at-once prompting strategy and GPT-4o-mini), which provides more reliable probabilistic outputs.

\subsection{ Random Classifier Baseline}\label{sec:random_baseline}

To provide context for interpreting our models' performance, we compute the theoretical F1 score of a random baseline classifier. This classifier does not learn from input features—it simply predicts the positive class with probability \( p \). To create a fair baseline, we set \( p \) equal to the true proportion of positive labels in the dataset, denoted as \( r \).

This calibration ensures that the classifier predicts positive with the same frequency as positives appear in the data.

Let the dataset contain \( N \) examples. Then:

\begin{itemize}
    \item The number of actual positives is \( rN \)
    \item The number of predicted positives is \( pN = rN \)
    \item The number of true positives (TP) is the expected overlap between actual and predicted positives: \( \text{TP} = r \times r \times N = r^2N \)
    \item The number of false positives (FP) is the portion of predicted positives that are not actual positives: \( \text{FP} = (1 - r) \times r \times N = r(1 - r)N \)
    \item The number of false negatives (FN) is the portion of actual positives not predicted: \( \text{FN} = r \times (1 - r) \times N = r(1 - r)N \)
\end{itemize}

Now we compute precision and recall:

\begin{align*}
\text{Precision} &= \frac{\text{TP}}{\text{TP} + \text{FP}} \\
&= \frac{r^2N}{r^2N + r(1 - r)N} \\
&= \frac{r^2}{r^2 + r(1 - r)} \\
&= \frac{r^2}{r} = r
\end{align*}

\begin{align*}
\text{Recall} &= \frac{\text{TP}}{\text{TP} + \text{FN}} \\
&= \frac{r^2N}{r^2N + r(1 - r)N} \\
&= \frac{r^2}{r^2 + r(1 - r)} \\
&= \frac{r^2}{r} = r
\end{align*}

So when \( p = r \), both precision and recall equal \( r \), and the F1 score becomes:

\[
\text{F1}_{\text{random}} = \frac{2 \cdot r \cdot r}{r + r} = \frac{2r^2}{2r} = r
\]

This confirms that a calibrated random classifier has equal precision and recall, both equal to the positive label rate \( r \), and thus F1 score also equals \( r \).

For example, in the MIC dataset’s \textit{liberty} classification task, the positive rate is \( r = 0.1923 \), so:

\[
\text{F1}_{\text{random}} = 0.1923
\]

\subsection{ Evaluation Metrics for Human Values}\label{appendix:humval_eval_metrics}

\paragraph{Macro F1}
We used the unmodified \textsc{Webis-22}'s evaluation methods for the reported results of the dataset.  
For every value dimension $v$, the script first counts
\emph{relevant} instances ($\mathrm{Rel}_v$), defined as test statements whose gold label for $v$ equals~1,  
and \emph{positive} predictions ($\mathrm{Pos}_v$), defined as statements for which our system outputs~1.  
True positives are
\[
\mathrm{TP}_v = \lvert \mathrm{Rel}_v \cap \mathrm{Pos}_v \rvert ,
\]
and true negatives are
\[
\mathrm{TN}_v = \lvert \overline{\mathrm{Rel}_v} \cap \overline{\mathrm{Pos}_v} \rvert ,
\]
that is, instances where both the gold label and the system prediction equal~0.

From these counts the evaluation derives  
\begin{align}
\mathrm{Precision}_v &= \frac{\mathrm{TP}_v}{\mathrm{Pos}_v},\\[2pt]
\mathrm{Recall}_v &= \frac{\mathrm{TP}_v}{\mathrm{Rel}_v},\\[2pt]
\mathrm{F1}_v &= \frac{2\,\mathrm{Precision}_v\,\mathrm{Recall}_v}
                    {\mathrm{Precision}_v + \mathrm{Recall}_v},\\[2pt]
\mathrm{Accuracy}_v &= \frac{\mathrm{TP}_v + \mathrm{TN}_v}{N},
\end{align}
where $N$ is the number of test instances.  
If $\mathrm{Rel}_v = 0$ for a region, the dimension $v$ is skipped, and it does not contribute to macro scores.  

Macro precision and macro recall are the arithmetic means of the per-dimension values that remain:
\begin{align}
\mathrm{MacroPrec} &= \frac{1}{|\mathcal{V}|}\sum_{v\in\mathcal{V}} \mathrm{Precision}_v,\\[2pt]
\mathrm{MacroRec}  &= \frac{1}{|\mathcal{V}|}\sum_{v\in\mathcal{V}} \mathrm{Recall}_v,
\end{align}
with $\mathcal{V}$ the set of dimensions that occur at least once in the gold labels for the region.  
The reported macro~F1 is then recomputed from these two aggregates,
\[
\mathrm{MacroF1} \;=\; 
\frac{2\,\mathrm{MacroPrec}\,\mathrm{MacroRec}}
     {\mathrm{MacroPrec} + \mathrm{MacroRec}},
\]
and macro accuracy is the mean of per-dimension accuracies.  
Predicted probabilities were thresholded at $0.5$ before evaluation, as required by the task.  
All metrics are reported separately for the four geographic regions: USA, Africa, China, and India.

\paragraph{Accuracy}
For both the \textsc{ValEval-MFT} and \textsc{ValEval-Schwartz} datasets, we follow their evaluation protocol and report accuracy as the main metric.  
For each test example, we identify the target value and compare the system's predicted label for that value with the gold label.  
Each label must be one of three strings:
\begin{itemize}
    \item \texttt{Yes} – the response supports the value,
    \item \texttt{No} – the response contradicts the value, or
    \item \texttt{Not related} – the response does not refer to the value.
\end{itemize}
We compute accuracy as the proportion of examples for which the predicted label exactly matches the gold label, using \texttt{sklearn.metrics.accuracy\_score}.
  
\paragraph{F1 and Accuracy.}
For \textsc{ValueNet}, we follow the official evaluation protocol and compute prediction precision, recall, F1 score, and accuracy by first rounding the model outputs to the nearest integer.  
Accuracy is reported per dimension, while precision, recall, and F1 score are computed per value direction: $-1$ for \texttt{No}, $1$ for \texttt{Yes}, and $0$ for \texttt{Not related}.

\subsection{Costs of Prompting and Finetuning}
\label{sec:cost_table}

Our approach is cost-efficient and provides an accessible measurement tool for social science researchers, particularly because it does not require fine-tuning. This makes it more accessible to those with limited GPU resources. For example, using the GPT-4o-mini API—the most lightweight and efficient model in the GPT-4 family—results in comparably low costs to the DeepSeek-v3 API. 

\subsubsection{API-Level Cost Comparison}\label{appendix:cost}
\begin{table}[!htbp]
\centering
\caption{Cost Comparison per 1M Tokens: GPT-4o-mini vs DeepSeek-v3 (pricing as of April 2025)}
\begin{tabularx}{\linewidth}{lXX}
\toprule
Category & GPT-4o-mini API & DeepSeek-v3 API \\
\midrule
Cached input           & \$0.075 & \$0.070 \\
Input        & \$0.150 & \$0.270 \\
Output                 & \$0.600 & \$1.100 \\
\bottomrule
\end{tabularx}
\end{table}

\subsubsection{Dataset-Specific Costs}
Table~\ref{tab:cost} reports estimated per-1K input token costs for GPT-4o-mini when using the \textit{all@once} prompt, with the number of examples and the average length (in tokens) for each dataset. 

\begin{table}[!htbp]
\centering
\caption{Estimated dataset-specific costs (per 1K input tokens) for GPT-4o-mini using the \textit{all@once} prompt.}
\begin{tabularx}{\linewidth}{lXXXX}
\toprule
 & Reddit & Twitter & MIC & SC \\
\midrule
Cost (1K tokens) & \$0.072 & \$0.068 & \$0.065 & \$0.066 \\
\# of examples     & 17,886  & 34,987  & 6,235   & 5,122   \\
Avg. tokens / ex.  & 41.7    & 19.3    & 52.1    & 47.8    \\
\bottomrule
\end{tabularx}
\label{tab:cost}
\end{table}

\subsubsection{Fine-tuning Costs}

Table~\ref{tab:finetune_cost} shows estimated costs of fine-tuning GPT-4o-mini using the \modelname prompt on varying amounts of Mformer’s training data (\textit{In-domain} Group).

\begin{table}[!htbp]
\centering
\caption{Estimated fine-tuning costs for GPT-4o-mini with the \modelname prompt on Mformer’s training data (\textit{In-domain} Group).}
\begin{tabularx}{0.7\linewidth}{lX}
\toprule
\# of Examples & Estimated Cost \\
\midrule
30,000 & \$39.00 \\
3,000  & \$19.00 \\
300    & \$0.63 \\
50     & \$0.33 \\
\bottomrule
\end{tabularx}
\label{tab:finetune_cost}
\end{table}

\subsection{Result stability of ~\modelname}\label{sec:prompt_stability}
Performance appears robust with respect to the inclusion of definitions and changes in phrasing in \Cref{tab:wording_change}. Specifically, when changing the leading sentence from ``whether this morality-related text involves the five dimensions'' to ``whether each of the five dimensions of moral foundations is relevant in the text,'' and revising the instruction from ``For each dimension, determine whether the text involves the given dimension'' to ``Determine whether each dimension is relevant in the text,'' the AUC remains unchanged or differs by only 0.01.

\begin{table}[!htbp]
\centering
\caption{Impact of Wording Changes on AUC (all@once prompt)}
\begin{tabularx}{\linewidth}{llXXXXX}
\toprule
Prompt & Change & A & C & F & L & S \\
\midrule
all@once & No  & 0.76 & 0.81 & 0.83 & 0.72 & 0.73 \\
all@once & Yes & 0.76 & 0.81 & 0.83 & 0.73 & 0.73 \\
\bottomrule
\end{tabularx}\label{tab:wording_change}
\end{table}

\subsubsection{Label Ordering in Classifier Chain Method}\label{appendix:order}
We examined the impact of label ordering in the classifier chain method. As shown in \Cref{tab:chain_order}, the ordering effects are relatively small: performance differences on the Reddit dataset across orders are within 0.02 on average. Importantly, all classifier chain variants consistently outperform the simple 1-by-1 baseline across the five MFT dimensions.

\begin{table*}[t]
\centering
\caption{Performance of the classifier chain method under different label orders, compared to the 1-by-1 baseline. Reported scores are F1 values.}
\label{tab:chain_order}
\begin{tabular}{lcccccc}
\toprule
Dimension & 1-by-1 & Chain (F,C,L,S,A) & Chain (C,F,L,S,A) & Chain (A,F,L,C,S) & Mean & Std \\
\midrule
Care       & 0.35 & 0.59 & 0.59 & 0.57 & 0.58 & 0.01 \\
Fairness   & 0.37 & 0.55 & 0.53 & 0.56 & 0.55 & 0.02 \\
Loyalty    & 0.31 & 0.40 & 0.38 & 0.40 & 0.39 & 0.01 \\
Authority  & 0.17 & 0.46 & 0.47 & 0.47 & 0.47 & 0.01 \\
Sanctity   & 0.26 & 0.36 & 0.34 & 0.35 & 0.35 & 0.01 \\
\bottomrule
\end{tabular}
\end{table*}
Performance is also stable under different orderings of the five Moral Foundations dimensions. \Cref{tab:order_change} reports AUC scores for the all@once prompt under five permutations on the Twitter dataset, where each foundation appears first in one of the variants. The original ordering (C,F,L,A,S) matches the one used in \Cref{fig:prompt_structure}B.

\begin{table}[!h]
\centering
\caption{Effect of Label Order on AUC (all@once prompt)}
\begin{tabularx}{\linewidth}{lXXXXX}
\toprule
Label Ordering & A & C & F & L & S \\
\midrule
(C,F,L,A,S) & 0.76 & 0.81 & 0.83 & 0.73 & 0.73 \\
(A,L,S,F,C) & 0.78 & 0.80 & 0.83 & 0.72 & 0.73 \\
(F,C,L,A,S) & 0.76 & 0.81 & 0.84 & 0.73 & 0.74 \\
(L,A,S,F,C) & 0.76 & 0.80 & 0.83 & 0.73 & 0.73 \\
(S,A,L,F,C) & 0.78 & 0.80 & 0.83 & 0.72 & 0.74 \\
\midrule
Mean & 0.77 & 0.80 & 0.83 & 0.73 & 0.73 \\
SD   & 0.01 & 0.01 & 0.00 & 0.01 & 0.01 \\
\bottomrule
\end{tabularx}\label{tab:order_change}
\end{table}

\subsubsection{Stability Across Repeated Runs}

We ran \modelname{} five times using the same prompt (all@once, GPT-4o-mini). \Cref{tab:stability} shows that the resulting AUC scores have extremely low variance, suggesting that the output extraction procedure is stable and deterministic.

\begin{table}[!h]
\centering
\caption{AUC Scores Across Five Runs (all@once, GPT-4o-mini)}
\begin{tabularx}{\linewidth}{lXXXXX}
\toprule
Run & A & C & F & L & S \\
\midrule
time1 & 0.76 & 0.81 & 0.83 & 0.72 & 0.73 \\
time2 & 0.76 & 0.81 & 0.83 & 0.72 & 0.73 \\
time3 & 0.76 & 0.81 & 0.83 & 0.72 & 0.73 \\
time4 & 0.76 & 0.81 & 0.83 & 0.72 & 0.72 \\
time5 & 0.76 & 0.81 & 0.83 & 0.72 & 0.73 \\
\midrule
Mean & 0.76 & 0.81 & 0.83 & 0.72 & 0.73 \\
SD   & 0.00 & 0.00 & 0.00 & 0.00 & 0.00 \\
\bottomrule
\end{tabularx}\label{tab:stability}
\end{table}

\section{In-domain Evaluations on MFT}\label{sec:Results}
\secmoveup
\subsection{Evaluating Prompting Strategies}\label{appendix:mformer_mova}
\label{sec:results}
\sectextmoveup
We evaluate \modelname~on the three datasets used by recent approaches~\cite{nguyen2024measuring,rathje2024gpt,Dehghani2024gpt}. Table \ref{table:dataset:mf_dataset_stats_internal} provides dataset profiles.

\headingmoveup
\paragraph{Training and evaluation settings.}
We reuse the 90-10 train-test split by Mformer, the only work that evaluates across three datasets \cite{nguyen2024measuring} so that we can directly compare results. The lexical classifier in \Cref{subsec:lexLLM} learns $\omega_j$ and $\lambda_j$ for each foundation $j$ on the training portion of all three domains, performing a 10-fold cross-validation on the training portion to search for the learning rate that maximizes the AUC score.

\headingmoveup
\paragraph{Evaluation metrics.} 
We adopt both \textbf{AUC} and \textbf{F1} metrics to evaluate the performance of LLM models. AUC measures how well a classifier ranks positive instances higher than negative ones, regardless of the threshold used. It is particularly suitable for this task because it considers probability scores rather than binary predictions and remains robust under imbalanced datasets, where certain foundations—such as \textit{sanctity} or \textit{loyalty}—appear in fewer than 10\% of instances in several datasets (\Cref{table:dataset:mf_dataset_stats_internal}). 
We also report the F1 score for comparing our approach to baseline methods, such as the one proposed by \citet{rathje2024gpt}. We conduct \textit{Wilcoxon signed-rank test} as a non-parametric test to detect significant differences between different methods (details in \Cref{sec:Results_Wilcoxon}).

\begin{table}[!htbp] 
\centering \small
\begin{tabular}{@{}p{2.3cm}@{} p{1.3cm} p{1.3cm} p{1.3cm}@{}} 
\toprule 
Source & Twitter & News & Reddit \\ 
\midrule 
\# of Examples & 34,987 & 34,262 & 17,886 \\ 
Avg. tokens & 19.3 & 28.0 & 41.7 \\ 
\midrule 
\% Authority & 33.4 & 24.9 & 19.2 \\ 
\% Care & 40.6 & 24.8 & 26.5 \\ 
\% Fairness & 35.9 & 24.2 & 29.5 \\ 
\% Loyalty & 31.1 & 24.4 & 11.1 \\ 
\% Sanctity & 22.3 & 19.9 & 9.8 \\ 
\bottomrule 
\end{tabular} 
\caption{Profile of three in-domain datasets with number of examples, and percentage of each moral foundation.}
\label{table:dataset:mf_dataset_stats_internal} 
\end{table}
\afterfigmoveup

\begin{figure}[!htbp]
    \centering
    \includegraphics[width=1\linewidth]{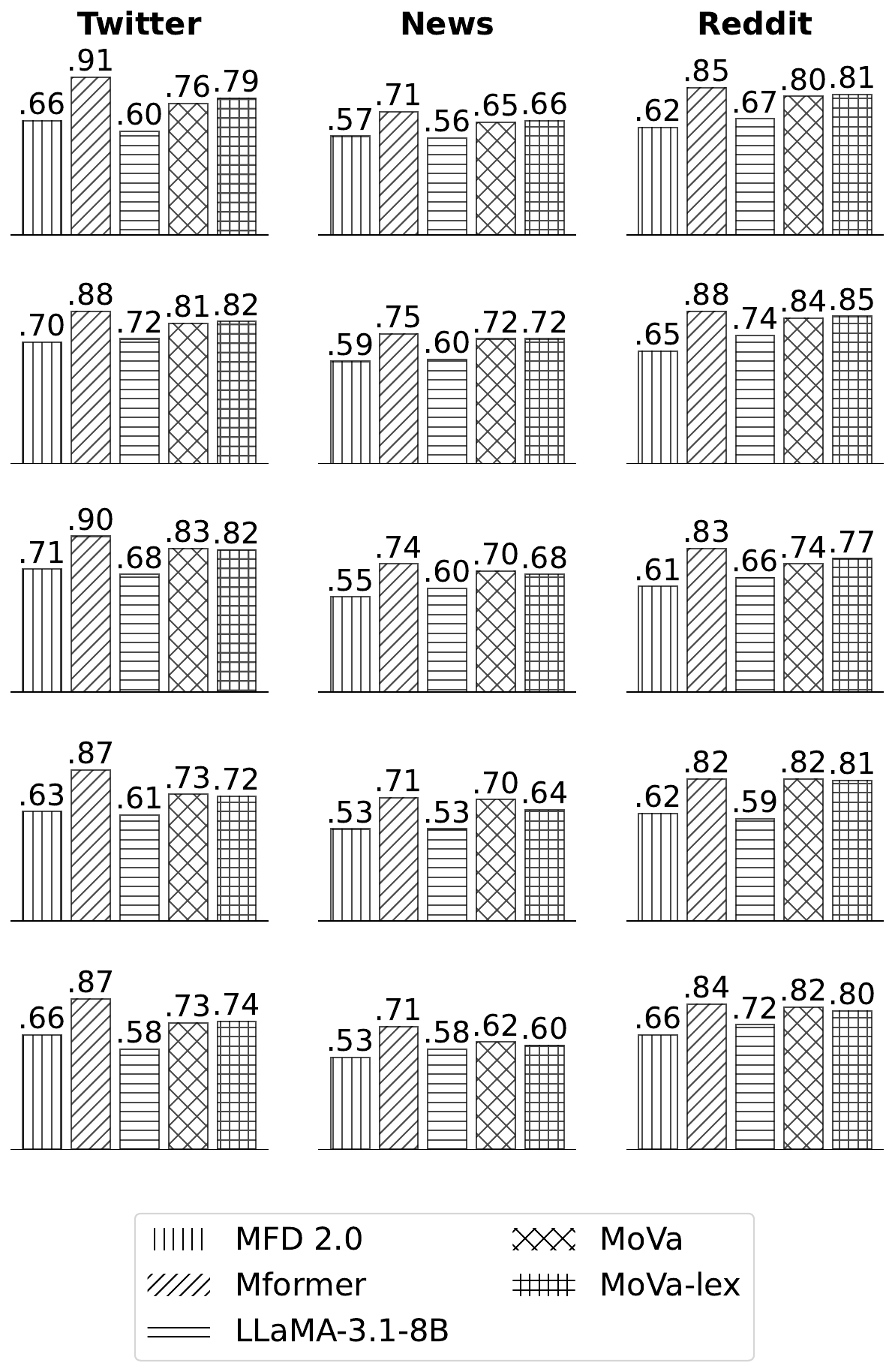}
    \caption{AUC scores of \modelname, \modelname~+ performance-driven lexicon prompt and LLaMA-3-8B against other baselines, MFD 2.0 and Mformer, on the Reddit, Twitter, and News test sets.}
    \label{fig:in_domain_models}
\end{figure}

\Cref{fig:in_domain_models} reports the AUC scores of different approaches. Among these, MFD2.0, \modelname~and LLaMA-3.1-8B with no learning for this task, Mformer~\cite{nguyen2024measuring} and \modelname-lex are tuned  on the respective datasets.

Mformer, a fine-tuned transformer network, has the highest performance among approaches with learning component. \modelname~tops the performance among those without learning. 
\modelname-lex, designed to mitigate errors between \modelname~and MFD2.0 lexicon, improves upon \modelname~in 7 out of the 15 tasks (5 task x 3 datasets).
The observation that learning-based approaches outperform non-learning ones on the datasets they are trained is consistent with those observed in recent work \cite{rathje2024gpt,Dehghani2024gpt}. However, the potential for non-learning approaches is still underexplored for new domains and new tasks (\Cref{sec:mft,sec:schwartz,sec:MAC_commonmorality}).

\begin{figure}[!htbp]
    \centering
    \includegraphics[width=1\linewidth]{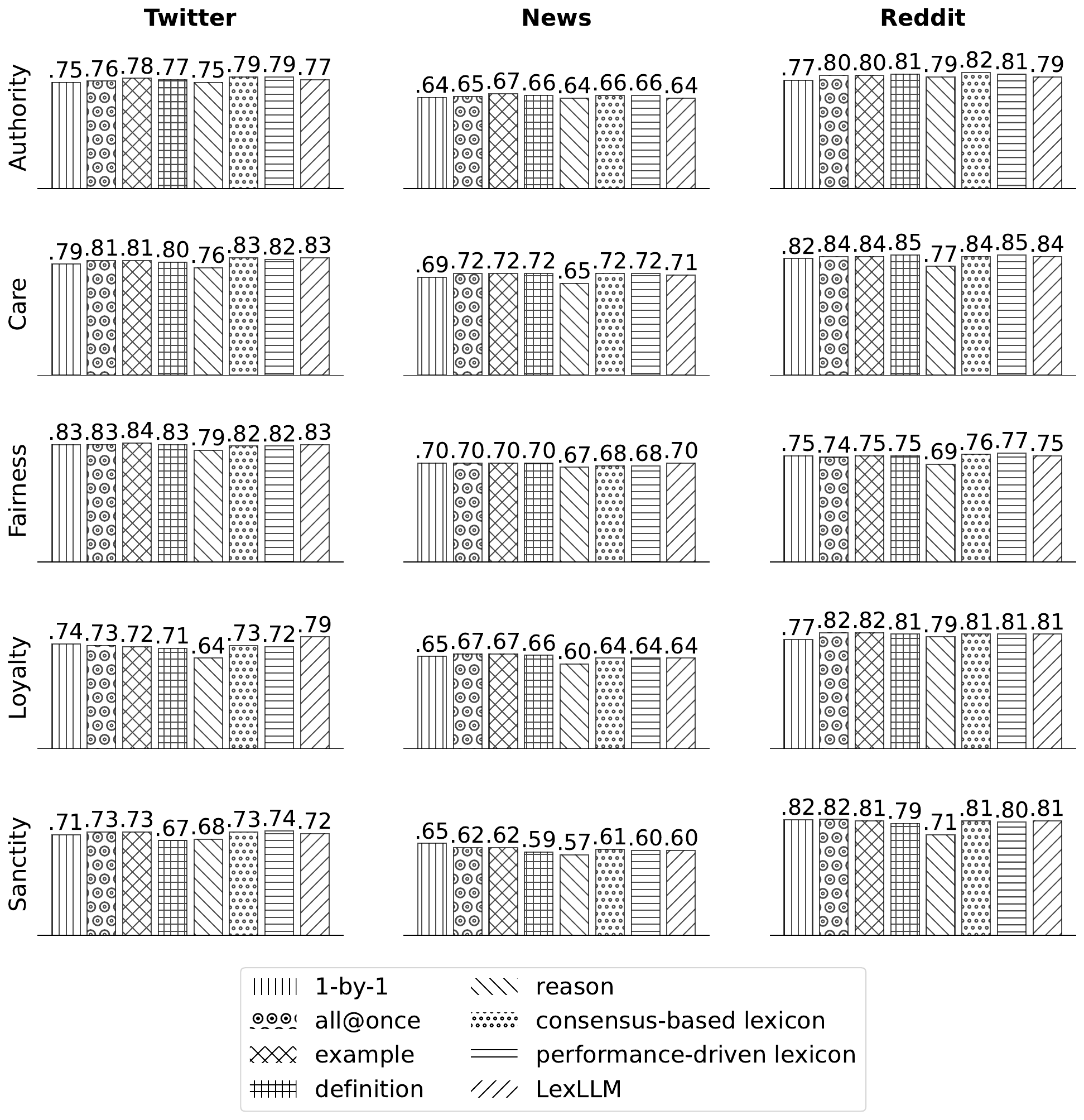}
    \caption{AUC on three Mformer's test sets for eight prompting strategies.}
    \label{fig:prompting_results}
\end{figure}

\Cref{subsec:other_LLMs,sec:Results_Wilcoxon} present detailed comparisons and statistical analyses of different prompting strategies and LLMs.
\modelname~with \emph{all@once} prompt significantly outperforms separate binary classification (\emph{1-by-1}) (\textit{p} < 0.05). \textit{\modelname}+\textit{example} also significantly outperforms \textit{1-by-1} (\textit{p} < 0.05). However, extensions of \textit{all@once}—including \textit{definition}, \textit{example} and \textit{reason}, do not offer significant improvement compared to \textit{all@once} (\Cref{fig:prompting_results}). 
The \textit{\modelname}+\textit{reason} prompt performs significantly worse than all others, suggesting that scoring moral relevance may not require analytical reasoning \citep{Sprague2024ToCO}.

For combining lexicon with LLMs, although none of these three methods shows a statistically significant advantage over \emph{all@once}, each yields some notable improvements. Among the 7 tasks, the Performance-driven lexicon improves upon \emph{all@once} in AUC on all three {\it authority} tasks, two {\it care} tasks, one {\it loyalty} task, and one {\it sanctity} task. Consensus-based lexicon prompting improves 5 of 15 tasks, and LexLLM improves 4. See detailed lexicon results (word lists and weight tables) in \Cref{appenddix:lexllm}. These observations suggest that task difficulty may vary across moral foundations, and robust combinations between LLMs and linguistic resources such as expert-curated lexicon is promising but an open direction.

Apart from GPT-4o-mini, we also use the \emph{all@once} prompt on several other LLMs, including GPT-3.5-Turbo, LLaMA-3.1-8B, LLaMA-3.1-70B, and DeepSeek-R1-Distill-Llama-8B (\Cref{subsec:other_LLMs}). 
LLaMA-3.1-8B shows a decrease in AUC of $0.10$ to $0.15$ compared to GPT-4o-mini (\Cref{fig:in_domain_models}).
LLaMA-3.1-70B exhibits marginal changes in AUC, ranging from an increase of 0.01 to a decrease of 0.01, with a significantly higher computational cost.
DeepSeek-R1-Distill-Llama-8B consistently scored below 0.60 AUC, likely due to its limited instruction-following capability for structured output.
GPT-3.5-Turbo shows a decrease in AUC of 0.04 to 0.10 compared to GPT-4o-mini.

\subsection{Other LLMs we tried}\label{subsec:other_LLMs}
\begin{table*}[!tb]
    \centering
    \resizebox{\textwidth}{!}{
        \begin{tabular}{l c c c c c}
            \toprule
            LLMs & Authority\_AUC & Care\_AUC & Fairness\_AUC & Loyalty\_AUC & Sanctity\_AUC \\
            \midrule
            gpt-4o-mini & 0.80 & 0.84 & 0.74 & 0.82 & 0.82 \\
            gpt-3.5-turbo & 0.77 & 0.80 & 0.70 & 0.74 & 0.73 \\
            LLaMA-3.1-8B & 0.67 & 0.74 & 0.66 & 0.59 & 0.72 \\
            DeepSeek-R1-Distill-Llama-8B & 0.56 & 0.60 & 0.58 & 0.58 & 0.59 \\
            LLaMA-3.1-70B & 0.68 & 0.73 & 0.67 & 0.67 & 0.71 \\
            \bottomrule
        \end{tabular}
    }
    \caption{AUC scores across LLMs with all@once prompt for five moral foundations in the Reddit dataset.}
    \label{tab:LLM_auc_scores}
\end{table*}

Apart from GPT-4o-mini, we also applied the \emph{all@once} prompt in several other LLMs, including LLaMA-3.1-8B, LLaMA-3.1-70B, and DeepSeek-R1-Distill-Llama-8B, to assess their performance in moral foundation classification. Our primary goal was to examine whether model size, instruction tuning, or architectural differences significantly impacted classification accuracy. These experiments faced hardware limits. Running on no more than 4 NVIDIA A100 GPUs was insufficient for the largest models and caused slow runtimes. We therefore restricted our experiments to the smaller models listed below.

\paragraph{LLaMA-3.1-8B} achieved worse performance, with AUC scores ranging from 0.59 to 0.74 across different moral foundations, which is approximately 10-15\% lower than GPT-4o-mini. The model performed best on \textit{Care} and \textit{Sanctity}, but struggled with \textit{Loyalty}, where its AUC was below 0.60. Scaling up to LLaMA-3.1-70B led to marginal improvements, particularly in \textit{Loyalty}, where it reached an AUC of 0.67, but overall, the gains were small compared to the increased computational cost, suggesting that model size alone does not necessarily enhance classification.

\paragraph{LLaMA-3.1-70B} demonstrated a slight improvement over LLaMA-3.1-8B across most moral foundations. However, the performance gap between the two models was marginal, suggesting that increasing model size alone does not guarantee significantly better moral foundation classification. The most notable gain was observed in the \textit{Loyalty} category, where LLaMA-3.1-70B showed a meaningful increase compared to its 8B counterpart. Despite its larger parameter count, its AUC scores still fell short of GPT-4o-mini, reinforcing the importance of high-quality instruction tuning over model size.

\paragraph{DeepSeek-R1-Distill-Llama-8B} 
DeepSeek-R1-Distill-Llama-8B exhibited the weakest performance consistently below 0.60 among the tested models, underperforming in all five moral foundations compared to LLaMA models and GPT-based models. One key issue encountered was its weaker instruction-following ability. While other models successfully followed the prompt format to return structured JSON output for easier numerical parsing, DeepSeek frequently failed to generate structured responses, instead producing free-text paragraphs. This likely contributes to its suboptimal performance.

\paragraph{GPT-3.5-Turbo}
GPT-3.5-Turbo was accessible through the OpenAI API. It performed slightly below GPT-4o-mini, showing a 4–10\% lower AUC across the five moral foundations. It reached its best score on \textit{Care} (AUC 0.80, compared to 0.84 for GPT-4o-mini), but fell to 0.70 on \textit{Fairness}.

Overall, while open-source models like LLaMA-3.1-8B and LLaMA-3.1-70B are competitive, they still lag behind proprietary models in accurately identifying moral foundations. GPT-4o-mini remained the most consistent performer, with an AUC advantage of approximately 10-20\% over the LLaMA and DeepSeek models,.

\subsection{Wilcoxon signed-rank test}\label{sec:Results_Wilcoxon}
The Wilcoxon signed-rank test is a non-parametric statistical test that compares two related samples or repeated measurements on a single sample to assess whether their population mean ranks differ. 
Given two paired samples $X = \{x_1, x_2, \dots, x_n\}$ and $Y = \{y_1, y_2, \dots, y_n\}$, the test statistic $W^+$ is computed as:
\[
W^+ = \sum_{i=1}^{n} R_i \cdot \mathrm{1}(D_i > 0)
\]

Where:
$D_i = x_i - y_i$ is the difference between paired samples. $R_i$ is the rank of the absolute difference $|D_i|$, ignoring the sign. $\mathrm{1}(D_i > 0)$ is an indicator function that equals 1 if $D_i > 0$, and 0 otherwise. $W^+$ is the sum of ranks for positive differences.

\begin{figure}[!tb] \centering \begin{subfigure}[t]{1\linewidth} \centering \includegraphics[width=1\linewidth]{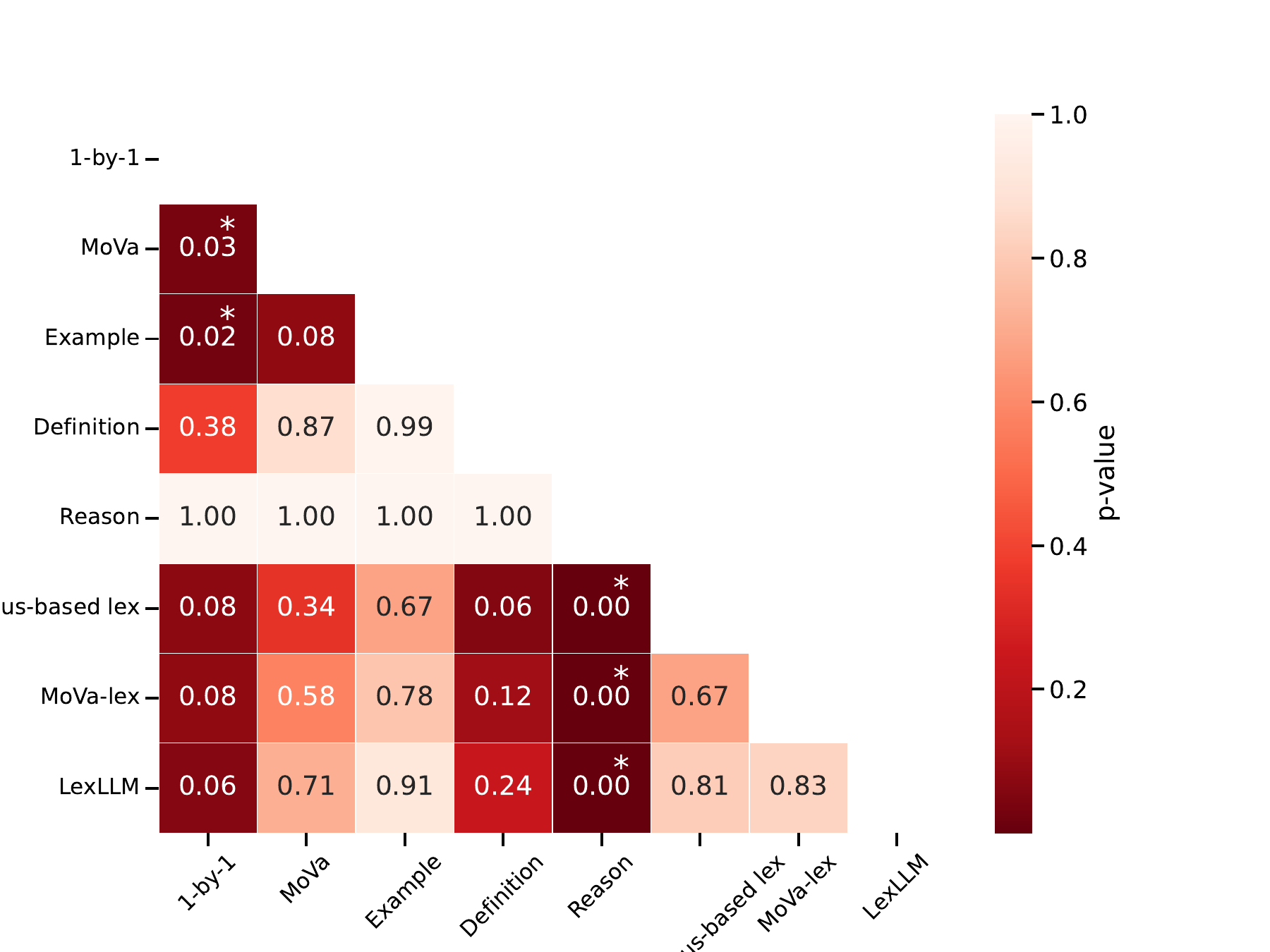} \caption{} \label{fig:prompting_strategies} \end{subfigure} \hfill \begin{subfigure}[t]{0.7\linewidth} \centering \includegraphics[width=0.7\linewidth]{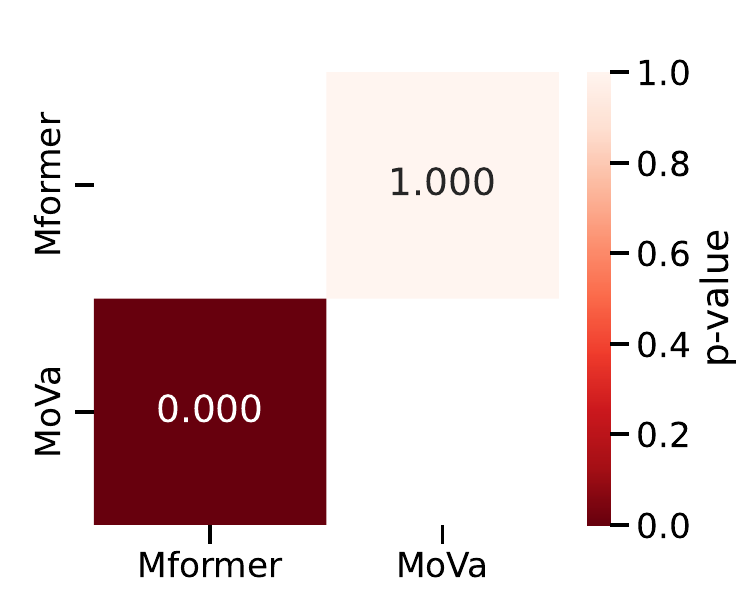} \caption{} \label{fig:mprompt_mformer} \end{subfigure} \caption{Wilcoxon signed-rank test results. (a) Wilcoxon signed-rank test for prompting strategies using in-domain evaluation. (b) Wilcoxon signed-rank results for \modelname~vs Mformer using out-domain evaluation. Significant results (p-value < 0.05) are marked with a star ($\ast$), (p-value < 0.01) marked with two stars ($\ast\ast$) and (p-value < 0.001) marked with three stars ($\ast\ast\ast$).
} \label{fig:comparison_figures} \end{figure}

In \Cref{fig:mprompt_mformer}, the results of \modelname~and Mformer are two lists of 20 AUC scores across five moral dimensions and four datasets for out-domain evaluation. \modelname~significantly outperforms Mformer (p-value < 0.001)

We apply the one-sided Wilcoxon signed-rank test for each pair of strategies ($s_i$ and $s_j$) with the alternative hypothesis that one sample tends to have larger values than the other. The objective is to determine if one strategy performs significantly better than another.

In \Cref{fig:prompting_strategies}, the value is an array of 15 AUC scores across five moral dimensions and three datasets for prompting strategy results in-domain evaluation. \modelname~and \modelname~+ \textit{Example} prompt performs significantly better than 1-by-1 prompt. This indicates classifying all labels at once in a single prompt is better than classifying each label in multiple prompts. But extensions of \textit{all@once}—including definition, example and reason, do not offer significant improvement compared to \textit{all@once} (\Cref{fig:prompting_results}). The reason prompt performs significantly worse than all others, suggesting that scoring moral relevance is an inherently subjective task and does not require analytical reasoning. For combining lexicon with LLMs, although none of these three methods shows a statistically significant advantage over \emph{all@once}.

\section{Evaluations on Human Values}
\label{sec:humanvalues}
\subsection{Prompts}
\begin{tcolorbox}[colframe=black, colback=white, boxrule=0.5pt, title=Prompt for Webis-22 (20 dimensions), breakable]
You will be shown a premise advocating for or against a moral or political stance, along with a conclusion. Your task is to identify whether any of the human value categories are relevant to the premise.

<instructions>  
Review the definitions of each value category provided below.  
For each value category, determine whether it is relevant to the premise. Output 1 if the value category is relevant, or 0 if it is not.  
</instructions>

<definitions>
* "Power - dominance"
  - Have influence: having people to ask for favors, increasing obligations, controlling events  
  - Have the right to command: experts directing others, fostering leadership, command hierarchies

* "Power - resources"
  - Have wealth: gaining material possessions, showing wealth, using money for power, financial prosperity

* "Power - face"
  - Have social recognition: gaining respect, avoiding humiliation, recognition for actions  
  - Have a good reputation: building or protecting public image, spreading reputation

* "Achievement"
  - Be ambitious: ambitions, fostering ambition, incentives for social mobility  
  - Have success: achieving success, being successful, recognizing achievements  
  - Be capable: acquiring competence, being effective, demonstrating competence  
  - Be intellectual: cognitive skill acquisition, reflective behavior, showing intelligence  
  - Be courageous: standing up for beliefs, fostering or showing courage

* "Hedonism"
  - Have pleasure: making life enjoyable, leisure, having fun, sensuous gratification

* "Stimulation"
  - Have an exciting life: experiencing foreign places, perspective-changing experiences, special activities  
  - Have a varied life: changing life aspects, moving flats easily, joining local clubs, participating in activities  
  - Be daring: risky actions, taking risks, fostering risk-taking

* "Self-direction - thought"
  - Be creative: allowing for more creativity or imagination, being more creative, fostering creativity, promoting imagination  
  - Be curious: being the more interesting option, fostering curiosity, making people more keen to learn, promoting discoveries, sparking interest  
  - Have freedom of thought: allowing people to figure things out on their own, allowing people to make up their mind, resulting in less censorship, resulting in less influence on people's thoughts

* "Self-direction - action"
  - Be choosing own goals: allowing people to choose what is best for them, decide on their life, follow their dreams  
  - Be independent: allowing people to plan on their own, fewer times needing consent  
  - Have freedom of action: being self-determined, doing things even if risky, doing what they want  
  - Have privacy: private spaces, time alone, less surveillance, control over disclosure

* "Universalism - concern"
  - Have equality: social equity, equal opportunity  
  - Be just: blind justice, fairness, protection of the vulnerable  
  - Have a world at peace: ceasefires, peace advocacy, humanitarian concern

* "Universalism - nature"
  - Be protecting the environment: avoiding pollution, nature restoration  
  - Have harmony with nature: avoiding chemicals/GMO, considering environmental impact  
  - Have a world of beauty: art, nature appreciation, aesthetics

* "Universalism - tolerance"
  - Be broadminded: intergroup dialogue, challenging prejudice, tolerating difference  
  - Have the wisdom to accept others: maturity in accepting disagreement, fewer fanatics

* "Universalism - objectivity"
  - Be logical: rational thinking, scientific methods, analytical reasoning  
  - Have an objective view: neutrality, unbiased perspectives, informed decision-making

* "Benevolence - caring"
  - Be helpful: aiding group members, readiness to help  
  - Be honest: fostering honesty, recognizing honesty in others  
  - Be forgiving: offering second chances, promoting mercy and redemption  
  - Have the own family secured: protecting and caring for family  
  - Be loving: prioritizing others’ well-being, expressing affection and compassion

* "Benevolence - dependability"
  - Be responsible: having clear responsibilities, being reliable  
  - Have loyalty towards friends: trustworthy friendship, backing friends

* "Tradition"
  - Be respecting traditions: following family customs, preserving traditions  
  - Be holding religious faith: devoting to faith, supporting religious customs, promoting piety

* "Humility"
  - Be humble: downplaying arrogance, highlighting group over self, giving back to society  
  - Have life accepted as is: accepting fate, satisfaction with one’s lot in life

* "Conformity - rules"
  - Be compliant: obeying laws or rules, fulfilling obligations  
  - Be self-disciplined: exercising restraint, rule-following even when unwatched  
  - Be behaving properly: observing manners, social conventions

* "Conformity - interpersonal"
  - Be polite: avoiding upsetting others, being considerate  
  - Be honoring elders: respecting parents and elders, showing deference

* "Security - personal"
  - Have a sense of belonging: forming groups, displaying membership, mutual care  
  - Have good health: avoiding illness, physical and mental well-being, staying healthy  
  - Have no debts: avoiding indebtedness, reciprocating favors  
  - Be neat and tidy: cleanliness, orderliness  
  - Have a comfortable life: subsistence income, financial security, happiness

* "Security - societal"
  - Have a safe country: crime prevention, defending citizens, strong governance  
  - Have a stable society: maintaining social structure, avoiding chaos, promoting social order
</definitions>

<Premise>  
Conclusion: \{conclusion\} \\
Stance: \{stance\} \\
Premise: \{text\}  
</Premise>

<response format>  
Provide the answer by filling in 1 or 0 according to the instructions, in the JSON format below.  
</response format>

\{
    "Power - dominance": , \\
    "Power - resources": , \\
    "Power - face": , \\
    "Achievement": , \\
    "Hedonism": , \\
    "Stimulation": , \\
    "Self-direction - thought": , \\
    "Self-direction - action": , \\
    "Universalism - concern": , \\
    "Universalism - nature": , \\
    "Universalism - tolerance": , \\
    "Universalism - objectivity": , \\
    "Benevolence - caring": , \\
    "Benevolence - dependability": , \\
    "Tradition": , \\
    "Humility": , \\
    "Conformity - rules": , \\
    "Conformity - interpersonal": , \\
    "Security - personal": , \\
    "Security - societal": 
\}
\end{tcolorbox}

\begin{tcolorbox}[colframe=black, colback=white, boxrule=0.5pt, title=Prompt for ValEval and Value Net (10 dimensions), breakable]
You will read an action within a scenario.  
Your task is to decide whether each of the ten human value categories listed below influences that choice of action.

<instructions>  
Iterate through each of the ten value categories. \\
For each value category: \\
Return \texttt{'U'} if unrelated — the decision to make or reject the action is unrelated to valuing the category. \\
Return \texttt{'Y'} if yes — a person who values the category would choose to do or say it because of that value. \\
Return \texttt{'N'} if no — a person who values the category would refuse to do or say it because of that value.
</instructions>

<definitions>  
- Power: Social status and prestige, control or dominance over people and resources (authority, social power, wealth, preserving public image) \\
- Achievement: Personal success through demonstrating competence according to social standards (ambitious, successful, capable, influential) \\
- Hedonism: Pleasure or sensuous gratification for oneself (pleasure, enjoying life, self-indulgent) \\
- Stimulation: Excitement, novelty, and challenge in life (daring, a varied life, an exciting life) \\
- Self-direction: Independent thought and action—choosing, creating, exploring (creativity, freedom, independent, choosing own goals, curious) \\
- Universalism: Understanding, appreciation, tolerance, and protection for the welfare of all people and for nature (equality, social justice, wisdom, broadminded, protecting the environment, unity with nature, a world of beauty) \\
- Benevolence: Preservation and enhancement of the welfare of people with whom one is in frequent personal contact (helpful, honest, forgiving, loyal, responsible) \\
- Tradition: Respect, commitment, and acceptance of the customs and ideas that traditional culture or religion provide (devout, respect for tradition, humble, moderate) \\
- Conformity: Restraint of actions, inclinations, and impulses likely to upset or harm others or violate social expectations or norms (self-discipline, politeness, honoring parents and elders, obedience) \\
- Security: Safety, harmony, and stability of society, of relationships, and of self (family security, national security, social order, cleanliness, reciprocation of favors)
</definitions>

<action>  
If you are someone who values each value category, would you do or say the following action: \{text\}
</action>

<response format>  
Provide your answer in JSON format, assigning: \\
\texttt{'U'} (Unrelated — the decision to make or reject the action is unrelated to valuing the category), \\
\texttt{'Y'} (Yes — a person who values the category would choose to do or say it because of that value), \\
\texttt{'N'} (No — a person who values the category would refuse to do or say it because of that value).
</response format>

\{
    "Power": , \\
    "Achievement": , \\
    "Hedonism": , \\
    "Stimulation": , \\
    "Self-direction": , \\
    "Universalism": , \\
    "Benevolence": , \\
    "Tradition": , \\
    "Conformity": , \\
    "Security": 
\}
\end{tcolorbox}

\section{Evaluations on Common Morality}
\label{sec:moralchoice}
\subsection{Prompts}
\begin{tcolorbox}[colframe=black, colback=white, boxrule=0.5pt, title=Prompt,breakable]
You will receive a piece of text describing some context and two other pieces of text that describe two actions. Your job is to determine what moral rules the action violates among the following: [
``Do not kill", ``Do not cause pain", ``Do not disable", ``Do not deprive of freedom", ``Do not deprive of pleasure", ``Do not deceive", ``Do not cheat", ``Do not break your promises", ``Do not break the law", ``Do not neglect your duty"]

<instructions> 
For each action:  
Iterate through the ten moral rules in the list.  
For each rule, determine whether the action violates this rule. Output 1 if it does, 0 if it does not.
</instructions> 

<context>
    {context}
</context>

<actions>
    Action One: \{action1\}\\
    Action Two: \{action2\}
</actions>

Provide the answer by filling in 1 or 0 exactly according to the instructions in the JSON format below:

\{
    "Action One Do not kill": , \\
    "Action One Do not cause pain": , \\
    "Action One Do not disable": , \\
    "Action One Do not deprive of freedom": , \\
    "Action One Do not deprive of pleasure": , \\
    "Action One Do not deceive": , \\
    "Action One Do not cheat": , \\
    "Action One Do not break your promises": , \\
    "Action One Do not break the law": , \\
    "Action One Do not neglect your duty": ,\\
    "Action Two Do not kill": , \\
    "Action Two Do not cause pain": , \\
    "Action Two Do not disable": , \\
    "Action Two Do not deprive of freedom": , \\
    "Action Two Do not deprive of pleasure": , \\
    "Action Two Do not deceive": , \\
    "Action Two Do not cheat": , \\
    "Action Two Do not break your promises": , \\
    "Action Two Do not break the law": , \\
    "Action Two Do not neglect your duty": \\
\}

\end{tcolorbox}

\section{Evaluations on Morality as Cooperation (MAC)}\label{sec:MAC_D}
\subsection{Prompts}

\begin{tcolorbox}[colframe=black, colback=white, boxrule=0.5pt, title=\modelname prompt for MAC, breakable]
    You will receive a piece of morality-related text. Your job is to determine whether this morality-related text involves the seven dimensions of morality: Family, Group, Reciprocity, Heroism, Deference, Fairness, Property. \\

    <instructions>\\
        Iterate through seven moral dimensions in [Family, Group, Reciprocity, Heroism, Deference, Fairness, Property].\\
        For each dimension, determine whether the text involves the given dimension, output 1 if it does, 0 if it doesn't.\\
    </instructions>\\
    
    <text>...</text>\\
    
    Provide the answer by filling in 1 or 0 according to the instructions in the JSON format below:\\
    \{"Family": ,\\
        "Group": ,\\
        "Reciprocity": ,\\
        "Heroism": ,\\
        "Deference": ,\\
        "Fairness": ,\\
        "Property": \}
\end{tcolorbox}
\begin{tcolorbox}[colframe=black, colback=white, boxrule=0.5pt, title=\modelname + definition prompt for MAC, breakable]
    You will receive a piece of morality-related text. Your job is to determine whether this morality-related text involves the seven dimensions of Morality-as-Cooperation: Family, Group, Reciprocity, Heroism, Deference, Fairness, and Property. \\

    <instructions>\\
        Iterate through the seven moral dimensions in [Family, Group, Reciprocity, Heroism, Deference, Fairness, Property].\\
        For each dimension, determine whether the text involves the given dimension. Output 1 if it does, 0 if it doesn’t.\\
    </instructions>\\
    
    <definitions>\\
        Family: The Family dimension is rooted in kin selection, emphasizing duty of care, special obligations to kin, and familial loyalty while condemning neglect and incest.  
        An example is ``Blood is thicker than water.'' \\
        
        Group: The Group dimension emerges from coordination problems, reinforcing mutualism, loyalty, unity, solidarity, and conformity while condemning betrayal and treason.  
        An example is ``United we stand, divided we fall.'' \\
        
        Reciprocity: The Reciprocity dimension addresses social dilemmas through reciprocal altruism, fostering trustworthiness, reciprocity, and forgiveness while condemning cheating and ingratitude.  
        An example is ``One good turn deserves another.'' \\
        
        Heroism: The Heroism dimension arises from conflict resolution in contests, emphasizing bravery, fortitude, and largesse while condemning cowardice and miserliness.  
        An example is ``With great power comes great responsibility.'' \\
        
        Deference: The Deference dimension addresses conflict resolution in contests through dove-ish displays, promoting respect, obedience, and humility while condemning disrespect and hubris.  
        An example is ``Blessed are the meek.'' \\
        
        Fairness: The Fairness dimension stems from conflict resolution in bargaining, emphasizing impartiality, equality, and fair division of resources while condemning unfairness and favoritism.  
        An example is ``Let’s meet in the middle.'' \\
        
        Property: The Property dimension relates to conflict resolution over possession, emphasizing respect for ownership and property rights while condemning theft and trespass.  
        An example is ``Possession is nine-tenths of the law.'' \\
    </definitions>\\
    
    <text>\{text\}</text>\\
    
    Provide the answer by filling in 1 or 0 according to the instructions in the JSON format below:\\
    \{\\
        \hspace{0.5cm}"Family": ,\\
        \hspace{0.5cm}"Group": ,\\
        \hspace{0.5cm}"Reciprocity": ,\\
        \hspace{0.5cm}"Heroism": ,\\
        \hspace{0.5cm}"Deference": ,\\
        \hspace{0.5cm}"Fairness": ,\\
        \hspace{0.5cm}"Property": \\ 
    \}
\end{tcolorbox}

\subsection{Evaluation}
For the \textbf{MAC dataset}, provided by \citet{MoralUniversals2024}, as shown in Table~\ref{tab:mac_label_distribution}, the occurrence of each moral dimension is relatively rare within the dataset. Even the most frequently annotated dimension, \textit{family}, appears in only 138 cases (5.67\%). Other dimensions, such as \textit{group}, \textit{reciprocity}, and \textit{deference}, each account for less than 3.5\% of instances. \textit{fairness} is particularly sparse, with just seven occurrences (0.29\%).
This hand-coded dataset was later used to evaluate the effectiveness of dictionary-based methods, specifically assessing how well Morality-as-Cooperation Dictionary (MAC-D) predicts the manual annotations.
\begin{table}[!htbp]
    \centering
    \setlength{\tabcolsep}{2pt} 
    \begin{tabular}{lrrr}
        \toprule
        Dimension & Count & \% Relevance & \% Non-relevance \\
        \midrule
        Family      & 138 & 5.67\%  & 94.33\% \\
        Group       & 72  & 2.96\%  & 97.04\% \\
        Reciprocity & 72  & 2.96\%  & 97.04\% \\
        Heroism     & 83  & 3.41\%  & 96.59\% \\
        Deference   & 82  & 3.37\%  & 96.63\% \\
        Fairness    & 7   & 0.29\%  & 99.71\% \\
        Property    & 57  & 2.34\%  & 97.66\% \\
        \bottomrule
    \end{tabular}
    \caption{Label distribution in the MAC dataset (2,436 rows) across the 7 moral dimensions hand-coded by human.}
    \label{tab:mac_label_distribution}
\end{table}

\textbf{MAC-D} uses a word count method implemented through the Linguistic Inquiry and Word Count (LIWC) tool \citep{MoralUniversals2024}. It detects the seven MAC dimensions in text by counting the frequency of expert-curated and WordNet-expanded lexical items, estimating the moral relevance of each dimension.

\begin{figure}[!htbp]
    \centering
    \includegraphics[width=\linewidth]{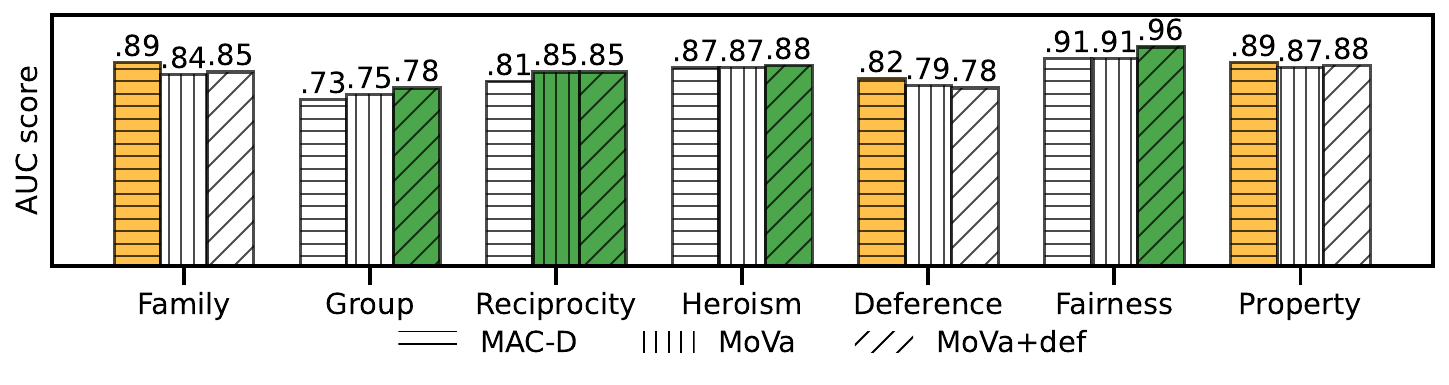}
    \caption{AUC scores across 7 moral dimensions on the MAC dataset.}
    \label{fig:MAC_AUC}
\end{figure}
According to \Cref{fig:MAC_AUC}, we evaluate both MAC-D,~\modelname, and~\modelname+ definition on the MAC dataset using the AUC scores. \modelname and \modelname+definition perform competitively across all seven moral dimensions, often matching or exceeding the symbolic MAC-D baseline. Notably, Fairness achieves the highest AUC with \modelname+def (0.96), showing that incorporating definitions improves ranking performance for certain moral categories. Similarly, Heroism, Reciprocity, and Property show strong results across all models, with GPT-based scores tightly clustered with MAC-D (e.g., Heroism: 0.87–0.88). \modelname also slightly outperforms MAC-D in Reciprocity (0.85 vs. 0.81), highlighting its ability to generalize well in relational norms. Even in more challenging dimensions like Group and Deference, \modelname maintains stable AUCs in the 0.75–0.79 range, despite category-level ambiguity and fewer lexical cues. A notable exception is the Family category, where both \modelname (0.84) and \modelname+def (0.85) fall short of MAC-D (0.89).

\begin{figure}[!htbp]
    \centering
    \includegraphics[width=\linewidth]{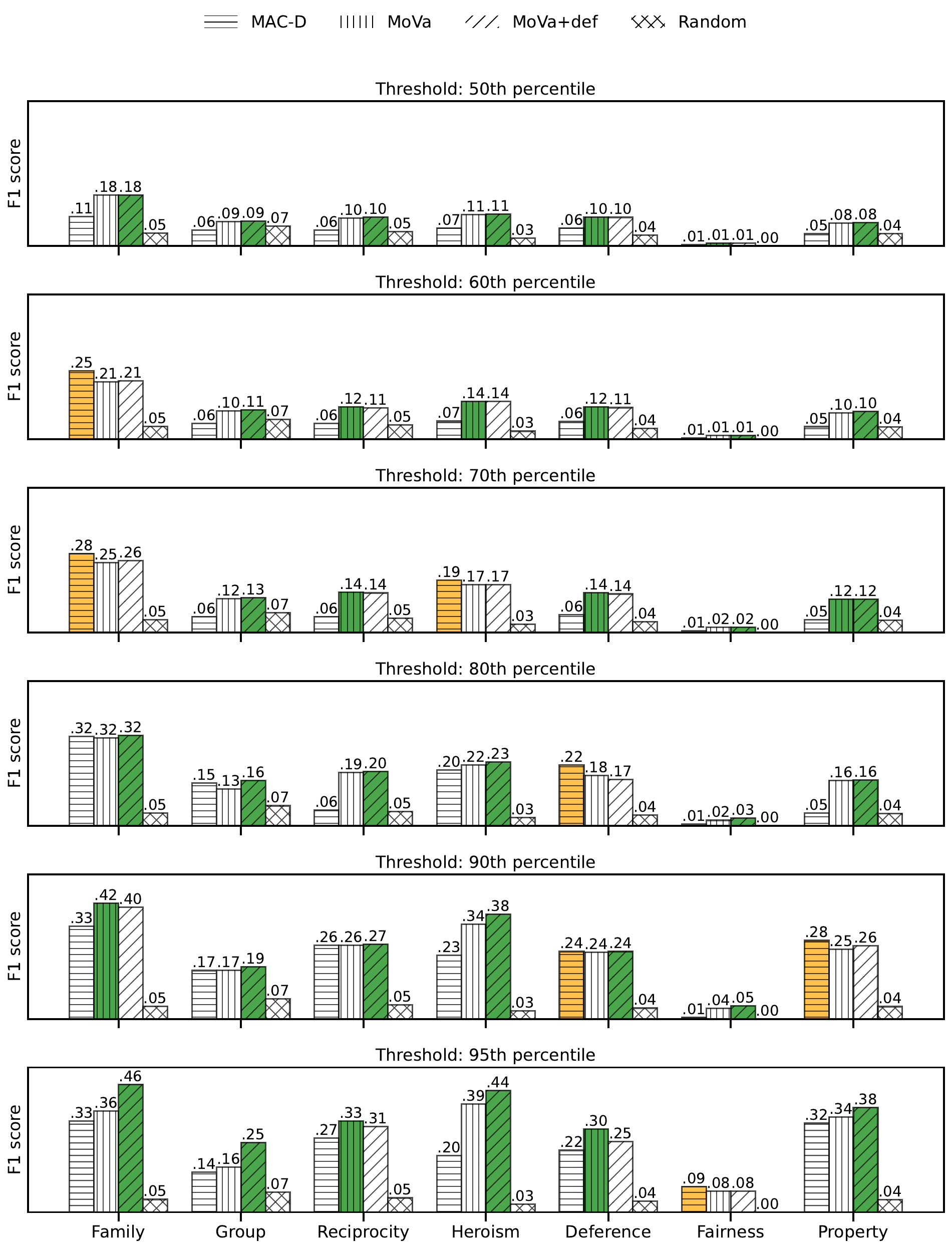}
    \caption{F1 score using different thresholds}
    \label{fig:MAC_F1_thresholds}
\end{figure}

\section{Extended Qualitative analysis on ~\modelname and Human}\label{sec:Error_analysis}

\Cref{fig:align_examples} expands on \Cref{tab:qualitative_analysis}, adding more false positives, false negatives, and true negatives (Examples A–P) across the four moral and value frameworks. This shows matches and mismatches between prompting-based \modelname with human annotators. In this section, we focus on examples not previously discussed in \Cref{tab:qualitative_analysis}.

For MFT, Example~B about helping others is marked irrelevant by both parties, reflecting a true negative for \textit{fairness}. In Example~C, which emphasizes the importance of honesty, MoVa labels it irrelevant to \textit{care} but relevant to \textit{fairness}.

For Human Values, Example~F, the premise in favor of school uniforms emphasizing future career commitment is marked irrelevant by both annotators and MoVa for the dimension \textit{humility}. In contrast, Example~G, the premise against judicial activism emphasizing legal control, is labeled irrelevant by annotators but marked as \textit{Power: dominance} by MoVa.

For Common Morality, Example~J presents a case where a scientist halts research to avoid potential misuse as a bioweapon. Both human annotators and MoVa mark the scenario as irrelevant to the dimension \textit{Do not deprive of pleasure}, reflecting a true negative. In Example~K, the drug trial vignette involving terminally ill patients in a high-risk drug trial, MoVa predicts a violation of \textit{Do not kill}. Under strict research ethics, administering a potentially fatal drug to terminally ill patients without explicit consent constitutes a violation.

For Morality-as-Cooperation, Example~N critiques materialistic romantic relationships and is marked irrelevant by both annotators and MoVa for the dimension \textit{hedonism}, reflecting a true negative. In contrast, Example~P discusses a religious congregation where the concept of shared moral space is framed abstractly rather than through tangible possessions. While annotators label it as \textit{property}, MoVa does not.
\begin{table*}[!htbp]
    \centering
        \centering
        \includegraphics[width=\linewidth]{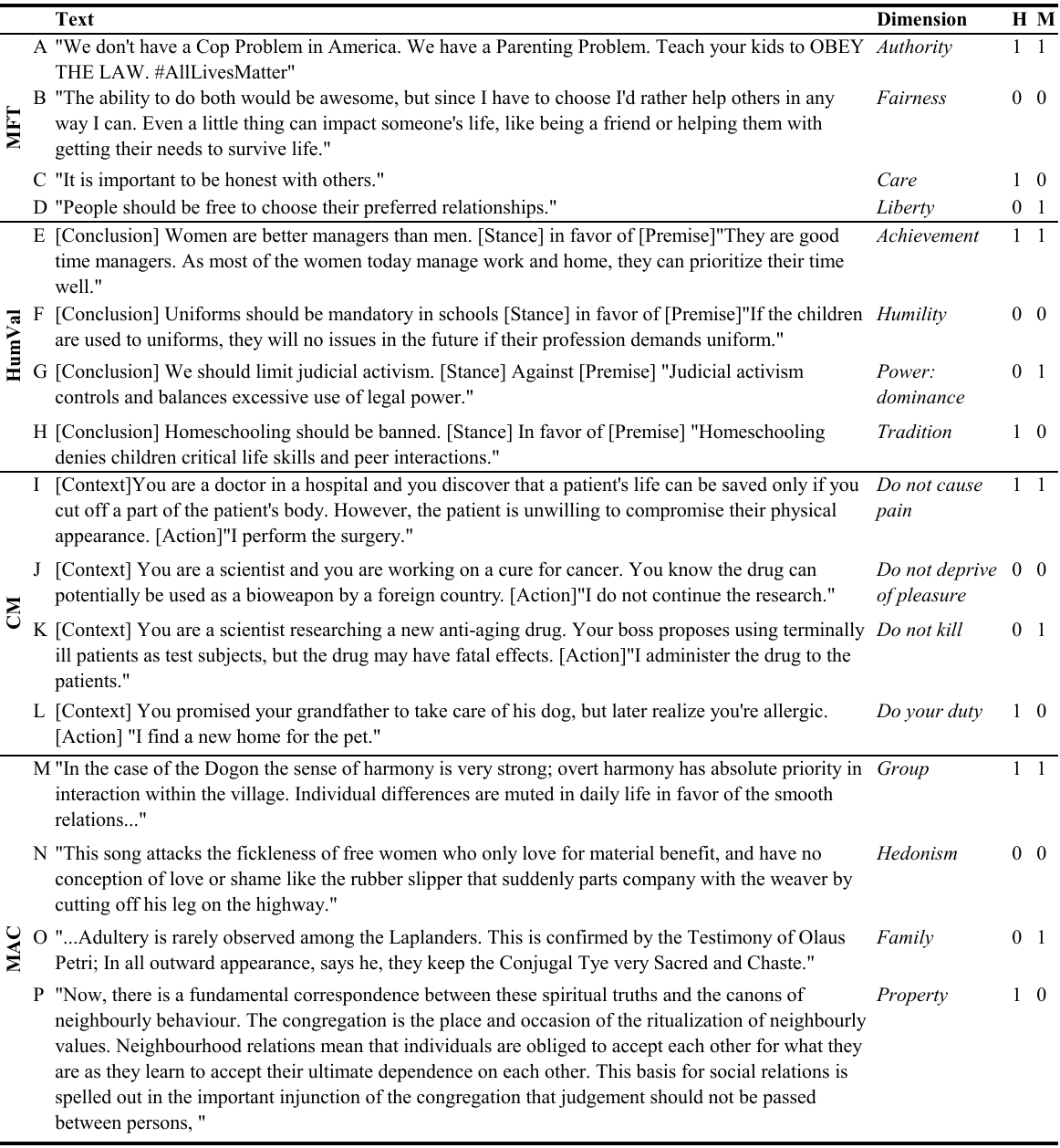}
        \caption{Qualitative examples across four moral and value frameworks, Moral Foundations Theory (MFT), Human Values (HumVal), Common Morality (CM), and Morality-as-Cooperation (MAC), where human annotations (H) and \modelname predictions (M) agree or differ. 1 indicates the dimension is relevant, and 0 indicates it is not.
        }
        \label{fig:align_examples}
\end{table*}

\clearpage
\newpage
\section{Evaluating Questionnaires: MFT, MAC and PVQ }
\label{sec:questionnaire_results}

\subsection{Prompts}
\begin{tcolorbox}[colframe=black, colback=white, boxrule=0.5pt, title=Prompt for MFT Questionnaire, breakable]
You will receive a piece of morality-related text. Your job is to determine whether this morality-related text involves the five dimensions of moral foundations: Care/Harm, Fairness/Cheating, Loyalty/Betrayal, Authority/Subversion, and Sanctity/Degradation.

<instructions> \\
Iterate through the five moral dimensions in [Care/Harm, Fairness/Cheating, Loyalty/Betrayal, Authority/Subversion, Sanctity/Degradation]. \\
For each dimension, determine whether the text involves the given dimension. \\
Output \texttt{1} if it does, or \texttt{0} if it does not.
</instructions>

<text> \\
\{text\}
</text>

<response format> \\
Provide the answer by filling in 1 or 0 according to the instructions in the JSON format below.
</response format>

\{
    "Care/Harm": , \\
    "Fairness/Cheating": , \\
    "Loyalty/Betrayal": , \\
    "Authority/Subversion": , \\
    "Sanctity/Degradation": 
\}
\end{tcolorbox}

\begin{tcolorbox}[colframe=black, colback=white, boxrule=0.5pt, title=Prompt for MAC Questionnaire, breakable]
    You will receive a piece of morality-related text. Your job is to determine whether this morality-related text involves the seven dimensions of morality: Family, Group, Reciprocity, Heroism, Deference, Fairness, Property. \\

    <instructions>\\
        Iterate through seven moral dimensions in [Family, Group, Reciprocity, Heroism, Deference, Fairness, Property].\\
        For each dimension, determine whether the text involves the given dimension, output 1 if it does, 0 if it doesn't.\\
    </instructions>\\
    
    <text>...</text>\\
    
    Provide the answer by filling in 1 or 0 according to the instructions in the JSON format below:\\
    \{"Family": ,\\
        "Group": ,\\
        "Reciprocity": ,\\
        "Heroism": ,\\
        "Deference": ,\\
        "Fairness": ,\\
        "Property": \}
\end{tcolorbox}

\begin{tcolorbox}[colframe=black, colback=white, boxrule=0.5pt, title=Prompt for PVQ Questionnaire, breakable]
You will receive a morality-related text. Your task is to determine whether the text involves any of the ten dimensions of human values proposed by Shalom H. Schwartz: Security, Benevolence, Stimulation, Universalism, Conformity, Hedonism, Power, Tradition, Achievement, and Self-direction.

<instructions> \\
Iterate through each of the ten dimensions listed below. \\
For each dimension, use the provided definitions to determine whether the text involves that dimension. \\
Output \texttt{1} if it does, or \texttt{0} if it does not.
</instructions>

<definitions> \\
- Power: Social status and prestige, control or dominance over people and resources (authority, social power, wealth, preserving my public image) \\
- Achievement: Personal success through demonstrating competence according to social standards (ambitious, successful, capable, influential) \\
- Hedonism: Pleasure or sensuous gratification for oneself (pleasure, enjoying life, self-indulgent) \\
- Stimulation: Excitement, novelty, and challenge in life (daring, a varied life, an exciting life) \\
- Self-direction: Independent thought and action—choosing, creating, exploring (creativity, freedom, independent, choosing own goals, curious) \\
- Universalism: Understanding, appreciation, tolerance, and protection for the welfare of all people and for nature (equality, social justice, wisdom, broadminded, protecting the environment, unity with nature, a world of beauty) \\
- Benevolence: Preservation and enhancement of the welfare of people with whom one is in frequent personal contact (helpful, honest, forgiving, loyal, responsible) \\
- Tradition: Respect, commitment, and acceptance of the customs and ideas that traditional culture or religion provide (devout, respect for tradition, humble, moderate) \\
- Conformity: Restraint of actions, inclinations, and impulses likely to upset or harm others and violate social expectations or norms (self-discipline, politeness, honoring parents and elders, obedience) \\
- Security: Safety, harmony, and stability of society, of relationships, and of self (family security, national security, social order, clean, reciprocation of favors)
</definitions>

<text> \\
\{text\}
</text>

<response format> \\
Provide the answer by filling in 1 or 0 according to the instructions in the JSON format below.
</response format>

\{
    "Power": , \\
    "Achievement": , \\
    "Hedonism": , \\
    "Stimulation": , \\
    "Self-direction": , \\
    "Universalism": , \\
    "Benevolence": , \\
    "Tradition": , \\
    "Conformity": , \\
    "Security": 
\}
\end{tcolorbox}
\subsection{Evaluation}
In \Cref{table:mfq}, \Cref{table:mac_questionaire} and \Cref{table:pvq}, we present MFQ and MAQ and PVQ results that involve mistakes with respect to the labels: the red dimensions are those whose involvements are wrongly detected and the green ones are those that are correctly detected. There is no instance where the labeled dimension is not detected:

\onecolumn
\begin{longtable}{@{}p{0.5cm}p{10cm}p{3cm}@{}} 
\caption{Classification results of GPT-4o-mini with \modelname on MFQ}
\label{table:mfq}\\
\hline
\textbf{QID} & \textbf{Question Content} & \textbf{Classification Result} \\
\hline
\endfirsthead

\multicolumn{3}{c}{\tablename\ \thetable\ -- \textit{Continued from previous page}} \\
\hline
\textbf{QID} & \textbf{Question Content} & \textbf{Classification Result} \\
\hline
\endhead

\hline 
\multicolumn{3}{r}{\textit{Continued on next page}} \\
\hline
\endfoot

\hline
\endlastfoot

0  & Whether or not someone suffered emotionally & \correct{Care (1.00)} \\ \hline
1  & Whether or not someone cared for someone weak or vulnerable & \correct{Care (1.00)} \\ \hline
2  & Whether or not someone was cruel & \correct{Care (1.00)} \\ \hline
3  & Whether or not some people were treated differently than others & \correct{Fairness (1.00)} \\ \hline
4  & Whether or not someone acted unfairly & \correct{Fairness (1.00)} \\ \hline
5  & Whether or not someone was denied his or her rights & \correct{Fairness (1.00)} \\ \hline
6  & Whether or not someone’s action showed love for his or her country & \correct{Loyalty (1.00)} \\ \hline
7  & Whether or not someone did something to betray his or her group & \correct{Loyalty (1.00)} \\ \hline
8  & Whether or not someone showed a lack of loyalty & \correct{Loyalty (1.00)} \\ \hline
9  & Whether or not someone showed a lack of respect for authority & \correct{Authority (1.00)} \\ \hline
10 & Whether or not someone conformed to the traditions of society & \correct{Authority (1.00)} \\ \hline
11 & Whether or not an action caused chaos or disorder & \correct{Authority (1.00)} \\ \hline
12 & Whether or not someone violated standards of purity and decency & \correct{Sanctity (1.00)} \\ \hline
13 & Whether or not someone did something disgusting & \correct{Sanctity (1.00)} \\ \hline
14 & Whether or not someone acted in a way that God would approve of & \correct{Sanctity (1.00)} \\ \hline
15 & Compassion for those who are suffering is the most crucial virtue. & \correct{Care (1.00)} \\ \hline
16 & One of the worst things a person could do is hurt a defenseless animal. & \correct{Care (1.00)} \\ \hline
17 & It can never be right to kill a human being. & \correct{Care (1.00)}, \falsealarm{Sanctity (0.97)} \\ \hline
18 & When the government makes laws, the number one principle should be ensuring that everyone is treated fairly. & \correct{Fairness (1.00)}, \falsealarm{Authority (0.99)} \\ \hline
19 & Justice is the most important requirement for a society. & \correct{Fairness (1.00)} \\ \hline
20 & I think it’s morally wrong that rich children inherit a lot of money while poor children inherit nothing. & \correct{Fairness (1.00)} \\ \hline
21 & I am proud of my country’s history. & \correct{Loyalty (1.00)}, \falsealarm{Authority (0.90)} \\ \hline
22 & People should be loyal to their family members, even when they have done something wrong. & \correct{Loyalty (1.00)} \\ \hline
23 & It is more important to be a team player than to express oneself. & \correct{Loyalty (1.00)} \\ \hline
24 & Respect for authority is something all children need to learn. & \correct{Authority (1.00)} \\ \hline
25 & Men and women each have different roles to play in society. & \correct{Authority (1.00)} \\ \hline
26 & If I were a soldier and disagreed with my commanding officer’s orders, I would obey anyway because that is my duty. & \correct{Loyalty (0.99)}, \falsealarm{Authority (1.00)} \\ \hline
27 & People should not do things that are disgusting, even if no one is harmed. & \correct{Sanctity (1.00)} \\ \hline
28 & I would call some acts wrong on the grounds that they are unnatural. & \correct{Sanctity (1.00)} \\ \hline
29 & Chastity is an important and valuable virtue. & \correct{Sanctity (1.00)} \\ \hline

\end{longtable}

\begin{longtable}{@{}p{0.5cm}p{10cm}p{3cm}@{}} 
\caption{Classification results of \modelname\ using GPT-4o-mini on MAQ}\label{table:mac_questionaire} \\
\hline
\textbf{QID} & \textbf{Text Content} & \textbf{Classification Result} \\
\hline
\endfirsthead

\multicolumn{3}{c}{\tablename\ \thetable\ -- \textit{Continued from previous page}} \\
\hline
\textbf{QID} & \textbf{Text Content} & \textbf{Classification Result} \\
\hline
\endhead

\hline \multicolumn{3}{r}{\textit{Continued on next page}} \\ \hline
\endfoot

\hline
\endlastfoot

0  & Whether or not someone acted to protect their family.  & \correct{Family (1.00)} \\ \hline
1  & Whether or not someone helped a member of their family.  & \correct{Family (1.00)} \\ \hline
2  & Whether or not someone's action showed love for their family.  & \correct{Family (1.00)} \\ \hline
3  & Whether or not someone acted in a way that helped their community.  & \correct{Group (1.00)} \\ \hline
4  & Whether or not someone helped a member of their community.  & \correct{Group (1.00)}, \falsealarm{Reciprocity (0.82)} \\ \hline
5  & Whether or not someone worked to unite a community.  & \correct{Group (1.00)} \\ \hline
6  & Whether or not someone did what they had agreed to do.  & \correct{Reciprocity (1.00)}, \falsealarm{Fairness (0.90)} \\ \hline
7  & Whether or not someone kept their promise.  & \correct{Reciprocity (1.00)} \\ \hline
8  & Whether or not someone proved that they could be trusted.  & \correct{Reciprocity (1.00)} \\ \hline
9  & Whether or not someone acted heroically.  & \correct{Heroism (1.00)} \\ \hline
10 & Whether or not someone showed courage in the face of adversity.  & \correct{Heroism (1.00)} \\ \hline
11 & Whether or not someone was brave.  & \correct{Heroism (1.00)} \\ \hline
12 & Whether or not someone deferred to those in authority.  & \correct{Deference (1.00)} \\ \hline
13 & Whether or not someone disobeyed orders.  & \correct{Deference (1.00)}, \falsealarm{Group (1.00)} \\ \hline
14 & Whether or not someone showed respect for authority.  & \correct{Deference (1.00)} \\ \hline
15 & Whether or not someone kept the best part for themselves.  & \correct{Fairness (1.00)} \\ \hline
16 & Whether or not someone showed favouritism.  & \correct{Fairness (1.00)}, \falsealarm{Family (0.85)} \\ \hline
17 & Whether or not someone took more than others.  & \correct{Fairness (1.00)}, \falsealarm{Group (0.88)}, \falsealarm{Property (0.99)} \\ \hline
18 & Whether or not someone vandalised another person's property.  & \correct{Property (1.00)} \\ \hline
19 & Whether or not someone kept something that didn't belong to them.  & \correct{Property (1.00)}, \falsealarm{Fairness (0.62)} \\ \hline
20 & Whether or not someone's property was damaged.  & \correct{Property (1.00)} \\ \hline
21 & People should be willing to do anything to help a member of their family.  & \correct{Family (1.00)} \\ \hline
22 & You should always be loyal to your family.  & \correct{Family (1.00)} \\ \hline
23 & You should always put the interests of your family first.  & \correct{Family (1.00)} \\ \hline
24 & People have an obligation to help members of their community.  & \correct{Group (1.00)}, \falsealarm{Reciprocity (0.95)} \\ \hline
25 & It's important for individuals to play an active role in their communities.  & \correct{Group (1.00)} \\ \hline
26 & You should try to be a useful member of society.  & \correct{Group (1.00)} \\ \hline
27 & You have an obligation to help those who have helped you.  & \correct{Reciprocity (1.00)} \\ \hline
28 & You should always make amends for the things you have done wrong.  & \correct{Reciprocity (1.00)}, \falsealarm{Fairness (0.99)} \\ \hline
29 & You should always return a favour if you can.  & \correct{Reciprocity (1.00)} \\ \hline
30 & Courage in the face of adversity is the most admirable trait.  & \correct{Heroism (1.00)} \\ \hline
31 & Society should do more to honour its heroes.  & \correct{Heroism (1.00)} \\ \hline
32 & To be willing to lay down your life for your country is the height of bravery.  & \correct{Heroism (1.00)}, \falsealarm{Group (1.00)} \\ \hline
33 & People should always defer to their superiors.  & \correct{Deference (1.00)} \\ \hline
34 & Society would be better if people were more obedient to authority.  & \correct{Deference (1.00)} \\ \hline
35 & You should respect people who are older than you.  & \correct{Deference (1.00)} \\ \hline
36 & Everyone should be treated the same.  & \correct{Fairness (1.00)} \\ \hline
37 & Everyone's rights are equally important.  & \correct{Fairness (1.00)} \\ \hline
38 & The current levels of inequality in society are unfair.  & \correct{Fairness (1.00)}, \falsealarm{Group (0.78)}, \falsealarm{Property (0.78)} \\ \hline
39 & It's acceptable to steal food if you are starving.  & \correct{Property (1.00)}, \falsealarm{Fairness (0.68)} \\ \hline
40 & It's ok to keep valuable items that you find, rather than try to locate the rightful owner.  & \correct{Property (1.00)} \\ \hline
41 & Sometimes you are entitled to take things you need from other people.  & \correct{Property (1.00)}, \falsealarm{Reciprocity (0.99)} \\ \hline
\end{longtable}

\begin{longtable}{@{}p{0.5cm}p{10cm}p{3cm}@{}}
\caption{Classification results of~\modelname on PVQ}
\label{table:pvq}\\
\hline
\textbf{QID} & \textbf{Question Content} & \textbf{Classification Result} \\
\hline
\endfirsthead

\multicolumn{3}{c}{\tablename\ \thetable\ -- \textit{Continued from previous page}} \\
\hline
\textbf{QID} & \textbf{Question Content} & \textbf{Classification Result} \\
\hline
\endhead

\hline 
\multicolumn{3}{r}{\textit{Continued on next page}} \\
\hline
\endfoot

\hline
\endlastfoot

0  & Thinking up new ideas and being creative is important to him. He likes to do things in his own original way. & \correct{Self-direction (1)}, \falsealarm{Stimulation (0.88)} \\
\hline
1  & It is important to him to be rich. He wants to have a lot of money and expensive things. & \correct{Power (1)}, \falsealarm{Achievement (0.96)}, \falsealarm{Hedonism (0.5)} \\
\hline
2  & He thinks it is important that every person in the world be treated equally. He believes everyone should have equal opportunities in life. & \correct{Universalism (1)} \\
\hline
3  & It’s very important to him to show his abilities. He wants people to admire what he does. & \correct{Achievement (1)} \\
\hline
4  & It is important to him to live in secure surroundings. He avoids anything that might endanger his safety. & \correct{Security (1)} \\
\hline
5  & He thinks it is important to do lots of different things in life. He always looks for new things to try. & \correct{Stimulation (1)}, \falsealarm{Self-direction (1)} \\
\hline
6  & He believes that people should do what they’re told. He thinks people should follow rules at all times, even when no one is watching. & \correct{Conformity (1)} \\
\hline
7  & It is important to him to listen to people who are different from him. Even when he disagrees with them, he still wants to understand them. & \correct{Universalism (1)}, \falsealarm{Benevolence (0.97)}, \falsealarm{Self-direction (0.62)} \\
\hline
8  & He thinks it’s important not to ask for more than what you have. He believes that people should be satisfied with what they have. & \correct{Conformity (0.99)} \\
\hline
9  & He seeks every chance he can to have fun. It is important to him to do things that give him pleasure. & \correct{Hedonism (1)}, \falsealarm{Stimulation (0.8)} \\
\hline
10 & It is important to him to make his own decisions about what he does. He likes to be free to plan and to choose his activities for himself. & \correct{Self-direction (1)} \\
\hline
11 & It’s very important to him to help the people around him. He wants to care for their well-being. & \correct{Benevolence (1)} \\
\hline
12 & Being very successful is important to him. He likes to impress other people. & \correct{Power (1)}, \falsealarm{Achievement (1)} \\
\hline
13 & It is very important to him that his country be safe. He thinks the state must be on watch against threats from within and without. & \correct{Security (1)} \\
\hline
14 & He likes to take risks. He is always looking for adventures. & \correct{Stimulation (1)}, \falsealarm{Self-direction (0.89)} \\
\hline
15 & It is important to him to always behave properly. He wants to avoid doing anything people would say is wrong. & \correct{Conformity (1)} \\
\hline
16 & It is important to him to be in charge and tell others what to do. He wants people to do what he says. & \correct{Power (1)} \\
\hline
17 & It is important to him to be loyal to his friends. He wants to devote himself to people close to him. & \correct{Benevolence (1)} \\
\hline
18 & He strongly believes that people should care for nature. & \correct{Universalism (1)} \\
\hline
19 & Religious belief is important to him. He tries hard to do what his religion requires. & \correct{Tradition (1)}, \falsealarm{Conformity (1)} \\
\hline
20 & It is important to him that things be organized and clean. He really does not like things to be a mess. & \correct{Security (1)}, \falsealarm{Conformity (1)} \\
\hline
21 & He thinks it’s important to be interested in things. He likes to be curious and to try to understand all sorts of things. & \correct{Self-direction (1)}, \falsealarm{Stimulation (0.9)} \\
\hline
22 & He believes all the world’s people should live in harmony. Promoting peace among all groups in the world is important to him. & \correct{Universalism (1)}, \falsealarm{Benevolence (0.99)} \\
\hline
23 & He thinks it is important to be ambitious. He wants to show how capable he is. & \correct{Achievement (1)} \\
\hline
24 & He thinks it is best to do things in traditional ways. It is important to him to keep up the customs he has learned. & \correct{Tradition (1)} \\
\hline
25 & Enjoying life’s pleasures is important to him. He likes to spoil himself. & \correct{Hedonism (1)} \\
\hline
26 & It is important to him to respond to the needs of others. He tries to support those he knows. & \correct{Benevolence (1)} \\
\hline
27 & He believes he should always show respect to his parents and to older people. It is important to him to be obedient. & \correct{Conformity (1)}, \falsealarm{Tradition (1)} \\
\hline
28 & He wants everyone to be treated justly, even people he doesn’t know. It is important to him to protect the weak in society. & \correct{Universalism (1)}, \falsealarm{Benevolence (1)} \\
\hline
29 & He likes surprises. It is important to him to have an exciting life. & \correct{Stimulation (1)} \\
\hline
30 & He tries hard to avoid getting sick. Staying healthy is very important to him. & \correct{Security (1)} \\
\hline
31 & Getting ahead in life is important to him. He strives to do better than others. & \correct{Achievement (1)}, \falsealarm{Power (0.82)} \\
\hline
32 & Forgiving people who have hurt him is important to him. He tries to see what is good in them and not to hold a grudge. & \correct{Benevolence (1)}, \falsealarm{Universalism (1)} \\
\hline
33 & It is important to him to be independent. He likes to rely on himself. & \correct{Self-direction (1)} \\
\hline
34 & Having a stable government is important to him. He is concerned that the social order be protected. & \correct{Security (1)} \\
\hline
35 & It is important to him to be polite to other people all the time. He tries never to disturb or irritate others. & \correct{Conformity (1)}, \falsealarm{Tradition (0.85)} \\
\hline
36 & He really wants to enjoy life. Having a good time is very important to him. & \correct{Hedonism (1)} \\
\hline
37 & It is important to him to be humble and modest. He tries not to draw attention to himself. & \correct{Tradition (0.73)}, \falsealarm{Conformity (1)} \\
\hline
38 & He always wants to be the one who makes the decisions. He likes to be the leader. & \correct{Power (1)} \\
\hline
39 & It is important to him to adapt to nature and to fit into it. He believes that people should not change nature. & \correct{Universalism (1)} \\
\hline

\end{longtable}


\begin{thebibliography}{60}
\providecommand{\natexlab}[1]{#1}

\bibitem[{Abdulhai et~al.(2024)Abdulhai, Serapio-Garc{\'i}a, Crepy, Valter,
  Canny, and Jaques}]{abdulhai2023moral}
Marwa Abdulhai, Gregory Serapio-Garc{\'i}a, Clement Crepy, Daria Valter, John
  Canny, and Natasha Jaques. 2024.
\newblock \href {https://doi.org/10.18653/v1/2024.emnlp-main.982} {Moral
  foundations of large language models}.
\newblock In \emph{Proceedings of the 2024 Conference on Empirical Methods in
  Natural Language Processing}, pages 17737--17752, Miami, Florida, USA.
  Association for Computational Linguistics.

\bibitem[{Abdurahman et~al.(2024)Abdurahman, Atari, Karimi‐Malekabadi, Xue,
  Trager, Park, Golazizian, Omrani, and Dehghani}]{Dehghani2024gpt}
Suhaib Abdurahman, Mohammad Atari, Farzan Karimi‐Malekabadi, Mona~J. Xue,
  Jackson Trager, Peter~S. Park, Preni Golazizian, Ali Omrani, and Morteza
  Dehghani. 2024.
\newblock \href {https://doi.org/10.1093/pnasnexus/pgae245} {Perils and
  opportunities in using large language models in psychological research}.
\newblock \emph{PNAS Nexus}, 3(7):pgae245.

\bibitem[{Alfano et~al.(2024)Alfano, Cheong, and Curry}]{MoralUniversals2024}
Mark Alfano, Marc Cheong, and Oliver~Scott Curry. 2024.
\newblock \href {https://doi.org/10.1016/j.heliyon.2024.e25940} {Moral
  universals: A machine-reading analysis of 256 societies}.
\newblock \emph{Heliyon}, 10(6):e25940.

\bibitem[{Amin et~al.(2017)Amin, Bednarczyk, Ray, Melchiori, Graham,
  Huntsinger, and Omer}]{aminAssociationMoralValues2017}
Avnika~B. Amin, Robert~A. Bednarczyk, Cara~E. Ray, Kala~J. Melchiori, Jesse
  Graham, Jeffrey~R. Huntsinger, and Saad~B. Omer. 2017.
\newblock Association of moral values with vaccine hesitancy.
\newblock \emph{Nature Human Behaviour}, 1(12):873--880.

\bibitem[{Bolt and Liao(2022)}]{bolt2022item}
Daniel~M. Bolt and Xiangyi Liao. 2022.
\newblock \href {https://doi.org/10.1007/s11336-022-09842-0} {Item complexity:
  A neglected psychometric feature of test items?}
\newblock \emph{Psychometrika}, 87(4):1195--1213.

\bibitem[{Clifford et~al.(2015)Clifford, Iyengar, Cabeza, and
  {Sinnott-Armstrong}}]{cliffordMoralFoundationsVignettes2015}
Scott Clifford, Vijeth Iyengar, Roberto Cabeza, and Walter {Sinnott-Armstrong}.
  2015.
\newblock Moral foundations vignettes: A standardized stimulus database of
  scenarios based on moral foundations theory.
\newblock \emph{Behavior Research Methods}, 47(4):1178--1198.

\bibitem[{Curry(2016)}]{curryMoralityCooperationProblemCentred2016}
Oliver~Scott Curry. 2016.
\newblock Morality as {{Cooperation}}: {{A Problem-Centred Approach}}.
\newblock In Todd~K. Shackelford and Ranald~D. Hansen, editors, \emph{The
  {{Evolution}} of {{Morality}}}, pages 27--51. Springer International
  Publishing, Cham.

\bibitem[{Curry et~al.(2019)Curry, Chesters, and Van~Lissa}]{curry2019mapping}
Oliver~Scott Curry, Matthew~Jones Chesters, and Caspar~J Van~Lissa. 2019.
\newblock Mapping morality with a compass: Testing the theory of
  ‘morality-as-cooperation’with a new questionnaire.
\newblock \emph{Journal of Research in Personality}, 78:106--124.

\bibitem[{Demszky et~al.(2023)Demszky, Yang, Yeager, Bryan, Clapper, Chandhok,
  Eichstaedt, Hecht, Jamieson, Johnson et~al.}]{demszky2023using}
Dorottya Demszky, Diyi Yang, David~S Yeager, Christopher~J Bryan, Margarett
  Clapper, Susannah Chandhok, Johannes~C Eichstaedt, Cameron Hecht, Jeremy
  Jamieson, Meghann Johnson, et~al. 2023.
\newblock Using large language models in psychology.
\newblock \emph{Nature Reviews Psychology}, 2(11):688--701.

\bibitem[{Dickinson et~al.(2016)Dickinson, McLeod, Bloomfield, and
  Allred}]{dickinson2016moral}
Janis~L Dickinson, Poppy McLeod, Robert Bloomfield, and Shorna Allred. 2016.
\newblock Which moral foundations predict willingness to make lifestyle changes
  to avert climate change in the usa?
\newblock \emph{PloS one}, 11(10):e0163852.

\bibitem[{Dongbo(2024)}]{dongbo2024mirt}
Tu~Dongbo. 2024.
\newblock \href {https://doi.org/10.1007/978-981-97-7874-4_37}
  {Multidimensional item response theory (mirt)}.
\newblock In Z.~Kan, editor, \emph{The ECPH Encyclopedia of Psychology}.
  Springer, Singapore.

\bibitem[{Duan et~al.(2024)Duan, Yi, Zhang, Lu, Xie, and Gu}]{duan2024denevil}
Shitong Duan, Xiaoyuan Yi, Peng Zhang, Tun Lu, Xing Xie, and Ning Gu. 2024.
\newblock \href {https://openreview.net/forum?id=m3RRWWFaVe} {Denevil: Towards
  deciphering and navigating the ethical values of large language models via
  instruction learning}.
\newblock In \emph{Proceedings of the 12th International Conference on Learning
  Representations (ICLR)}.

\bibitem[{Emelin et~al.(2021)Emelin, Le~Bras, Hwang, Forbes, and
  Choi}]{emelin2021moralstories}
Denis Emelin, Ronan Le~Bras, Jena~D. Hwang, Maxwell Forbes, and Yejin Choi.
  2021.
\newblock \href {https://doi.org/10.18653/v1/2021.emnlp-main.54} {Moral
  stories: Situated reasoning about norms, intents, actions, and their
  consequences}.
\newblock In \emph{Proceedings of the 2021 Conference on Empirical Methods in
  Natural Language Processing}, pages 698--718, Online and Punta Cana,
  Dominican Republic. Association for Computational Linguistics.

\bibitem[{Forbes et~al.(2020)Forbes, Hwang, Shwartz, Sap, and
  Choi}]{forbesSocialChemistry1012020}
Maxwell Forbes, Jena~D Hwang, Vered Shwartz, Maarten Sap, and Yejin Choi. 2020.
\newblock Social {{Chemistry}} 101: {{Learning}} to {{Reason}} about {{Social}}
  and {{Moral Norms}}.
\newblock In \emph{Proceedings of the 2020 {{Conference}} on {{Empirical
  Methods}} in {{Natural Language Processing}} ({{EMNLP}})}, pages 653--670.

\bibitem[{Frimer et~al.(2019)Frimer, Boghrati, Haidt, Graham, and
  Dehgani}]{frimerMoralFoundationsDictionary2019c}
Jeremy~A. Frimer, Reihane Boghrati, Jonathan Haidt, Jesse Graham, and Morteza
  Dehgani. 2019.
\newblock Moral foundations dictionary 2.0.
\newblock \url{https://osf.io/ezn37/}.

\bibitem[{Ganguli et~al.(2022)Ganguli, Lovitt, Kernion, Askell, Bai, Kadavath,
  Mann, Perez, Schiefer, Ndousse et~al.}]{ganguli2022redteaming}
Deep Ganguli, Liane Lovitt, Jackson Kernion, Amanda Askell, Yuntao Bai, Saurav
  Kadavath, Ben Mann, Ethan Perez, Nicholas Schiefer, Kamal Ndousse, et~al.
  2022.
\newblock \href {https://arxiv.org/abs/2209.07858} {Red teaming language models
  to reduce harms: Methods, scaling behaviors, and lessons learned}.
\newblock \emph{arXiv preprint arXiv:2209.07858}.

\bibitem[{Gert(2004)}]{gert2004common}
Bernard Gert. 2004.
\newblock \emph{Common morality: Deciding what to do}.
\newblock Oxford University Press.

\bibitem[{Gert et~al.(2006)Gert, Culver, and Clouser}]{gert2006bioethics}
Bernard Gert, Charles~M. Culver, and K.~Danner Clouser. 2006.
\newblock \href {https://academic.oup.com/book/10242} {\emph{Bioethics: A
  Systematic Approach}}, 2 edition.
\newblock Oxford University Press, New York.

\bibitem[{Graham and Haidt(2012)}]{grahamMoralFoundationsDictionary2012}
Jesse Graham and Jonathan Haidt. 2012.
\newblock Moral foundations dictionary.
\newblock \url{https://moralfoundations.org/other-materials/}.

\bibitem[{Graham et~al.(2009)Graham, Haidt, and
  Nosek}]{grahamLiberalsConservativesRely2009a}
Jesse Graham, Jonathan Haidt, and Brian~A Nosek. 2009.
\newblock Liberals and conservatives rely on different sets of moral
  foundations.
\newblock \emph{Journal of personality and social psychology}, 96(5):1029.

\bibitem[{Grimmer and Stewart(2013)}]{grimmer2013text}
Justin Grimmer and Brandon~M Stewart. 2013.
\newblock Text as data: The promise and pitfalls of automatic content analysis
  methods for political texts.
\newblock \emph{Political analysis}, 21(3):267--297.

\bibitem[{Guo et~al.(2023)Guo, Mokhberian, and
  Lerman}]{guoDataFusionFramework2023}
Siyi Guo, Negar Mokhberian, and Kristina Lerman. 2023.
\newblock \href {https://doi.org/10.1609/icwsm.v17i1.22145} {A {{Data Fusion
  Framework}} for {{Multi-Domain Morality Learning}}}.
\newblock \emph{Proceedings of the International AAAI Conference on Web and
  Social Media}, 17:281--291.

\bibitem[{Haidt(2012)}]{haidtRighteousMindWhy2012}
Jonathan Haidt. 2012.
\newblock \emph{The {{Righteous Mind}}: {{Why Good People Are Divided}} by
  {{Politics}} and {{Religion}}}.
\newblock {Vintage}.

\bibitem[{Haidt and Joseph(2004)}]{haidtIntuitiveEthicsHow2004a}
Jonathan Haidt and Craig Joseph. 2004.
\newblock Intuitive ethics: {{How}} innately prepared intuitions generate
  culturally variable virtues.
\newblock \emph{Daedalus}, 133(4):55--66.

\bibitem[{Hoover et~al.(2020)Hoover, {Portillo-Wightman}, Yeh, Havaldar,
  Davani, Lin, Kennedy, Atari, Kamel, Mendlen, Moreno, Park, Chang, Chin,
  Leong, Leung, Mirinjian, and Dehghani}]{hooverMoralFoundationsTwitter2020a}
Joe Hoover, Gwenyth {Portillo-Wightman}, Leigh Yeh, Shreya Havaldar,
  Aida~Mostafazadeh Davani, Ying Lin, Brendan Kennedy, Mohammad Atari, Zahra
  Kamel, Madelyn Mendlen, Gabriela Moreno, Christina Park, Tingyee~E. Chang,
  Jenna Chin, Christian Leong, Jun~Yen Leung, Arineh Mirinjian, and Morteza
  Dehghani. 2020.
\newblock Moral {{Foundations Twitter Corpus}}: {{A Collection}} of {{35K
  Tweets Annotated}} for {{Moral Sentiment}}.
\newblock \emph{Social Psychological and Personality Science},
  11(8):1057--1071.

\bibitem[{Hopp et~al.(2021)Hopp, Fisher, Cornell, Huskey, and
  Weber}]{hoppExtendedMoralFoundations2021}
Frederic~R. Hopp, Jacob~T. Fisher, Devin Cornell, Richard Huskey, and Ren{\'e}
  Weber. 2021.
\newblock The extended {{Moral Foundations Dictionary}} ({{eMFD}}):
  {{Development}} and applications of a crowd-sourced approach to extracting
  moral intuitions from text.
\newblock \emph{Behavior Research Methods}, 53(1):232--246.

\bibitem[{Hupkes et~al.(2023)Hupkes, Giulianelli, Dankers, Artetxe, Elazar,
  Pimentel, Christodoulopoulos, Lasri, Saphra, Sinclair
  et~al.}]{hupkes2023taxonomy}
Dieuwke Hupkes, Mario Giulianelli, Verna Dankers, Mikel Artetxe, Yanai Elazar,
  Tiago Pimentel, Christos Christodoulopoulos, Karim Lasri, Naomi Saphra,
  Arabella Sinclair, et~al. 2023.
\newblock A taxonomy and review of generalization research in {NLP}.
\newblock \emph{Nature Machine Intelligence}, 5(10):1161--1174.

\bibitem[{Ji et~al.(2023)Ji, Liu, Dai, Pan, Zhang, Bian, Chen, Sun, Wang, and
  Yang}]{ji2023socialrisks}
Jiaming Ji, Mickel Liu, Juntao Dai, Xuehai Pan, Chi Zhang, Ce~Bian, Boyuan
  Chen, Ruiyang Sun, Yizhou Wang, and Yaodong Yang. 2023.
\newblock \href
  {https://proceedings.neurips.cc/paper_files/paper/2023/hash/4dbb61cb68671edc4ca3712d70083b9f-Abstract-Datasets_and_Benchmarks.html}
  {{BeaverTails}: Towards improved safety alignment of llm via a
  human-preference dataset}.
\newblock In \emph{Advances in Neural Information Processing Systems 36
  (NeurIPS 2023), Datasets and Benchmarks Track}.

\bibitem[{Ji et~al.(2024)Ji, Chen, Jin, Xu, Hua, and
  Zhang}]{ji2024moralbenchmoralevaluationllms}
Jianchao Ji, Yutong Chen, Mingyu Jin, Wujiang Xu, Wenyue Hua, and Yongfeng
  Zhang. 2024.
\newblock \href {https://arxiv.org/abs/2406.04428} {{MoralBench}: Moral
  evaluation of {LLMs}}.
\newblock \emph{Preprint}, arXiv:2406.04428.

\bibitem[{Ji et~al.(2025)Ji, Chen, Jin, Xu, Hua, and Zhang}]{pschobench2024}
Jianchao Ji, Yutong Chen, Mingyu Jin, Wujiang Xu, Wenyue Hua, and Yongfeng
  Zhang. 2025.
\newblock \href {https://doi.org/10.1145/3748239.3748246} {Moralbench: Moral
  evaluation of {LLMs}}.
\newblock \emph{ACM SIGKDD Explorations Newsletter}, 27(1):62--71.

\bibitem[{Jiang et~al.(2022{\natexlab{a}})Jiang, Xu, Zhu, Han, Zhang, and
  Zhu}]{EvaluatingAIpersonality2022}
Guangyuan Jiang, Manjie Xu, Song-Chun Zhu, Wenjuan Han, Chi Zhang, and Yixin
  Zhu. 2022{\natexlab{a}}.
\newblock Evaluating and inducing personality in pre-trained language models.
\newblock In \emph{Neural Information Processing Systems}.

\bibitem[{Jiang et~al.(2022{\natexlab{b}})Jiang, Hwang, Bhagavatula, Bras,
  Liang, Dodge, Sakaguchi, Forbes, Borchardt, Gabriel, Tsvetkov, Etzioni, Sap,
  Rini, and Choi}]{jiang2022delphi}
Liwei Jiang, Jena~D. Hwang, Chandra Bhagavatula, Ronan~Le Bras, Jenny Liang,
  Jesse Dodge, Keisuke Sakaguchi, Maxwell Forbes, Jon Borchardt, Saadia
  Gabriel, Yulia Tsvetkov, Oren Etzioni, Maarten Sap, Regina Rini, and Yejin
  Choi. 2022{\natexlab{b}}.
\newblock \href {https://arxiv.org/abs/2110.07574} {Can machines learn
  morality? the {Delphi} experiment}.
\newblock \emph{Preprint}, arXiv:2110.07574.

\bibitem[{Kiesel et~al.(2022)Kiesel, Alshomary, Handke, Cai, Wachsmuth, and
  Stein}]{kiesel2022valueeval}
Johannes Kiesel, Milad Alshomary, Nicolas Handke, Xiaoni Cai, Henning
  Wachsmuth, and Benno Stein. 2022.
\newblock \href {https://doi.org/10.18653/v1/2022.acl-long.306} {Identifying
  the human values behind arguments}.
\newblock In \emph{Proceedings of the 60th Annual Meeting of the Association
  for Computational Linguistics (Volume 1: Long Papers)}, pages 4459--4471,
  Dublin, Ireland. Association for Computational Linguistics.

\bibitem[{Kobbe et~al.(2020)Kobbe, Rehbein, Hulpuș, and
  Stuckenschmidt}]{kobbeExploringMoralityArgumentation2020}
Jonathan Kobbe, Ines Rehbein, Ioana Hulpuș, and Heiner Stuckenschmidt. 2020.
\newblock Exploring {{Morality}} in {{Argumentation}}.
\newblock In \emph{Proceedings of the 7th {{Workshop}} on {{Argument Mining}}},
  pages 30--40.

\bibitem[{Lee et~al.(2022)Lee, Li, and Vu}]{lee2022meta}
Hung-yi Lee, Shang-Wen Li, and Thang Vu. 2022.
\newblock \href {https://doi.org/10.18653/v1/2022.naacl-main.49} {Meta learning
  for natural language processing: A survey}.
\newblock In \emph{Proceedings of the 2022 Conference of the North American
  Chapter of the Association for Computational Linguistics: Human Language
  Technologies}, pages 666--684, Seattle, United States. Association for
  Computational Linguistics.

\bibitem[{Li et~al.(2020)Li, Wen, Hau, Yuan, and Peng}]{li2020crossloadings}
Yan Li, Zhonglin Wen, Kit-Tai Hau, Ke-Hai Yuan, and Yifan~F. Peng. 2020.
\newblock \href {https://doi.org/10.1080/10705511.2020.1745075} {Effects of
  cross-loadings on determining the number of factors to retain}.
\newblock \emph{Structural Equation Modeling: A Multidisciplinary Journal},
  27(4):584--599.

\bibitem[{Liscio et~al.(2022)Liscio, Dondera, Geadau, Jonker, and
  Murukannaiah}]{liscioCrossDomainClassificationMoral2022}
Enrico Liscio, Alin Dondera, Andrei Geadau, Catholijn Jonker, and Pradeep
  Murukannaiah. 2022.
\newblock \href {https://doi.org/10.18653/v1/2022.findings-naacl.209}
  {Cross-domain classification of moral values}.
\newblock In \emph{Findings of the Association for Computational Linguistics:
  {{NAACL}} 2022}, pages 2727--2745, Seattle, United States. Association for
  Computational Linguistics.

\bibitem[{Liu et~al.(2024)Liu, Feng, Xue, Wang, Wu, Lu, Zhao, Deng, Zhang, Ruan
  et~al.}]{liu2024deepseek}
Aixin Liu, Bei Feng, Bing Xue, Bingxuan Wang, Bochao Wu, Chengda Lu, Chenggang
  Zhao, Chengqi Deng, Chenyu Zhang, Chong Ruan, et~al. 2024.
\newblock Deepseek-v3 technical report.
\newblock \emph{arXiv preprint arXiv:2412.19437}.

\bibitem[{Lu et~al.(2022)Lu, Zhao, Li, Lu, Peng, Gao, and Zhu}]{lu2022valuenet}
Y.~Lu, Y.~Zhao, J.~Li, P.~Lu, B.~Peng, J.~Gao, and S.-C. Zhu. 2022.
\newblock \href {https://doi.org/10.1609/aaai.v36i10.21368} {{ValueNet}: A new
  dataset for human value driven dialogue system}.
\newblock In \emph{Proceedings of the AAAI Conference on Artificial
  Intelligence}, volume~36, pages 11183--11191.

\bibitem[{Mokhberian et~al.(2020)Mokhberian, Abeliuk, Cummings, and
  Lerman}]{mokhberianMoralFramingIdeological2020}
Negar Mokhberian, Andr{\'e}s Abeliuk, Patrick Cummings, and Kristina Lerman.
  2020.
\newblock \href {https://doi.org/10.1007/978-3-030-60975-7_16} {Moral framing
  and ideological bias of news}.
\newblock In Samin Aref, Kalina Bontcheva, Marco Braghieri, Frank Dignum, Fosca
  Giannotti, Francesco Grisolia, and Dino Pedreschi, editors, \emph{Social
  Informatics (SocInfo 2020)}, volume 12467 of \emph{Lecture Notes in Computer
  Science}, pages 206--219. Springer International Publishing, Cham.

\bibitem[{Nguyen et~al.(2024)Nguyen, Chen, Carroll, Tran, Klein, and
  Xie}]{nguyen2024measuring}
Tuan~Dung Nguyen, Ziyu Chen, Nicholas~George Carroll, Alasdair Tran, Colin
  Klein, and Lexing Xie. 2024.
\newblock Measuring moral dimensions in social media with mformer.
\newblock In \emph{Proceedings of the International AAAI Conference on Web and
  Social Media}, volume~18, pages 1134--1147.

\bibitem[{Pawar et~al.(2024)Pawar, Park, Jin, Arora, Myung, Yadav, Haznitrama,
  Song, Oh, and Augenstein}]{pawarSurveyCulturalAwareness2024}
Siddhesh Pawar, Junyeong Park, Jiho Jin, Arnav Arora, Junho Myung, Srishti
  Yadav, Faiz~Ghifari Haznitrama, Inhwa Song, Alice Oh, and Isabelle
  Augenstein. 2024.
\newblock \href {https://doi.org/10.48550/arXiv.2411.00860} {Survey of
  {{Cultural Awareness}} in {{Language Models}}: {{Text}} and {{Beyond}}}.
\newblock \emph{Preprint}, arXiv:2411.00860.

\bibitem[{Piurko et~al.(2011)Piurko, Schwartz, and
  Davidov}]{piurko2011SchwartzPolitics}
Yilmaz Piurko, Shalom~H. Schwartz, and Eldad Davidov. 2011.
\newblock \href {https://doi.org/10.1111/j.1467-9221.2011.00828.x} {Basic
  personal values and the meaning of left‐right political orientations in 20
  countries}.
\newblock \emph{Political Psychology}, 32(4):537--561.

\bibitem[{Rathje et~al.(2024)Rathje, Mirea, Sucholutsky, Marjieh, Robertson,
  and Van~Bavel}]{rathje2024gpt}
Steve Rathje, Dan-Mircea Mirea, Ilia Sucholutsky, Raja Marjieh, Claire~E
  Robertson, and Jay~J Van~Bavel. 2024.
\newblock {GPT} is an effective tool for multilingual psychological text
  analysis.
\newblock \emph{Proceedings of the National Academy of Sciences},
  121(34):e2308950121.

\bibitem[{Read et~al.(2011)Read, Pfahringer, Holmes, and
  Frank}]{read2011classifier}
Jesse Read, Bernhard Pfahringer, Geoffrey Holmes, and Eibe Frank. 2011.
\newblock \href {https://doi.org/10.1007/978-3-642-23775-3_17} {Classifier
  chains for multilabel classification}.
\newblock In \emph{Proceedings of the European Conference on Machine Learning
  and Principles and Practice of Knowledge Discovery in Databases (ECML PKDD)},
  pages 254--269. Springer.

\bibitem[{Ren et~al.(2024)Ren, Ye, Fang, Zhang, and
  Song}]{ren-etal-2024-valuebench}
Yuanyi Ren, Haoran Ye, Hanjun Fang, Xin Zhang, and Guojie Song. 2024.
\newblock \href {https://doi.org/10.18653/v1/2024.acl-long.111} {{ValueBench}:
  Towards comprehensively evaluating value orientations and understanding of
  large language models}.
\newblock In \emph{Proceedings of the 62nd Annual Meeting of the Association
  for Computational Linguistics (Volume 1: Long Papers)}, pages 2015--2040,
  Bangkok, Thailand. Association for Computational Linguistics.

\bibitem[{Scherrer et~al.(2023)Scherrer, Shi, Feder, and
  Blei}]{ScherrerMoralChoice}
Nino Scherrer, Claudia Shi, Amir Feder, and David Blei. 2023.
\newblock \href
  {https://proceedings.neurips.cc/paper_files/paper/2023/file/a2cf225ba392627529efef14dc857e22-Paper-Conference.pdf}
  {Evaluating the moral beliefs encoded in llms}.
\newblock In \emph{Advances in Neural Information Processing Systems},
  volume~36, pages 51778--51809. Curran Associates, Inc.

\bibitem[{Schwartz(1992)}]{schwartz1992universals}
Shalom~H. Schwartz. 1992.
\newblock Universals in the content and structure of values: Theoretical
  advances and empirical tests in 20 countries.
\newblock In Mark~P. Zanna, editor, \emph{Advances in Experimental Social
  Psychology}, volume~25, pages 1--65. Academic Press.

\bibitem[{Schwartz et~al.(2010)Schwartz, Caprara, and
  Vecchione}]{schwartz2010basic}
Shalom~H. Schwartz, Gian~Vittorio Caprara, and Michele Vecchione. 2010.
\newblock Basic personal values and political orientations.
\newblock In Peter~H. Hatemi and Rose McDermott, editors, \emph{Social
  Psychology of Political Behavior}, pages 137--154. Oxford University Press.

\bibitem[{Schwartz et~al.(2001)Schwartz, Melech, Lehmann, Burgess, Harris, and
  Owens}]{schwartz2001}
Shalom~H. Schwartz, Gila Melech, Arielle Lehmann, Steven Burgess, Mari Harris,
  and Vicki Owens. 2001.
\newblock Extending the cross-cultural validity of the theory of basic human
  values with a different method of measurement.
\newblock \emph{Journal of Cross-Cultural Psychology}, 32(5):519--542.

\bibitem[{Sechidis et~al.(2011)Sechidis, Tsoumakas, and
  Vlahavas}]{sechidis2011stratification}
Konstantinos Sechidis, Grigorios Tsoumakas, and Ioannis Vlahavas. 2011.
\newblock On the stratification of multi-label data.
\newblock In \emph{Proceedings of the European Conference on Machine Learning
  and Knowledge Discovery in Databases (ECML-PKDD) Workshop on Learning from
  Multi-Label Data}, pages 145--158.

\bibitem[{Sprague et~al.(2024)Sprague, Yin, Rodriguez, Jiang, Wadhwa, Singhal,
  Zhao, Ye, Mahowald, and Durrett}]{Sprague2024ToCO}
Zayne Sprague, Fangcong Yin, Juan~Diego Rodriguez, Dongwei Jiang, Manya Wadhwa,
  Prasann Singhal, Xinyu Zhao, Xi~Ye, Kyle Mahowald, and Greg Durrett. 2024.
\newblock \href {https://api.semanticscholar.org/CorpusID:272708032} {To {CoT}
  or not to {CoT}? chain-of-thought helps mainly on math and symbolic
  reasoning}.
\newblock \emph{ArXiv}, abs/2409.12183.

\bibitem[{Tedeschi et~al.(2024)}]{tedeschi2024alertSafety}
Simone Tedeschi et~al. 2024.
\newblock \href {https://arxiv.org/abs/2404.08676} {{ALERT}: A comprehensive
  benchmark for assessing large language models' safety through red teaming}.
\newblock \emph{arXiv preprint arXiv:2404.08676}.

\bibitem[{Tennant et~al.(2025)Tennant, Hailes, and Musolesi}]{tennant2024moral}
Elizaveta Tennant, Stephen Hailes, and Mirco Musolesi. 2025.
\newblock Moral alignment for {LLM} agents.
\newblock In \emph{Proceedings of the International Conference on Learning
  Representations (ICLR)}.

\bibitem[{Trager et~al.(2022)Trager, Ziabari, Davani, Golazizian,
  {Karimi-Malekabadi}, Omrani, Li, Kennedy, Reimer, Reyes, Cheng, Wei,
  Merrifield, Khosravi, Alvarez, and
  Dehghani}]{tragerMoralFoundationsReddit2022}
Jackson Trager, Alireza~S. Ziabari, Aida~Mostafazadeh Davani, Preni Golazizian,
  Farzan {Karimi-Malekabadi}, Ali Omrani, Zhihe Li, Brendan Kennedy, Nils~Karl
  Reimer, Melissa Reyes, Kelsey Cheng, Mellow Wei, Christina Merrifield, Arta
  Khosravi, Evans Alvarez, and Morteza Dehghani. 2022.
\newblock \href {https://arxiv.org/abs/2208.05545} {The {{Moral Foundations
  Reddit Corpus}}}.
\newblock \emph{arXiv preprint arXiv:2208.05545}.

\bibitem[{Yao et~al.(2024{\natexlab{a}})Yao, Yi, Gong, Wang, and
  Xie}]{yao2024value}
Jing Yao, Xiaoyuan Yi, Yifan Gong, Xiting Wang, and Xing Xie.
  2024{\natexlab{a}}.
\newblock \href {https://doi.org/10.18653/v1/2024.naacl-long.486} {{Value
  FULCRA}: Mapping large language models to the multidimensional spectrum of
  basic human values}.
\newblock In \emph{Proceedings of the 2024 Conference of the North American
  Chapter of the Association for Computational Linguistics: Human Language
  Technologies (Volume 1: Long Papers)}, pages 8762--8785, Mexico City, Mexico.
  Association for Computational Linguistics.

\bibitem[{Yao et~al.(2024{\natexlab{b}})Yao, Yi, and Xie}]{yao2024clave}
Jing Yao, Xiaoyuan Yi, and Xing Xie. 2024{\natexlab{b}}.
\newblock \href {https://arxiv.org/abs/2407.10725} {{CLAVE}: An adaptive
  framework for evaluating values of llm generated responses}.
\newblock In \emph{Proceedings of the 38th Conference on Neural Information
  Processing Systems (NeurIPS 2024), Datasets and Benchmarks Track}.

\bibitem[{Zheng et~al.(2024)Zheng, Wang, Zhang, Duy~Tai, Sun, and
  Chua}]{zheng2024ali}
Jingnan Zheng, Han Wang, An~Zhang, Nguyen Duy~Tai, Jun Sun, and Tat-Seng Chua.
  2024.
\newblock \href
  {https://proceedings.neurips.cc/paper_files/paper/2024/file/b35c38f70065ac6c694089ca93a015bb-Paper-Conference.pdf}
  {{ALI-Agent}: Assessing {LLMs'} alignment with human values via agent-based
  evaluation}.
\newblock In \emph{Advances in Neural Information Processing Systems},
  volume~37, pages 99040--99088. Curran Associates, Inc.

\bibitem[{Zhou et~al.(2024)Zhou, Hu, Li, Zhang, Wu, King, and
  Meng}]{zhou-etal-2024-rethinking}
Jingyan Zhou, Minda Hu, Junan Li, Xiaoying Zhang, Xixin Wu, Irwin King, and
  Helen Meng. 2024.
\newblock \href {https://doi.org/10.18653/v1/2024.findings-naacl.144}
  {Rethinking machine ethics {--} can {LLM}s perform moral reasoning through
  the lens of moral theories?}
\newblock In \emph{Findings of the Association for Computational Linguistics:
  NAACL 2024}, pages 2227--2242, Mexico City, Mexico. Association for
  Computational Linguistics.

\bibitem[{Ziems et~al.(2022)Ziems, Yu, Wang, Halevy, and
  Yang}]{ziemsMoralIntegrityCorpus2022}
Caleb Ziems, Jane Yu, Yi-Chia Wang, Alon Halevy, and Diyi Yang. 2022.
\newblock The {{Moral Integrity Corpus}}: {{A Benchmark}} for {{Ethical
  Dialogue Systems}}.
\newblock In \emph{Proceedings of the 60th Annual Meeting of the Association
  for Computational Linguistics (Volume 1: {{Long}} Papers)}, pages 3755--3773,
  {Dublin, Ireland}. {Association for Computational Linguistics}.

\end{thebibliography}
\end{document}